\documentclass{article}

\PassOptionsToPackage{numbers, compress}{natbib}

 \usepackage[main, final]{neurips_2026}

\usepackage[utf8]{inputenc} 
\usepackage[T1]{fontenc}    
\usepackage{hyperref}       
\usepackage{url}            
\usepackage{booktabs}       
\usepackage{amsfonts}       
\usepackage{nicefrac}       
\usepackage{microtype}      
\usepackage{xcolor}         

\usepackage{amsmath}
\usepackage{amssymb}
\usepackage{mathtools}
\usepackage{amsthm}

\usepackage[capitalize,noabbrev]{cleveref}

\newcommand{\eg}{\emph{e.g.}}
\newcommand{\ie}{\emph{i.e.}}

\newcommand{\etc}{\emph{etc}}

\usepackage{dsfont} 
\usepackage{graphicx} %
\usepackage{float}
\usepackage{multicol}
\usepackage{multirow}
\usepackage{caption}
\usepackage{subcaption}
\usepackage{tabularx}
\usepackage{array} %
\usepackage{utfsym}
\usepackage[most]{tcolorbox}
\usepackage{listings}
\tcbuselibrary{listings}
\usepackage{enumitem}

\usepackage{fvextra}
\usepackage{makecell}

\newcommand{\bulletpara}[1]{%
    \noindent\textbullet\quad #1 %
}
\newcommand{\vrulewidth}{0.5pt}

\theoremstyle{plain}

\newtheorem{remark}{Remark}

\title{On the Faithfulness of Visual Thinking: \\ Measurement and Enhancement}

\author{
    \textbf{Zujing~Liu\textsuperscript{1,2}$^*$, ~Junwen~Pan\textsuperscript{2}$^*$}, ~Qi~She\textsuperscript{2}$^\ddag$, 
    ~Yuan~Gao\textsuperscript{1}$^\dag$,
    ~Gui-Song~Xia\textsuperscript{1}$^\dag$\\
    \textsuperscript{1}School of Artificial Intelligence, Wuhan University \qquad
    \textsuperscript{2}ByteDance \\
}

\begin{document}

\renewcommand*{\thefootnote}{\fnsymbol{footnote}}
\footnotetext{* Equal contributions. Work done during Zujing Liu's internship at ByteDance under the mentorship of Qi She.}
\footnotetext{$\ddagger$ Project leader.}
\footnotetext{$\dagger$ Corresponding authors.}
\renewcommand*{\thefootnote}{\arabic{footnote}}

\maketitle

\begin{abstract}
Recent large vision–language models (LVLMs) can generate vision–text multimodal chain-of-thought (MCoT) traces after reinforcement fine-tuning (RFT). However, we observe that the visual information in MCoT is often inaccurate even when answers are correct, and task accuracy remains nearly unchanged despite the visual information being corrupted. This indicates a lack of faithfulness in the vision part of MCoT reasoning.
We attribute this to \emph{the RL reward design in RFT, which solely incentivizes the format of interleaved vision-text cues}, encouraging the model to incorporate visual information into its text reasoning steps without considering its correctness. In this paper, we first probe MCoT faithfulness by measuring how much the prediction changes when its visual and textual thoughts are intervened. Surprisingly, the model's predictions remain nearly unchanged under visual intervention but change significantly under textual intervention, indicating that \textbf{the visual evidence is largely ignored}.
To further analyze the visual information, we introduce an automated LVLM-based evaluation metric that quantifies the faithfulness of visual cues from two perspectives: relevance and sufficiency. Our evaluation reveals that the visual information in current MCoT traces is simultaneously irrelevant and insufficient.
To address this issue, we propose Sufficient-Component Cause Model (SCCM) learning. This approach encourages the MCoT to generate sufficient yet minimal visual components that are \textbf{independently capable of leading to the correct answer}.
The proposed SCCM is annotation-free and plug-and-play, compatible with various RFT for MCoT. Empirical results demonstrate that SCCM substantially improves the visual faithfulness across a suite of fine-grained perception and reasoning benchmarks.
The code is available at \url{https://github.com/EugeneLiu01/Faithful_Thinking_with_Image}.

\end{abstract}

\vspace{-6mm}
\section{Introduction}
\vspace{-3mm}
Multimodal Chain-of-Thought (MCoT) reasoning marks a pivotal advancement in the capabilities of Large Vision-Language Models (LVLMs), specifically enhancing the interpretability and intuitiveness of their reasoning processes for human users \citep{wang2025multimodal}. Unlike conventional text-only Chain-of-Thought (CoT) \citep{wei2022chain, team2025kimi, guo2025seed1}, vision–text MCoT fundamentally integrates the visual modality into the reasoning pathway, mirroring human cognition, which inherently fuses vision and language information \citep{baddeley2012working,paivio2013imagery}. By grounding reasoning in both visual and textual contexts, MCoT provides a more transparent and relatable cognitive process, making complex model outputs more accessible and understandable.

Recent breakthroughs have demonstrated the potential of MCoT following the ``thinking with images'' paradigm \citep{su2025thinking,hu2024visual,su2025openthinkimg,openai2025o3}. A promising direction of involving ``image thoughts'' in reasoning is to utilize the pretrained visual grounding ability, by encapsulating it in an image zoom-in tool, enabling it to actively gather information from the original images via tool calls in an \emph{agentic paradigm} \citep{plaat2025agentic,qian2025toolrl}. This design facilitates reinforcement fine-tuning (RFT) \citep{schulman2017proximal,shao2024deepseekmath} in an agentic manner \citep{hu2024visual, li2025torl}, which can significantly improve the tool-calling ability of LLMs, such as RAG \citep{jin2025search} and AI Agent \citep{luo2025agent}, \etc.
Consequently, work represented by DeepEyes \citep{zheng2025deepeyes} and Pixel-Reasoner \citep{su2025pixel} has achieved promising performance on various fine-grained perception and reasoning benchmarks like \textit{V*} Bench \citep{wu2024v}.

\begin{figure*}[t]
    \centering
    \vspace{-4mm}
    \begin{minipage}[t]{0.76\textwidth}
        \centering
        \includegraphics[width=\textwidth]{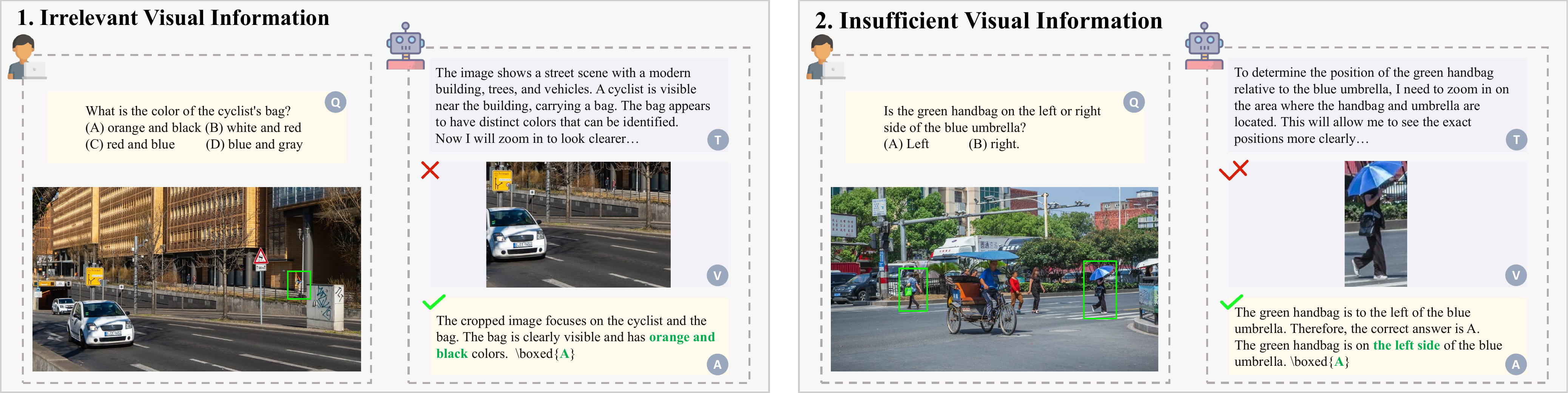}\vspace{-2mm}
        \subcaption{}
        \label{fig:failcases}
    \end{minipage}
    \begin{minipage}[t]{0.23\textwidth}
        \centering
        \includegraphics[width=\textwidth]{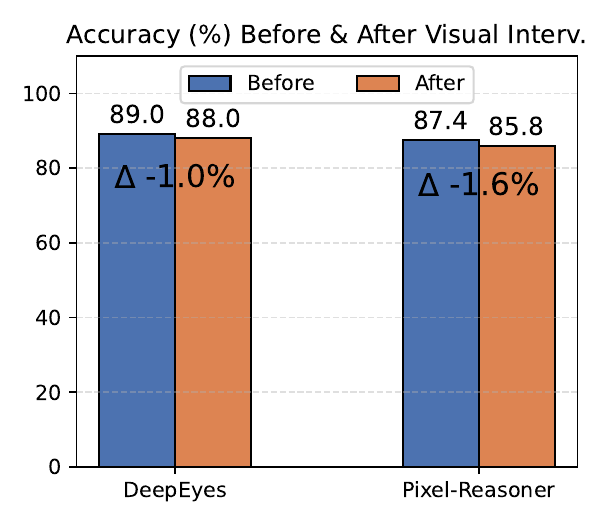}\vspace{-2mm}
        \subcaption{}
        \label{fig:changerates_tv}
    \end{minipage}
    \vspace{-3mm}
    \caption{\textbf{Unfaithfulness problem in ``Thinking with Images'' diagrams.}
    (a) Visual thinking failure modes, including irrelevant or insufficient visual information, where visual reasoning is misaligned with the final answer.
    (b) Quantitative analysis on \textit{V*} Bench with current work \citep{zheng2025deepeyes, su2025pixel}. We present the accuracy before and after interventions (\ie, mislocalized image region) on the \textbf{Visual} information within MCoT, showing that visual information intervention induced negligible change in accuracy.}\label{fig:mcoT_analysis}\vspace{-6.5mm}
\end{figure*}

However, we found that there are obvious mistakes in the generated MCoT by case study on \textit{V*} Bench, primarily divided into two categories: 1) irrelevant visual information; 2) incomplete and insufficient visual information, as illustrated in Figure \ref{fig:failcases}.
Furthermore, we intervened on the visual information in MCoT by replacing the original zoom-in crops with mislocalized image regions of incorrect bounding boxes, which semantically corrupts the visual information, and observed the accuracy changes on \textit{V*} Bench. Surprisingly, as shown in Figure \ref{fig:changerates_tv}, intervening on the visual information induced only a negligible change in accuracy.
This evidence demonstrates that inaccurate and insufficient visual information in MCoT may still yield definite, even accurate answers, suggesting that \textbf{the MCoT can be unfaithful}.
We attribute this unfaithfulness to the most widely used RL reward design \citep{zheng2025deepeyes,su2025pixel} \emph{which only encourages the \textbf{presence} rather than the \textbf{correctness and sufficiency} of interleaved visual cues}. As a result, it can be easily hacked by introducing arbitrarily ineffective visual cues without query-related content, and deriving the final answer based solely on the textual reasoning. This issue is more pronounced when given easy queries, where additional visual cues often offer limited benefit. %

The above analysis motivates us to make an in-depth evaluation of the faithfulness of MCoT. Specifically, we first probe the faithfulness of MCoT through intervention \citep{hagmayer2007causal} on its visual and textual parts, respectively, measuring how much the prediction changes when its visual and textual thoughts are corrupted. 
Notably, the model's predictions remain nearly unchanged under visual intervention but change severely with textual intervention, suggesting that \emph{visual evidence can be largely ignored and thus contributes less to the model's predictions than textual evidence}. 
To further diagnose the visual information in MCoT, we introduce an automated LVLM-based evaluation pipeline that quantifies faithfulness from two perspectives: relevance and sufficiency. Specifically, with an external LVLM as a judge, 1) it determines whether the visual components are \emph{relevant} to the model's prediction; and 2) for \emph{sufficiency}, it judges whether the visual components can correctly answer the user's query.
We conducted extensive evaluations to assess the visual faithfulness of MCoT generated by representative multimodal reasoning models \citep{zheng2025deepeyes, su2025pixel},
revealing that the visual components in MCoT are oftentimes insufficient for correct answers, and even unrelated to the model's final predictions.

To address this issue, we propose Sufficient-Component Cause Model (SCCM) learning \citep{rothman1976causes,flanders2006relationship}, which enforces the visual components to be \emph{sufficient-and-minimal} for correct answers, \ie, 1) the correct answer can be derived \emph{solely} from the visual components of MCoT, focusing on semantic accuracy, and 2) the visual components contain no extra information that is unrelated to correct answers, emphasizing semantic conciseness.
This design offers three advantages: 1) it encourages robust visual reasoning by requiring visual evidence to independently yield correct answers, thereby avoiding over-reliance on textual reasoning that disregards visual cues;
2) it enhances MCoT faithfulness by ensuring the correctness of visual cues, leading to rigorous visual reasoning; and 3) it facilitates a more traceable reasoning process and provides a more intuitive understanding of predictions.

The proposed SCCM learning is annotation-free and compatible with various RFT training for MCoT, substantially improving faithfulness metrics across representative fine-grained perception and reasoning benchmarks. Our main contributions include:
\begin{itemize}[leftmargin=*, itemsep=0pt, topsep=-1.5mm, parsep=0pt]
\item We reveal the unfaithfulness problem of vision-text MCoT where \emph{visual evidence is largely ignored}, and introduce an evaluation pipeline to quantify the faithfulness of MCoT.
\item We propose Sufficient-Component Cause Model learning, a simple and effective reward mechanism that enhances the multimodal reasoning ability by improving the faithfulness of the MCoT. 
\end{itemize}

\section{Related Work} \label{sec:related_work}

\textbf{Vision-language Models Reasoning.} Chain-of-Thought (CoT) \citep{wei2022chain} has been widely recognized as a key technology for enhancing the reasoning capabilities of large language models (LLMs). Inspired by the success of \citet{guo2025deepseek}, researchers are actively exploring the application of similar reinforcement learning approaches to large vision–language models (LVLMs) \citep{peng2025lmm,zhang2025r1,liu2025visual}. %
\citet{wang2025traceable} propose textual bounding boxes as traceable evidence to enhance visual grounding reasoning and leverage high-quality bounding box annotations to design IoU rewards in reinforcement learning, which primarily focus on the spatial accuracy of visual grounding and entail considerable annotation overhead.
While this approach actively incorporates visual cues into its reasoning process, the reasoning remains \textit{purely textual} and has not yet fully explored the distinctive reasoning paradigms that LVLMs may support, \eg, incorporating visual evidence explicitly into the reasoning process.

\textbf{Thinking with Images.} Unlike text-only reasoning that treats vision as a static, initial context \citep{su2025thinking}, the ``thinking with images'' paradigm actively leverages visual information as intermediate steps in the reasoning process, through various extrinsic operations, \eg, toolkit and code executor \citep{shen2024zoomeye,su2025openthinkimg,zheng2025deepeyes, openai2025o3} and intrinsic generation or imagination \citep{chern2025thinking, xu2025visual}. A promising paradigm is involving visual information in an agentic manner \citep{zheng2025deepeyes, su2025pixel}, which gathers key information from the original images via targeted tool calling, such as zoom-in tool. Despite these initial advances, the validity and reliability of such visual reasoning paradigms remain underexplored.

\textbf{Reasoning Faithfulness.} Faithfulness is formally defined as how well the stated explanation accurately reflects the actual reasoning process of the model. It has received sustained research attention in LLMs \citep{bao2024likely, tanneru2024hardness, paul2024making, barez2025chain}, and its evaluation is non-trivial due to the large parameter scale and the black-box nature of LLMs \citep{agarwal2024faithfulness}. \citet{lanham2023measuring} apply different interventions to the CoT and observe the resulting changes in final answers to evaluate CoT faithfulness. \citet{xiong2025measuring} utilize counterfactual interventions to investigate the faithfulness of the reasoning process.
In LVLMs, faithfulness is complicated by a long-standing issue: models may rely on linguistic priors or text shortcuts, leading to insufficient grounding in visual evidence \citep{goyal2017making, agrawal2018don, hsieh2023sugarcrepe, yu2025vfaith}. 
Prior work mainly examines input-level visual neglect, \ie, whether models utilize visual inputs when making predictions \citep{parcalabescu2023mm, zhang2024mathverse, chen2024we, parcalabescu2025vision}, rather than whether visual information is faithfully incorporated throughout the reasoning process. This gap is critical for emerging paradigms such as ``thinking with images'', where visual information is not only part of the initial input but can also be actively retrieved and revisited during reasoning.
Thus, beyond input-level grounding, it is crucial to examine whether models genuinely integrate such retrieved visual evidence into their intermediate reasoning, \ie, reasoning-level visual faithfulness, which remains largely underexplored.

\vspace{-1mm}
\section{Preliminary}

\begin{figure*}[ht]
\vspace{-2mm}
    \centering
    \begin{minipage}[t]{0.71\textwidth} 
        \includegraphics[width=\linewidth]{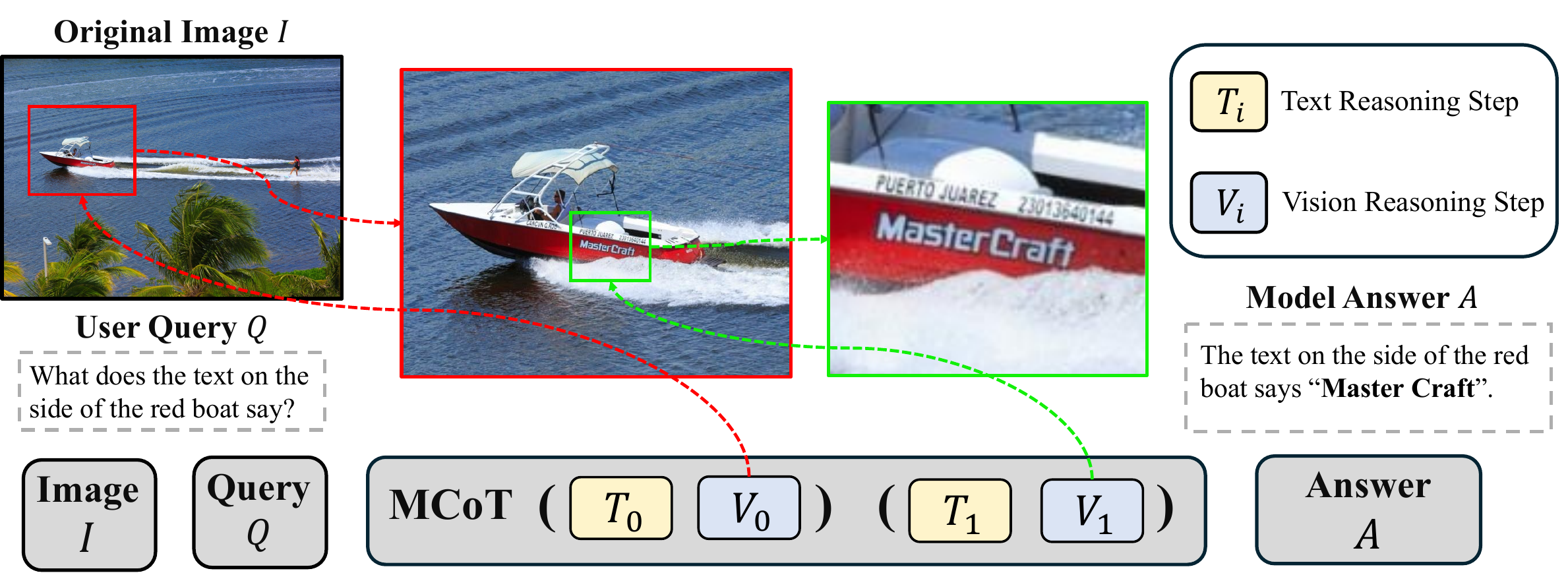}
        \subcaption{}
        \label{fig:think-with-image}
    \end{minipage}
    \hspace{2mm}
    \begin{minipage}[t]{0.25\textwidth} 
        \includegraphics[width=\linewidth]{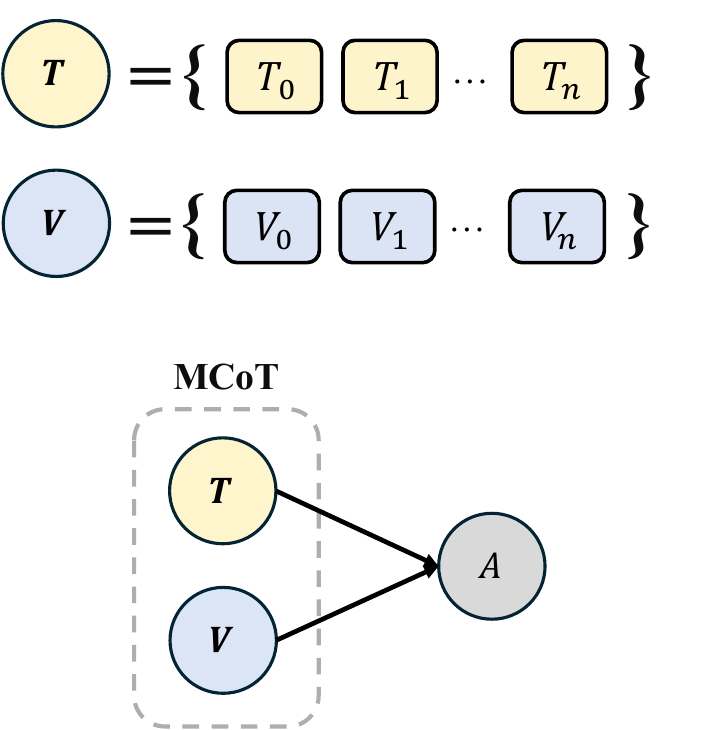}
        \subcaption{}
        \label{fig:scm}
    \end{minipage}
    \vspace{-2mm}
    \caption{(a) The illustration of ``thinking with images'' paradigm. The visual information in MCoT is introduced by zoom-in tool call as observation tokens, and appended to the ongoing reasoning process to guide subsequent steps; (b) The Structural Causal Model (SCM) of MCoT and the answer. The directed arrow between two nodes indicates a causal dependency between them, which can be analyzed via intervention experiments.} \label{fig:mcot_scm}
    \vspace{-4mm}
\end{figure*}

\textbf{Definition of Multimodal Chain-of-Thought (MCoT).}
We note that in the agentic ``thinking with images'' paradigm with visual grounding and tool calling \citep{zheng2025deepeyes,su2025pixel}, the visual information is incorporated via image zoom-in tool calls rather than being generated by the model. Thus, the visual information can be regarded as observation tokens, which are appended to the ongoing reasoning process to guide subsequent reasoning steps.
For Visual Question-Answering tasks, given an input image $I$ and a user query $Q$, the agentic multimodal reasoning process can be formulated as:
\begin{align}
    \mathbf{y} = \{(T_0, V_0), (T_1, V_1), ..., (T_t, V_t), A \ | \ I, Q\} \label{eq:mcot}
\end{align}
where $\textbf{T} = \{T_0, T_1, ..., T_t\}$ and $\textbf{V} = \{V_0, V_1, ..., V_t\}$ represent the textual and visual reasoning steps respectively, and $A$ is the final answer in the model's response $\mathbf{y}$. Therefore, the MCoT is defined as $\textbf{MCoT} = \{ (T_0, V_0), (T_1, V_1), ..., (T_t, V_t) \}$. Figure \ref{fig:think-with-image} shows the ``thinking with images'' paradigm.

\textbf{MCoT Faithfulness.} Faithfulness demands that the stated reasoning accurately, completely, and faithfully reflect the model's actual reasoning process, \ie, it accurately represents the reasoning process behind the model's prediction \citep{jacovi2020towards}. %
Specifically for ``thinking with images'' MCoT, the faithfulness is manifested in 1) \emph{Causal Consistency}. The textual $\mathbf{T}$ and visual $\mathbf{V}$ components of MCoT should be causally relevant to the final answer $A$, rather than merely forming fictitious associations with it;
2) \emph{Information Sufficiency}. Both the textual $\mathbf{T}$ and visual $\mathbf{V}$ components of MCoT independently retain sufficient information from the input $I$ and $Q$ to derive the correct answer. Otherwise, it indicates that MCoT has fabricated or omitted information.

\section{MCoT Faithfulness Evaluation} \label{sec:faithfulness_eval_pipeline}
\vspace{-2.5mm}
We first probe MCoT faithfulness by intervention experiments to analyze the influence of the textual and visual components of MCoT on the predicted answers in Sect. \ref{sec:interv_analysis}, revealing that the visual evidence is largely ignored. Sect. \ref{sec:quant_measuring} details our automated evaluation pipeline that quantifies the faithfulness of MCoT's visual components in terms of relevance and sufficiency.

\vspace{-2mm}
\subsection{Intervention Analysis} \label{sec:interv_analysis}
\vspace{-2mm}

We conduct intervention experiments to assess the influence of textual/visual components on predicted answers by measuring how much the final predicted answer changes after intervening on these components of the generated MCoT. Figure \ref{fig:interv_case} in Appendix \ref{app:how_intervention} shows an example.
This analysis can be interpreted within the Structural Causal Model (SCM) framework \citep{pearl2009causality}, illustrated in Figure \ref{fig:scm}.
In our setting, the textual components $\mathbf{T}$ and visual components $\mathbf{V}$ of MCoT are treated as intermediate variables in the reasoning process, and the predicted answer $A$ is the outcome.

Under the SCM framework, the effect of a variable is characterized by comparing outcomes under intervention and non-intervention, \eg, through the Average Treatment Effect \citep{rubin1974estimating}.
Inspired by this, we define \emph{Textual Effect} and \emph{Visual Effect} under the intervention notation as diagnostic metrics.\footnote{This analysis is intended as a diagnostic probe of model behavior under controlled perturbations, rather than strict causal identification. Nevertheless, it offers useful insights into the impact of textual and visual components on model predictions.}

\textbf{Textual Effect (TE).} It measures how much the model's predicted answer changes under \emph{intervention on textual components}, while keeping \emph{the visual components fixed}. 
\begin{equation}
\begin{aligned}
    TE = \mathbb{E}[A | do(\mathbf{T}), \mathbf{V}] - \mathbb{E}[A | \mathbf{T}, \mathbf{V}]
\end{aligned}
\end{equation}
Inspired by \citet{lanham2023measuring}, the intervened textual components $do(\mathbf{T})$ are approximated by injecting mistakes in the original text content with minor modification. We employ GPT-4o \citep{openai2024gpt4o} for mistake injection, with the corresponding prompt detailed in Appendix \ref{app:prompt_mistake_inject}.

\textbf{Visual Effect (VE).} It measures how much the model's predicted answer changes under \emph{intervention on visual components}, while keeping \emph{the textual components fixed}. 
\begin{equation}
\begin{aligned}
    VE = \mathbb{E}[A | do(\mathbf{V}), \mathbf{T}] - \mathbb{E}[A | \mathbf{V}, \mathbf{T}]
\end{aligned}
\end{equation}
For the visual component intervention $do(\mathbf{V})$, we approximate it by replacing the cropped image from zoom-in tool call in MCoT with a mislocalized crop, obtained by randomly shifting and scaling the original bounding box, while enforcing its IoU with the original box below a threshold, \eg, 0.1.

\begin{remark} \label{remark:rm1}
The proposed Textual/Visual Effect captures how much the predicted answer changes under specific interventions. However, token-level comparison of model predictions for each query is computationally complex.
To simplify the analysis, we convert the model's predicted answer into an accuracy measure (1 for correct, 0 for incorrect).
The effect is then defined as the difference between the mean accuracy of all queries with and without intervention.
This formulation also enables McNemar's test \citep{mcnemar1947note} to evaluate the statistical significance of the observed changes.
\end{remark}

\begin{figure*}[ht]
  \vspace{-1mm}
    \centering
    \begin{minipage}[t]{0.48\linewidth} 
        \centering
        \subcaption{DeepEyes}\vspace{-0.5mm}
        \includegraphics[width=\linewidth]{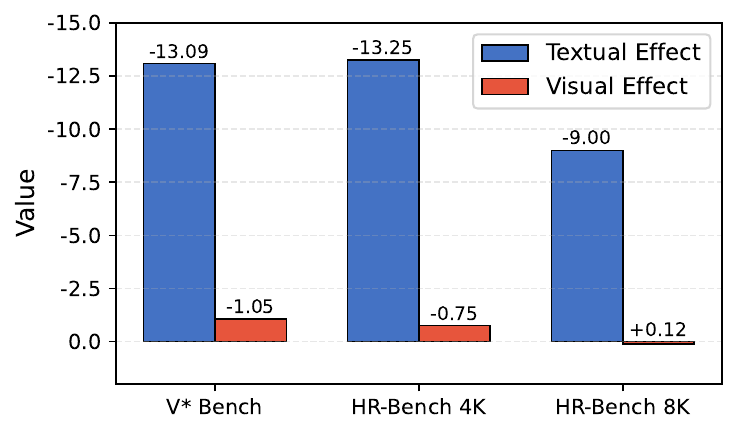}
    \end{minipage}
    \begin{minipage}[t]{0.48\linewidth} 
        \centering
        \subcaption{Pixel-Reasoner}\vspace{-0.5mm}
        \includegraphics[width=\linewidth]{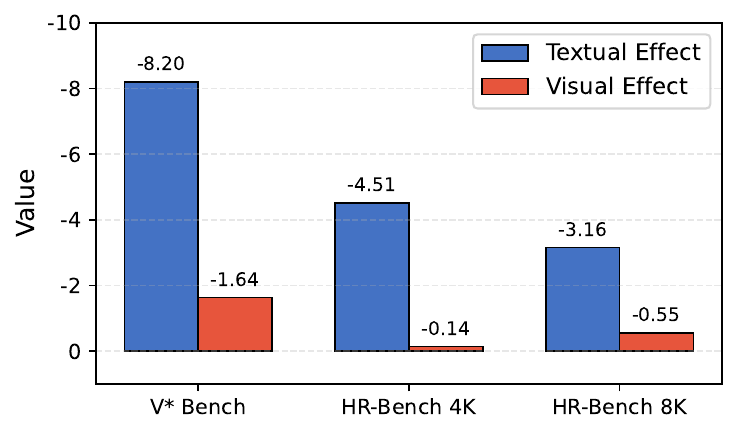}
    \end{minipage}
    \vspace{-3mm}
    \caption{Textual vs. visual effect across benchmarks for DeepEyes and Pixel-Reasoner.} %
    \label{fig:compare_te_ve}
    \vspace{-6mm}
\end{figure*}

Comparing the textual and visual effect across benchmarks for DeepEyes and Pixel-Reasoner in Figure \ref{fig:compare_te_ve}, we observe that the visual effect consistently exhibits much smaller magnitudes than the textual effect. This indicates that \emph{the visual information in MCoT has a limited impact on the model's underlying reasoning process} (\ie, the reasoning process relies primarily on textual information), suggesting that \textbf{the visual information is less faithful}. Interestingly, in some cases, interventions on visual components even lead to slight performance improvements, implying that visual information may not always bring beneficial impacts and useful evidence for the reasoning process.
Detailed results of intervention experiments can be found in Table \ref{tab:causal} of Sect. \ref{sec:interv_exp}.

\vspace{-2.5mm}
\subsection{Quantifying Faithfulness: Relevance and Sufficiency} \label{sec:quant_measuring}
\vspace{-1.5mm}

In this section, we propose a quantitative evaluation pipeline of visual faithfulness for further investigation. %
This pipeline assesses visual faithfulness from two perspectives: \emph{relevance and sufficiency}. For automated evaluation, a third-party LVLM, \ie, GPT-4o is involved as a judge.

\textbf{Relevance.} It evaluates whether the visual components $\mathbf{V}$ of MCoT are relevant to the predicted answer $A$, \ie, whether they provide visual evidence that supports the answer. 
We leverage GPT-4o as a judge to assess the relevance of visual components for the predicted answer, denoted as $\mathcal{J}_R(\mathbf{V}, A)$. The model outputs `Yes' if $\mathbf{V}$ is relevant to and supports $A$, and `No' otherwise.
The prompt for relevance assessment is detailed in Appendix \ref{app:prompt_relevance}. It is formally defined as:
\begin{equation}
\begin{aligned}
\text{Rel}(\mathbf{V}, A) = \mathds{1} \left [ \mathcal{J}_R(\mathbf{V}, A) = \text{`Yes'} \right ]
\end{aligned} \label{eq:relevance}
\end{equation}

\textbf{Sufficiency.} It evaluates whether the visual components $\mathbf{V}$ of MCoT contain sufficient information to correctly answer the given question. It is a prerequisite for accurate prediction, while also a key indicator of faithful MCoT reasoning that contains no fabricated or omitted information.
GPT-4o is employed again for predicting a new answer from only visual components $\mathbf{V}$, denoted $\mathcal{J}_S(\mathbf{V})$, in which the prompt is detailed in Appendix \ref{app:prompt_sufficiency}. With ground-truth answer $A_{\mathcal{GT}}$, the sufficiency is derived from the accuracy of the new answer, formally defined as:
\begin{equation}
\begin{aligned}
\text{Suf}(\mathbf{V}) = \mathds{1} \left [ \mathcal{J}_S(\mathbf{V}) = A_{\mathcal{GT}} \right ]
\end{aligned} \label{eq:sufficiency}
\end{equation}

\begin{remark}
    Due to the inherent randomness in the judge model, we perform multiple rounds of judgment and aggregate the results across repetitions to enhance the robustness and reliability of the final outcomes. Available aggregation approaches include 1) majority voting and 2) averaging. We also assess the judge's reliability via quantitative agreement with human annotation in Appendix \ref{app:judger_reliability}.
\end{remark}

\begin{figure*}[ht]
    \centering
    \vspace{-2mm}
    \includegraphics[width=0.89\linewidth]{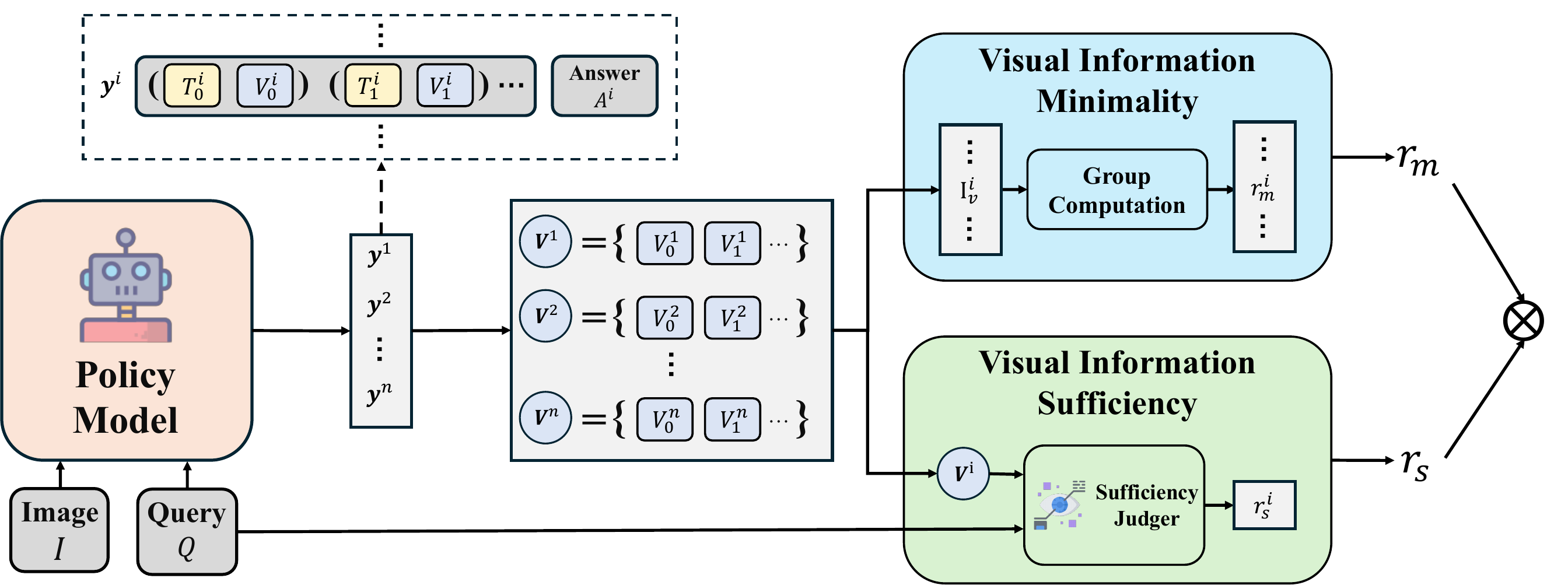}
    \vspace{-1.5mm}
    \caption{The overview of our proposed Sufficient-Component Cause Model (SCCM) learning to establish visual information as sufficient-component causes to correct answers. The SCCM framework requires that: 1) the visual information alone is sufficient to lead to the correct answer, enforced by the \textbf{Visual Information Sufficiency} reward $r_s$; and 2) the visual information involved is as minimal as possible, guided by the \textbf{Visual Information Minimality} reward $r_m$.} \label{fig:method_overview}
    \vspace{-4.5mm}
\end{figure*}

\vspace{-3mm}
\section{Sufficient-Component Cause Modeling} \label{sec:sccm}
\vspace{-2mm}

\textbf{Pitfalls of Existing Methods.}
Based on the proposed evaluation pipeline, we extensively assess current works \citep{zheng2025deepeyes,su2025pixel}, as detailed in Sect. \ref{sec:interv_exp} and \ref{sec:faithful_quant_exp}.
Our evaluation reveals that existing methods exhibit less faithfulness in the visual information of their MCoT. Specifically, this visual information is oftentimes insufficient for correct answers, which might even be unrelated to the model's final predictions. This suggests that the visual information has minimal impact on the underlying reasoning process.
\emph{We attribute this unfaithfulness to their RL reward design}, which only incentivizes the presence of interleaved visual cues via zoom-in tool call, while neglecting the correctness and sufficiency of those visual cues. In other words, their RL reward design may encourage arbitrary visual information generated by the zoom-in tool. This design flaw makes the reward easily hacked by introducing ineffective visual cues and deriving the final answer based solely on the textual reasoning. This case is likely to occur in the earlier stages when visual reasoning ability is underdeveloped, which subsequently makes the visual information largely ignored in the well-trained MCoT, and the model ultimately relies solely on its stronger textual reasoning for the actual reasoning process.

\textbf{Toward Mitigating Visual Unfaithfulness.}
To improve the accuracy of visual information involved in MCoT and encourage the model to correctly reason with it (\ie, ``thinking with images''), we propose Sufficient-Component Cause Model (SCCM) \citep{rothman1976causes} learning to establish visual information as a sufficient-component cause for correct answers in RFT training, with SCCM serving as a conceptual foundation.
SCCM learning encourages 1) \emph{Sufficiency}: the visual components in the MCoT must be sufficient to derive the correct answer; and 2) \emph{Minimality}: these sufficient components should be as minimal as possible\footnote{In SCCM, minimality bridges sufficiency to necessity: removing any component would invalidate the sufficient cause. Since the visual components in MCoT are instantiated through zoom-in crops, we operationalize this principle by minimizing the total visual information quantity, encouraging sufficient visual evidence to be compact and less redundant.}, without extra irrelevant information (\ie, the zoom-in bounding box should be tightest).
The overview of our SCCM learning is illustrated in Figure \ref{fig:method_overview}.

\textbf{Visual Information Sufficiency.} Our primary objective is to ensure that the visual information $\mathbf{V}$ in MCoT becomes independently sufficient to produce a correct answer. Formally, this requires $\text{Suf}(\mathbf{V}) = \mathds{1} \left [ \mathcal{J}_S(\mathbf{V}) = A_{\mathcal{GT}} \right ]$ \footnote{We note that for simplicity, we adopt the definition of $\mathcal{J}_S(\cdot)$ in Eq. \ref{eq:sufficiency}, while using a more cost-effective model, \ie, Qwen2.5-VL-72B, with simplified prompts for computational efficiency, detailed in Appendix \ref{app:training_details}.}, then the sufficiency reward is: 
\begin{equation}
\begin{aligned}
    r_{\text{s}}(\mathbf{y}^i) & = \text{Suf}(\mathbf{V}^i) =\mathds{1} \left [ \mathcal{J}_S(\mathbf{V}^i) = A_{\mathcal{GT}} \right ]
\end{aligned} \label{eq:suff_rwd}
\end{equation}
where $\mathbf{y}^i$ is the $i$-th rollout of the responses with MCoT, from the same input image $I$ and query $Q$. $\mathbf{V}^i$ is the visual components extracted from the MCoT of response $\mathbf{y}^i$. 

Visual information sufficiency offers several key benefits: (1) it imposes explicit supervision on visual components, encouraging the model to incorporate accurate and effective visual information; (2) it requires no additional annotations, such as ground-truth bounding boxes, making it widely applicable and plug-and-play without extra constraints on the training data; (3) by improving the correctness of visual cues, the model becomes more capable of deriving answers from the visual information itself,
enhancing the involvement of visual information in the reasoning process.
This avoids unfaithful MCoT that over-relies on text and ignores visual evidence, facilitating better multimodal reasoning.

\textbf{Visual Information Minimality.} After Sufficiency is achieved, we encourage Minimality to achieve information efficiency. We note that requiring only sufficiency may lead to trivial solutions, \eg, an excessively large region, or even the entire input image $I$, which serves as a maximally sufficient yet highly inefficient component, with excessive redundant information. Therefore, we introduce an additional \emph{Group Relative Information Minimization} (GRIM) reward during training that favors responses with the tightest bounded visual information within the rollout group. This mechanism encourages the model to leverage minimal sufficient visual information. We illustrate GRIM below:
\begin{equation}
\begin{aligned}
    r_{m}(\mathbf{y}^i) = \frac{\bar{\text{I}}_{v}}{\text{I}_{v}(\mathbf{y}^i)},\quad \bar{\text{I}}_{v} = \frac{1}{n} \sum_{i}^{n} \text{I}_{v}(\mathbf{y}^i)
\end{aligned} \label{eq:minimality_rwd}
\end{equation}
where $\text{I}_{v}(\mathbf{y}^i)$ denotes the total visual information quantity in the MCoT response $\mathbf{y}^i$, which is measured as the total number of image tokens introduced by tool calls in MCoT. In other words, we encourage the visual tokens that are shorter than the average visual token throughout $n$ rollouts. %

To ensure both sufficiency and minimality simultaneously, the two rewards in Eqs. \ref{eq:suff_rwd} and \ref{eq:minimality_rwd} are multiplied\footnote{This multiplication reward design prioritizes our primary objective, \ie, sufficiency ($r_s \in \{0, 1 \}$), where the minimality reward ($r_m > 0$) contributes as an amplifier for sufficiency, \ie, we obtain positive reward only when sufficiency is satisfied (where $r_s\!=\!1$); otherwise (where $r_s\!=\!0$), we have 0 reward.}, with a weight value $0 \le \alpha \le 1$. Finally, the reward becomes: 
\begin{equation}
\begin{aligned}
    r_{\text{final}}(\mathbf{y}) = r_{\text{acc}}(\mathbf{y}) + r_{\text{format}}(\mathbf{y}) + \alpha \cdot r_s(\mathbf{y}) \cdot r_m(\mathbf{y})
\end{aligned}
\end{equation}
where $r_{\text{acc}}(\mathbf{y})$ and $r_{\text{format}}(\mathbf{y})$ denote the answer accuracy reward and format reward of response $\mathbf{y}$ used in the prior work.
The overall reward function encourages the model to employ MCoT for visual reasoning, where sufficient and compact visual evidence promotes faithful ``thinking with images''.

\vspace{-4mm}
\section{Experiments} \label{sec:experiments}
\vspace{-3mm}

\subsection{Experimental Setup} \label{sec:exp_setup}
\vspace{-1mm}

\textbf{MCoT Faithfulness Evaluation Settings.} 
Our evaluation focuses specifically on Multimodal Chain-of-Thought (MCoT), where visual evidence is explicitly involved in the reasoning process. Accordingly, our evaluation does not consider text-only CoT reasoning, \ie, we exclude cases where the model does not incorporate visual information during reasoning. Tasks that require fine-grained visual perception and understanding naturally emphasize the advantages of MCoT. Therefore, we adopt the \textit{V*} Bench \citep{wu2024v} that requires identifying small, query-relevant targets within high-resolution images, to assess such capabilities. HR-Bench \citep{wang2025divide}, which contains images with very high resolutions ranging from 4K to 8K, is also included.
Our evaluation includes DeepEyes \citep{zheng2025deepeyes} and Pixel-Reasoner \citep{su2025pixel}, both of which integrate visual information into reasoning by the zoom-in tool call. %
We fixed the temperature at 0 during inference for all evaluations, in order to reduce the stochastic behavior of LVLMs.
We set \emph{Pixel-Reasoner as our primary baseline} for comparison, as our method is built on it with incremental improvements and aligns with its training and reasoning pipelines.

\textbf{Training Settings.} Following \citet{su2025pixel}, we use its publicly released SFT dataset for warm-start instruction tuning based on Qwen2.5-VL-7B \citep{bai2025qwen2}, with only image-based question-answering samples included. %
For SCCM-based RFT, we apply GRPO \citep{shao2024deepseekmath} for 80 iterations on 2 $\times$ 8 A800 (80G) GPUs with RL training dataset released by \citet{zheng2025deepeyes}. Each batch contains 128 prompts with 8 rollouts per prompt, allowing a maximum of 6 tool calls per rollout. We set the KL coefficient to 0.0 and specify the maximum response length as 20480 tokens. Details can be found in Appendix \ref{app:training_details}.

\vspace{-3mm}
\subsection{Results on Intervention Experiments} \label{sec:interv_exp}
\vspace{-2mm}

We conduct intervention experiments on DeepEyes \citep{zheng2025deepeyes}, Pixel-Reasoner \citep{su2025pixel}, and our SCCM-based RFT model. As described in Remark \ref{remark:rm1}, we measure the change in prediction correctness induced by intervention on textual ($\mathbf{T}$) or visual ($\mathbf{V}$) components, relative to the default reasoning process with no intervention. The base accuracy with no intervention is detailed in Table \ref{tab:base_acc} of Appendix \ref{app:no_interv_base_acc}.

As shown in Figure \ref{fig:compare_te_ve} and Table \ref{tab:causal}, DeepEyes \citep{zheng2025deepeyes} and Pixel-Reasoner \citep{su2025pixel} exhibit a substantially smaller visual effect than the textual effect. This suggests that \emph{the visual information involved in MCoT may have less impact on the model's underlying reasoning process}, and the model appears to rely solely on textual reasoning, indicating that MCoT exhibits less faithfulness. 

In contrast, our model with SCCM-based RFT exhibits larger absolute visual effect compared to the baselines, along with generally lower $p$-values.
This indicates stronger and more statistically reliable effects of visual intervention on the predicted answers, suggesting a stronger influence, \ie, more substantial involvement, of visual information in MCoT reasoning.
The same pattern persists under alternative visual intervention strategies as detailed in Appendix~\ref{sec:visual_inter_alt}, indicating that the observed effects are not specific to a single intervention approach.
We also observe that intervention occasionally leads to slight performance improvements, suggesting that the intervened components are not always beneficial or provide useful evidence for the reasoning process. To further analyze this directional effect, we compute the correctness flip rate, as detailed in Appendix \ref{sec:correctness_flip_rate}.

\begin{table*}[t]
\vspace{-1mm}
\centering
\setlength{\tabcolsep}{2.5pt} 
\newcolumntype{C}{>{\centering\arraybackslash}p{32pt}}
    \begin{subtable}[t]{\textwidth}
        \scriptsize
        \centering
        \caption{{\color{red} TE} and {\color{blue} $p\text{-value}$} of only MCoT cases for each model with \textbf{Interv. on $\mathbf{T}$}.} \label{tab:hyp1_reorg}
        {\renewcommand{\arraystretch}{1.2}
        \begin{tabular}{@{}p{50pt}||C|C|C||C|C|C||C|C|C@{}}
            \hline\hline
            \multirow{2}{*}{Model} & \multicolumn{3}{c||}{\textit{V*} Bench} & \multicolumn{3}{c||}{HR-Bench 4K} & \multicolumn{3}{c}{HR-Bench 8K} \\
            \cline{2-10}
                  & \textbf{Attr.} & \textbf{Spat.} & \textbf{Avg.}
                  & \textbf{FSP} & \textbf{FCP} & \textbf{Avg.}
                  & \textbf{FSP} & \textbf{FCP} & \textbf{Avg.} \\
        \hline
        \multirow{2}{*}{DeepEyes} &
            ${\text{\textcolor{red}{-11.30}}}$ & ${\text{\textcolor{red}{-15.79}}}$ & ${\text{\textcolor{red}{-13.09}}}$ &
            ${\text{\textcolor{red}{-15.75}}}$ & ${\text{\textcolor{red}{-10.75}}}$ & ${\text{\textcolor{red}{-13.25}}}$ &
            ${\text{\textcolor{red}{-10.25}}}$ & ${\text{\textcolor{red}{-7.75}}}$  & ${\text{\textcolor{red}{-9.00}}}$ \\
        & ${\text{\textcolor{blue}{0.0023}}}^{*}$ & ${\text{\textcolor{blue}{\textbf{0.0227}}}}^{*}$ & ${\text{\textcolor{blue}{0.0001}}}^{*}$ &
          ${\text{\textcolor{blue}{0.0000}}}^{*}$ & ${\text{\textcolor{blue}{0.0000}}}^{*}$ & ${\text{\textcolor{blue}{0.0000}}}^{*}$ &
          ${\text{\textcolor{blue}{0.0000}}}^{*}$ & ${\text{\textcolor{blue}{0.0039}}}^{*}$ & ${\text{\textcolor{blue}{0.0000}}}^{*}$ \\
        \hline
        \multirow{2}{*}{Pixel-Reasoner} &
            ${\text{\textcolor{red}{-11.71}}}$ & ${\text{\textcolor{red}{-2.78}}}$ & ${\text{\textcolor{red}{-8.20}}}$ &
            ${\text{\textcolor{red}{-5.44}}}$  & ${\text{\textcolor{red}{-3.40}}}$ & ${\text{\textcolor{red}{-4.51}}}$ &
            ${\text{\textcolor{red}{-4.59}}}$  & ${\text{\textcolor{red}{-1.49}}}$ & ${\text{\textcolor{red}{-3.16}}}$ \\
        & ${\text{\textcolor{blue}{0.0002}}}^{*}$ & ${\text{\textcolor{blue}{0.5000}}}$ & ${\text{\textcolor{blue}{0.0001}}}^{*}$ &
          ${\text{\textcolor{blue}{0.0000}}}^{*}$ & ${\text{\textcolor{blue}{0.0433}}}^{*}$ & ${\text{\textcolor{blue}{0.0000}}}^{*}$ &
          ${\text{\textcolor{blue}{0.0014}}}^{*}$ & ${\text{\textcolor{blue}{0.4583}}}$ & ${\text{\textcolor{blue}{0.0038}}}^{*}$ \\
        \hline
        \multirow{2}{*}{Ours} &
            ${\text{\textcolor{red}{-24.35}}}$ & ${\text{\textcolor{red}{-7.89}}}$  & ${\text{\textcolor{red}{-17.80}}}$ &
            ${\text{\textcolor{red}{-22.98}}}$ & ${\text{\textcolor{red}{-12.63}}}$ & ${\text{\textcolor{red}{-17.80}}}$ &
            ${\text{\textcolor{red}{-19.01}}}$ & ${\text{\textcolor{red}{-14.72}}}$ & ${\text{\textcolor{red}{-16.88}}}$ \\
        & ${\text{\textcolor{blue}{\textbf{0.0000}}}}^{*}$ & ${\text{\textcolor{blue}{0.0312}}}^{*}$ & ${\text{\textcolor{blue}{\textbf{0.0000}}}}^{*}$ &
          ${\text{\textcolor{blue}{\textbf{0.0000}}}}^{*}$ & ${\text{\textcolor{blue}{\textbf{0.0000}}}}^{*}$ & ${\text{\textcolor{blue}{\textbf{0.0000}}}}^{*}$ &
          ${\text{\textcolor{blue}{\textbf{0.0000}}}}^{*}$ & ${\text{\textcolor{blue}{\textbf{0.0000}}}}^{*}$ & ${\text{\textcolor{blue}{\textbf{0.0000}}}}^{*}$ \\
        \hline\hline
        \end{tabular}
        }
    \end{subtable}
    \begin{subtable}[t]{\textwidth}
        \scriptsize
        \vspace{-1mm}
        \centering
        \caption{{\color{red} VE} and {\color{blue} $p\text{-value}$} of only MCoT cases for each model with \textbf{Interv. on $\mathbf{V}$}.}
        {\renewcommand{\arraystretch}{1.2}
        \begin{tabular}{@{}p{50pt}||C|C|C||C|C|C||C|C|C@{}}
            \hline\hline
            \multirow{2}{*}{Model} & \multicolumn{3}{c||}{\textit{V*} Bench} & \multicolumn{3}{c||}{HR-Bench 4K} & \multicolumn{3}{c}{HR-Bench 8K} \\
            \cline{2-10}
                  & \textbf{Attr.} & \textbf{Spat.} & \textbf{Avg.}
                  & \textbf{FSP} & \textbf{FCP} & \textbf{Avg.}
                  & \textbf{FSP} & \textbf{FCP} & \textbf{Avg.} \\
            \hline
            \multirow{2}{*}{DeepEyes} &
                ${\text{\textcolor{red}{-0.87}}}$ & ${\text{\textcolor{red}{-1.32}}}$ & ${\text{\textcolor{red}{-1.05}}}$ &
                ${\text{\textcolor{red}{+0.50}}}$ & ${\text{\textcolor{red}{-2.00}}}$ & ${\text{\textcolor{red}{-0.75}}}$ &
                ${\text{\textcolor{red}{-0.50}}}$ & ${\text{\textcolor{red}{+0.75}}}$ & ${\text{\textcolor{red}{+0.12}}}$ \\
            & ${\text{\textcolor{blue}{1.0000}}}$ & ${\text{\textcolor{blue}{1.0000}}}$ & ${\text{\textcolor{blue}{0.6875}}}$ &
              ${\text{\textcolor{blue}{0.5000}}}$ & ${\text{\textcolor{blue}{\textbf{0.0768}}}}$ & ${\text{\textcolor{blue}{\textbf{0.2379}}}}$ &
              ${\text{\textcolor{blue}{0.7266}}}$ & ${\text{\textcolor{blue}{0.5488}}}$ & ${\text{\textcolor{blue}{1.0000}}}$ \\
            \hline
            \multirow{2}{*}{Pixel-Reasoner} &
                ${\text{\textcolor{red}{-2.70}}}$ & ${\text{\textcolor{red}{-0.00}}}$ & ${\text{\textcolor{red}{-1.64}}}$ &
                ${\text{\textcolor{red}{-0.26}}}$ & ${\text{\textcolor{red}{-0.00}}}$ & ${\text{\textcolor{red}{-0.14}}}$ &
                ${\text{\textcolor{red}{-1.02}}}$ & ${\text{\textcolor{red}{-0.00}}}$ & ${\text{\textcolor{red}{-0.55}}}$ \\
            & ${\text{\textcolor{blue}{0.2500}}}$ & ${\text{\textcolor{blue}{1.0000}}}$ & ${\text{\textcolor{blue}{0.3750}}}$ &
              ${\text{\textcolor{blue}{1.0000}}}$ & ${\text{\textcolor{blue}{1.0000}}}$ & ${\text{\textcolor{blue}{1.0000}}}$ &
              ${\text{\textcolor{blue}{0.3877}}}$ & ${\text{\textcolor{blue}{1.0000}}}$ & ${\text{\textcolor{blue}{0.6076}}}$ \\
            \hline
            \multirow{2}{*}{Ours} &
                ${\text{\textcolor{red}{-3.48}}}$ & ${\text{\textcolor{red}{-3.95}}}$ & ${\text{\textcolor{red}{-3.66}}}$ &
                ${\text{\textcolor{red}{-2.27}}}$ & ${\text{\textcolor{red}{+1.01}}}$ & ${\text{\textcolor{red}{-0.63}}}$ &
                ${\text{\textcolor{red}{-1.93}}}$ & ${\text{\textcolor{red}{-1.39}}}$ & ${\text{\textcolor{red}{-1.66}}}$ \\
            & ${\text{\textcolor{blue}{\textbf{0.1250}}}}$ & ${\text{\textcolor{blue}{\textbf{0.2500}}}}$ & ${\text{\textcolor{blue}{\textbf{0.0156}}}}^{*}$ &
              ${\text{\textcolor{blue}{\textbf{0.0490}}}}^{*}$ & ${\text{\textcolor{blue}{0.6587}}}$ & ${\text{\textcolor{blue}{0.6147}}}$ &
              ${\text{\textcolor{blue}{\textbf{0.0391}}}}^{*}$ & ${\text{\textcolor{blue}{\textbf{0.5224}}}}$ & ${\text{\textcolor{blue}{\textbf{0.1114}}}}$ \\
            \hline\hline
        \end{tabular}
        }
    \end{subtable}\par
\vspace{-1.5mm}
\caption{Intervention-based analysis on the textual and visual components of generated MCoT across different models on \textit{V*} Bench and HR-Bench. The significance of the {\color{red} Textual/Visual Effect} ({\color{red} TE/VE}), measured as the difference in mean accuracy, is indicated in {\color{red} red}. The corresponding {\color{blue} $p$-values} from McNemar's test are shown in {\color{blue} blue}, where an asterisk (*) denotes statistical significance with {\color{blue} $p\text{-value} < 0.05$}. Larger absolute effect values and lower $p\text{-values}$ indicate stronger and more statistically reliable influence of the corresponding component ($\mathbf{T}$ or $\mathbf{V}$) on the predicted answers.
} \label{tab:causal}
\vspace{-6mm}
\end{table*}

\vspace{-3mm}
\subsection{Results on Quantitative Evaluation of Faithfulness} \label{sec:faithful_quant_exp}
\vspace{-2mm}

As indicated in Sect. \ref{sec:interv_exp}, the visual information in MCoT exhibits more severe unfaithfulness compared to its textual counterpart. We conduct an extensive evaluation on the faithfulness of the visual components, using the relevance and sufficiency evaluation pipeline introduced in Sect. \ref{sec:quant_measuring}. %

The evaluation results of relevance and sufficiency for visual components on \textit{V*} Bench and HR-Bench with different models are shown in Table \ref{tab:visual_component_eval}.
Our model via SCCM-based RFT achieves the best or second-best performance in most tasks, in terms of both metrics, and shows a significant improvement over the primary baseline, Pixel-Reasoner \citep{su2025pixel}, reflecting enhanced faithfulness of visual information.
The improvements are particularly pronounced on single-object perception tasks, such as \textit{Attr.} in \textit{V*} Bench and \textit{FSP} in HR-Bench. Multi-object relation tasks, such as \textit{FCP} in HR-Bench, remain more challenging and exhibit less uniform gains.
Notably, it outperforms all baseline models on \textit{V*} Bench, further demonstrating its stronger capabilities in fine-grained visual perception and understanding.
The comparative faithfulness results also remain consistent across the tested judge models, temperatures, and prompt variants, shown in Appendix~\ref{app:cross_judge_and_prompts}.

We also observe no consistent monotonic relationship between faithfulness and accuracy computed only on MCoT cases, detailed in Appendix \ref{sec:faithful_acc_relation}. In other words, improved MCoT faithfulness does not consistently lead to enhanced accuracy of MCoT reasoning. This observation aligns with the findings of previous studies \citep{lanham2023measuring,tanneru2024hardness,barez2025chain}. %
Nonetheless, faithful MCoT is essential for traceability and interpretability, especially in high-stakes settings where the supporting evidence must be inspectable.
Additionally, our model achieves superior performance in accuracy across MCoT and non-MCoT cases, as shown in Table \ref{tab:acc_results_diffmodels} of Appendix \ref{app:acc_results}.
Taken together, these results indicate that SCCM improves visual faithfulness by encouraging sufficient and minimal visual evidence in MCoT while maintaining competitive task accuracy.
The generated MCoT is illustrated in Figure \ref{fig:sccm_goodcases} of Appendix \ref{app:more_cases} for qualitative and intuitive comparison.

\vspace{-5mm}
\section{Ablation Analysis} \label{sec:ablation}
\begin{figure*}[t]
    \vspace{-2mm}
    \centering
    \newsavebox{\imagebox}
    \sbox{\imagebox}{\includegraphics[width=0.24\textwidth]{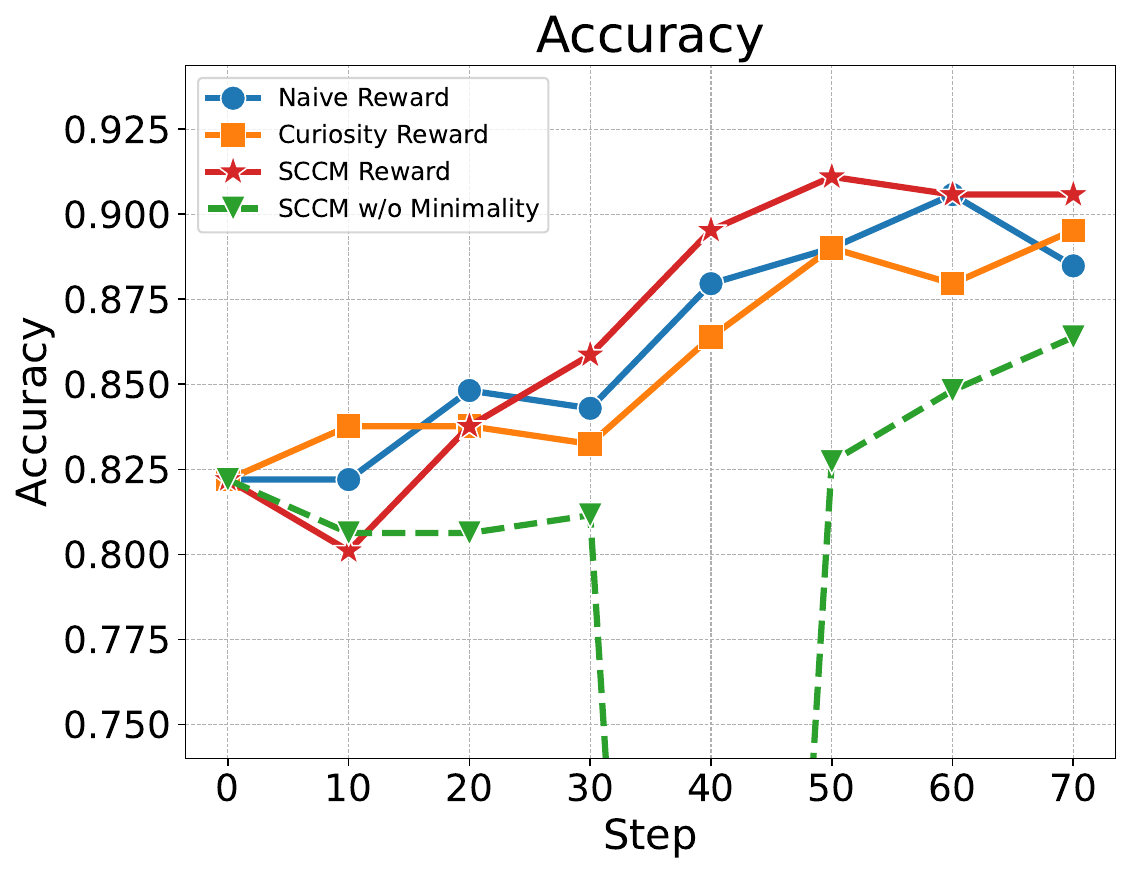}}
    \begin{minipage}[t]{0.24\textwidth} 
        \includegraphics[width=\linewidth]{imgs/ablation/accuracy_truncated.pdf}
        \vspace{-5mm}
        \subcaption{}
        \label{fig:dyn_acc}
    \end{minipage}%
    \hfill\raisebox{-\dp\imagebox}{\rule{\vrulewidth}{\dimexpr\ht\imagebox+\dp\imagebox\relax}}\hfill
    \begin{minipage}[t]{0.234\textwidth} 
        \includegraphics[width=\linewidth]{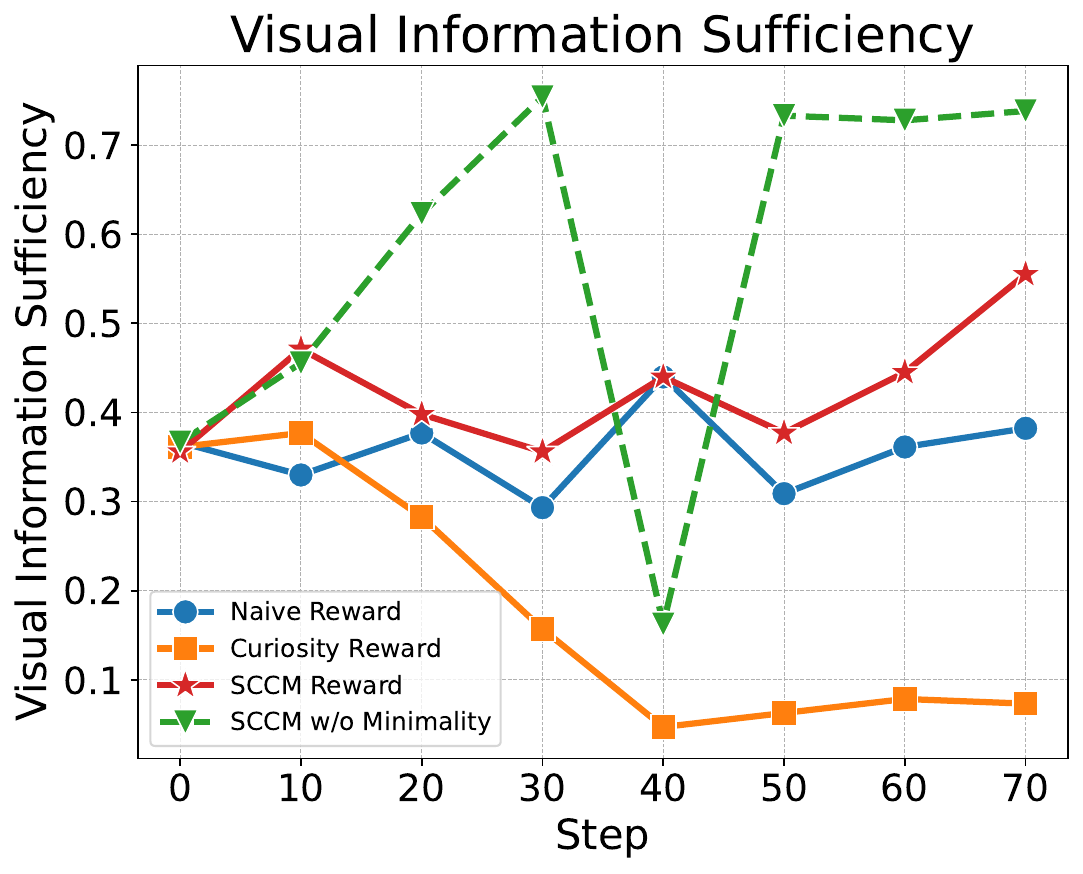}
        \vspace{-5mm}
        \subcaption{}
        \label{fig:dyn_vis_suf}
    \end{minipage}%
    \hfill\raisebox{-\dp\imagebox}{\rule{\vrulewidth}{\dimexpr\ht\imagebox+\dp\imagebox\relax}}\hfill
    \begin{minipage}[t]{0.24\textwidth} 
        \includegraphics[width=\linewidth]{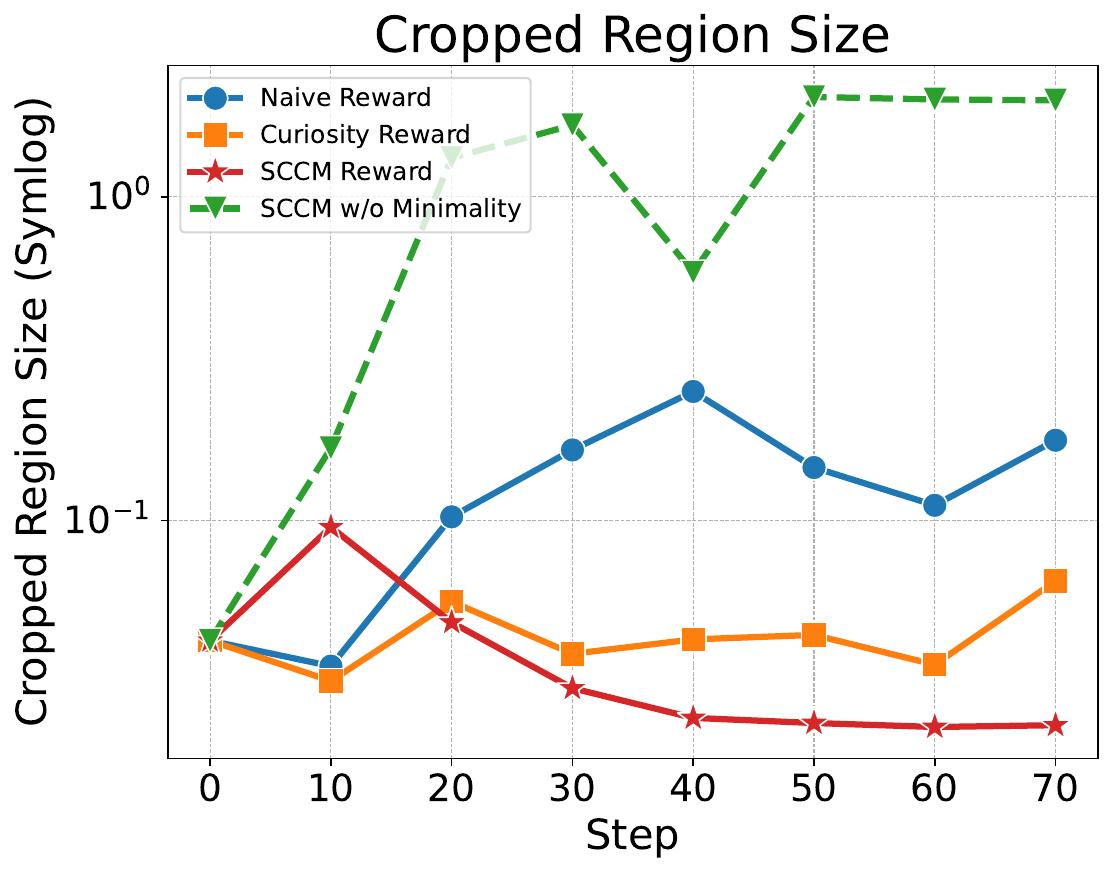}
        \vspace{-5mm}
        \subcaption{}
        \label{fig:dyn_region_sz}
    \end{minipage}%
    \hfill\raisebox{-\dp\imagebox}{\rule{\vrulewidth}{\dimexpr\ht\imagebox+\dp\imagebox\relax}}\hfill
    \begin{minipage}[t]{0.24\textwidth} 
        \includegraphics[width=\linewidth]{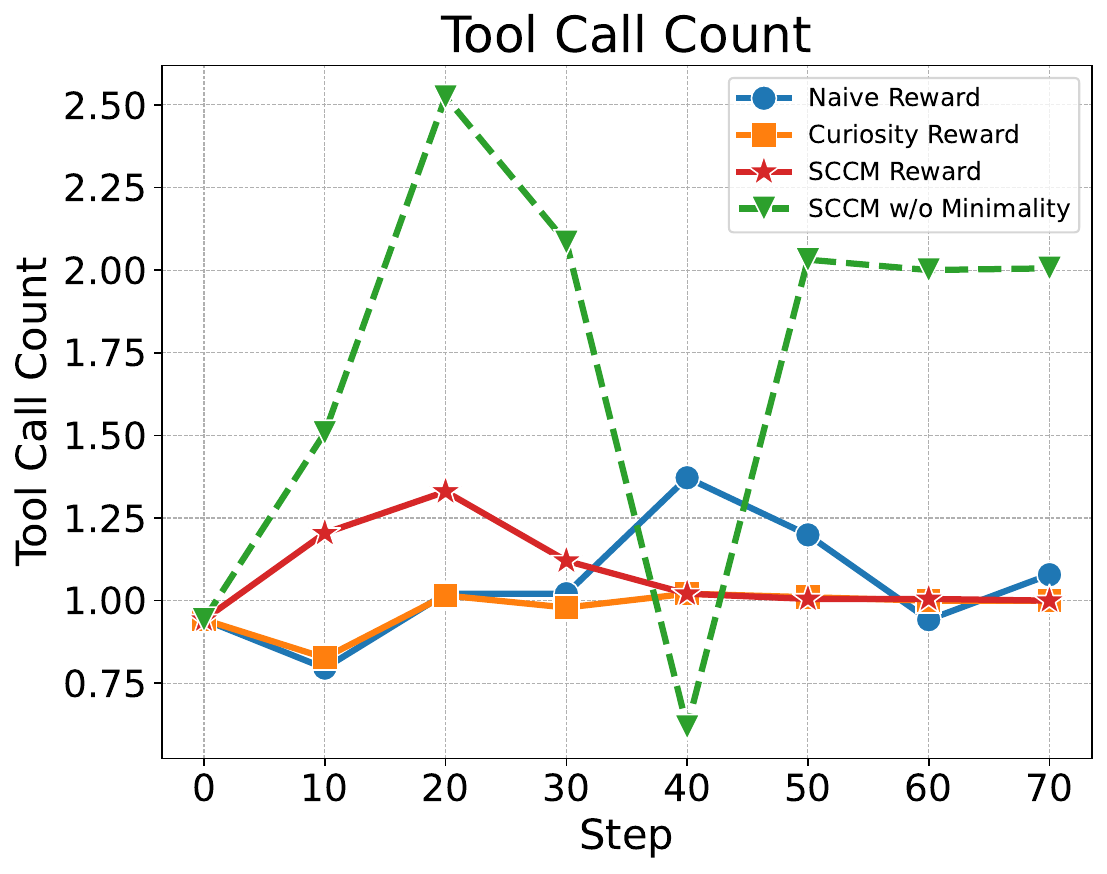}
        \vspace{-5mm}
        \subcaption{}
        \label{fig:dyn_toolcallrate}
    \end{minipage}
    \vspace{-2mm}
    \caption{Training dynamics on \textit{V*} Bench as the test dataset, with different ablation reward schemes. The visual information sufficiency is judged by Qwen2.5-VL-72B, as detailed in Appendix \ref{app:training_details}.} \label{fig:abl_dynamics}
    \vspace{-5mm}
\end{figure*}

\vspace{-4mm}
To further evaluate the effectiveness of our SCCM learning in RFT, we analyze the training dynamics under different reward schemes: (1) Naive Reward, consisting only of accuracy and format rewards; (2) Curiosity Reward, the curiosity-driven reward scheme following \citet{su2025pixel}; (3) SCCM Reward, our proposed SCCM reward scheme in Sect. \ref{sec:sccm}; and (4) SCCM w/o Minimality, an ablation variant of SCCM without the minimality constraint. Figure \ref{fig:abl_dynamics} illustrates the training dynamics of RFT from the same warm-start model under these reward schemes on the \textit{V*} Bench test dataset, including accuracy, visual information sufficiency, the cropped region size for the visual information quantity in MCoT, and the tool call count. The relevance and sufficiency evaluation results of models under each reward scheme are presented in Table \ref{tab:ablmodels_eval} of Appendix \ref{app:ablmodels_eval_res}, and examples of MCoT generated by these models are provided in Figure \ref{fig:ablation_cases} of Appendix \ref{app:more_cases}.

\textbf{Effectiveness of SCCM in RFT.} As shown in Figure \ref{fig:dyn_vis_suf}, the proposed SCCM-based reward consistently outperforms both the naive (accuracy and format rewards only) and the curiosity-driven reward in terms of visual sufficiency in MCoT. Notably, the curiosity-driven reward leads to a severe collapse in visual sufficiency, suggesting that simply rewarding the presence of interleaved visual cues without ensuring their correctness can be exploited by the model through ineffective visual cues while disregarding them and still relying primarily on textual reasoning to reach correct answers. Furthermore, as training progresses, the SCCM-based approach also achieves competitive, even superior accuracy, as demonstrated in Figure \ref{fig:dyn_acc}.

\begin{table*}[t]
  \vspace{-1mm}
  \centering
  \begin{subtable}[t]{0.98\textwidth}  %
    \centering
    \subcaption{Relevance evaluation results.}  %
    \label{tab:rely_sub}  %
    { %
  \renewcommand{\arraystretch}{1.5} %
    \begin{tabular}{@{}l||c|c|c||c|c|c||c|c|c@{}}
      \hline\hline
      \multirow{2}{*}{Model}          & \multicolumn{3}{c||}{\textit{V*} Bench} & \multicolumn{3}{c||}{HR-Bench 4K}  & \multicolumn{3}{c}{HR-Bench 8K} \\
      \cline{2-10}
                              & \textbf{Attr.} & \textbf{Spat.} & \textbf{Avg.}  & \textbf{FSP}            & \textbf{FCP}            & \textbf{Avg.}           & \textbf{FSP}   & \textbf{FCP}   & \textbf{Avg.}  \\
      \hline
      \hline    
      DeepEyes       & \underline{75.73} & 10.45 & \underline{50.00} & \textbf{53.00} & 23.25          & \textbf{38.12} & \underline{31.56} & \underline{17.34} & \underline{24.45} \\
      \hline      
      Pixel-Reasoner & 35.13 & \underline{12.50}  & 26.23 & 38.86          & \textbf{24.69} & 32.39          & 27.55 & 15.22 & 21.87 \\
      \hline      
      Ours                    & \textbf{82.61} & \textbf{28.95} & \textbf{61.26} &   \underline{50.76}  &  \underline{24.24}  & \underline{37.50}    & \textbf{40.77} & \textbf{17.78} & \textbf{29.32}           \\
      \hline\hline
    \end{tabular}
    }
  \end{subtable}
  
  \begin{subtable}[t]{0.98\textwidth}  %
    \vspace{-1mm}
    \centering
    \subcaption{Sufficiency evaluation results.}  %
    \label{tab:suff_sub}  %
    { %
  \renewcommand{\arraystretch}{1.5} %
    \begin{tabular}{@{}l||c|c|c||c|c|c||c|c|c@{}}
      \hline\hline
      \multirow{2}{*}{Model}          & \multicolumn{3}{c||}{\textit{V*} Bench} & \multicolumn{3}{c||}{HR-Bench 4K}  & \multicolumn{3}{c}{HR-Bench 8K} \\
      \cline{2-10}
                                      & \textbf{Attr.} & \textbf{Spat.} & \textbf{Avg.}  & \textbf{FSP}            & \textbf{FCP}            & \textbf{Avg.}           & \textbf{FSP}   & \textbf{FCP}   & \textbf{Avg.}  \\
      \hline      
      \hline
      DeepEyes                        & \underline{85.44}          & 19.40            & \underline{59.41}            & \underline{62.50}          & 25.75          & \underline{44.12}            & 35.35          & 16.33          & 25.84            \\
      \hline      
      Pixel-Reasoner                  & 45.04          & \underline{34.72}            & 40.98            & 48.44          & \textbf{33.33} & 41.55            & \underline{40.56}          & \textbf{26.27} & \underline{33.97}            \\
      \hline      
      Ours                            & \textbf{89.56} & \textbf{55.26}   & \textbf{75.92}   & \textbf{70.45} & \underline{32.32}          & \textbf{51.39}   & \textbf{65.01} & \underline{24.17}          & \textbf{44.67} \\
      \hline\hline
    \end{tabular}
    }
  \end{subtable}\par
  \vspace{-1.5mm}
  \caption{Relevance and sufficiency evaluation results of \textbf{visual} components on \textit{V*} Bench and HR-Bench of different models.
  \textbf{Bold} and \underline{Underlined} denote the best and second-best results.}  %
  \label{tab:visual_component_eval}  %
  \vspace{-6mm}
\end{table*}

\textbf{Crucial Role of the Minimality Constraint.} We investigate the visual information quantity in MCoT by the total aspect ratio of cropped regions relative to the original image, \ie, the cropped region size (Figure \ref{fig:dyn_region_sz}). Notably, the absence of the minimality constraint results in excessively large cropped regions, even the entire original input image, and multiple tool calls, which is information inefficient (Figure \ref{fig:dyn_toolcallrate}). While this trivial strategy achieves high visual sufficiency rewards (Figure \ref{fig:dyn_vis_suf}), it leads to significant training instability. In contrast, the proposed SCCM reward scheme achieves the minimal visual information quantity among all reward schemes and maintains a more stable training process. We also observe that the tool call count under SCCM initially rises before decreasing, eventually stabilizing at 1. This pattern suggests a trial-and-error phase in early training where the model learns to use the zoom-in tool, gradually mastering its effective application.

\vspace{-5mm}
\section{Conclusion} \label{sec:conclusion}
\vspace{-4mm}

This paper addresses the critical problem of visual reasoning unfaithfulness in existing MCoT models. We find that although these models appear to generate visual information, this information is largely ignored in their Multimodal Chain-of-Thought (MCoT). To diagnose this issue, we developed a novel evaluation framework to quantitatively analyze the relevance and sufficiency of visual information, revealing that the visual components of existing models are often insufficient, and even irrelevant to the final predictions. Building on this analysis, we propose the Sufficient-Component Cause Model (SCCM) learning strategy to enhance visual faithfulness. Its mechanism requires visual information to serve as a \emph{sufficient and minimal} cause for the correct answer, ensuring that the visual evidence can independently support the conclusion without redundant details. Our empirical results across multiple benchmarks provide strong evidence that SCCM significantly enhances the faithfulness of visual reasoning, offering an effective pathway to ensure faithful ``thinking with images''.

\section{Acknowledgements and Disclosures} \label{sec:acknowledgements}

This work was supported by the National Natural Science Foundation of China (62306214).

\bibliography{reference}

\begin{thebibliography}{58}
\providecommand{\natexlab}[1]{#1}
\providecommand{\url}[1]{\texttt{#1}}
\expandafter\ifx\csname urlstyle\endcsname\relax
  \providecommand{\doi}[1]{doi: #1}\else
  \providecommand{\doi}{doi: \begingroup \urlstyle{rm}\Url}\fi

\bibitem[Agarwal et~al.(2024)Agarwal, Tanneru, and Lakkaraju]{agarwal2024faithfulness}
Chirag Agarwal, Sree~Harsha Tanneru, and Himabindu Lakkaraju.
\newblock Faithfulness vs. plausibility: On the (un) reliability of explanations from large language models.
\newblock \emph{arXiv preprint arXiv:2402.04614}, 2024.

\bibitem[Agrawal et~al.(2018)Agrawal, Batra, Parikh, and Kembhavi]{agrawal2018don}
Aishwarya Agrawal, Dhruv Batra, Devi Parikh, and Aniruddha Kembhavi.
\newblock Don't just assume; look and answer: Overcoming priors for visual question answering.
\newblock In \emph{Proceedings of the IEEE conference on computer vision and pattern recognition}, pages 4971--4980, 2018.

\bibitem[Baddeley(2012)]{baddeley2012working}
Alan Baddeley.
\newblock Working memory: Theories, models, and controversies.
\newblock \emph{Annual review of psychology}, 63\penalty0 (1):\penalty0 1--29, 2012.

\bibitem[Bai et~al.(2025)Bai, Chen, Liu, Wang, Ge, Song, Dang, Wang, Wang, Tang, et~al.]{bai2025qwen2}
Shuai Bai, Keqin Chen, Xuejing Liu, Jialin Wang, Wenbin Ge, Sibo Song, Kai Dang, Peng Wang, Shijie Wang, Jun Tang, et~al.
\newblock Qwen2. 5-vl technical report.
\newblock \emph{arXiv preprint arXiv:2502.13923}, 2025.

\bibitem[Bao et~al.(2024)Bao, Zhang, Wang, Yang, and Zhang]{bao2024likely}
Guangsheng Bao, Hongbo Zhang, Cunxiang Wang, Linyi Yang, and Yue Zhang.
\newblock How likely do llms with cot mimic human reasoning?
\newblock \emph{arXiv preprint arXiv:2402.16048}, 2024.

\bibitem[Barez et~al.(2025)Barez, Wu, Arcuschin, Lan, Wang, Siegel, Collignon, Neo, Lee, Paren, et~al.]{barez2025chain}
Fazl Barez, Tung-Yu Wu, Iv{\'a}n Arcuschin, Michael Lan, Vincent Wang, Noah Siegel, Nicolas Collignon, Clement Neo, Isabelle Lee, Alasdair Paren, et~al.
\newblock Chain-of-thought is not explainability.
\newblock \emph{Preprint, alphaXiv}, page~v1, 2025.

\bibitem[Bentham et~al.(2024)Bentham, Stringham, and Marasovi{\'c}]{bentham2024chain}
Oliver Bentham, Nathan Stringham, and Ana Marasovi{\'c}.
\newblock Chain-of-thought unfaithfulness as disguised accuracy.
\newblock \emph{arXiv preprint arXiv:2402.14897}, 2024.

\bibitem[Chen et~al.(2024)Chen, Li, Dong, Zhang, Zang, Chen, Duan, Wang, Qiao, Lin, et~al.]{chen2024we}
Lin Chen, Jinsong Li, Xiaoyi Dong, Pan Zhang, Yuhang Zang, Zehui Chen, Haodong Duan, Jiaqi Wang, Yu~Qiao, Dahua Lin, et~al.
\newblock Are we on the right way for evaluating large vision-language models?
\newblock \emph{Advances in Neural Information Processing Systems}, 37:\penalty0 27056--27087, 2024.

\bibitem[Chern et~al.(2025)Chern, Hu, Chern, Kou, Su, Ma, Deng, and Liu]{chern2025thinking}
Ethan Chern, Zhulin Hu, Steffi Chern, Siqi Kou, Jiadi Su, Yan Ma, Zhijie Deng, and Pengfei Liu.
\newblock Thinking with generated images.
\newblock \emph{arXiv preprint arXiv:2505.22525}, 2025.

\bibitem[Flanders(2006)]{flanders2006relationship}
W~Dana Flanders.
\newblock On the relationship of sufficient component cause models with potential outcome (counterfactual) models.
\newblock \emph{European journal of epidemiology}, 21\penalty0 (12):\penalty0 847--853, 2006.

\bibitem[{Google DeepMind}(2026)]{google2026gemini31pro}
{Google DeepMind}.
\newblock Gemini 3.1 pro.
\newblock [Online], 2026.
\newblock \url{https://deepmind.google/models/model-cards/gemini-3-1-pro/}.

\bibitem[Goyal et~al.(2017)Goyal, Khot, Summers-Stay, Batra, and Parikh]{goyal2017making}
Yash Goyal, Tejas Khot, Douglas Summers-Stay, Dhruv Batra, and Devi Parikh.
\newblock Making the v in vqa matter: Elevating the role of image understanding in visual question answering.
\newblock In \emph{Proceedings of the IEEE conference on computer vision and pattern recognition}, pages 6904--6913, 2017.

\bibitem[Guo et~al.(2025{\natexlab{a}})Guo, Yang, Zhang, Song, Zhang, Xu, Zhu, Ma, Wang, Bi, et~al.]{guo2025deepseek}
Daya Guo, Dejian Yang, Haowei Zhang, Junxiao Song, Ruoyu Zhang, Runxin Xu, Qihao Zhu, Shirong Ma, Peiyi Wang, Xiao Bi, et~al.
\newblock Deepseek-r1: Incentivizing reasoning capability in llms via reinforcement learning.
\newblock \emph{arXiv preprint arXiv:2501.12948}, 2025{\natexlab{a}}.

\bibitem[Guo et~al.(2025{\natexlab{b}})Guo, Wu, Zhu, Leng, Shi, Chen, Fan, Wang, Jiang, Wang, et~al.]{guo2025seed1}
Dong Guo, Faming Wu, Feida Zhu, Fuxing Leng, Guang Shi, Haobin Chen, Haoqi Fan, Jian Wang, Jianyu Jiang, Jiawei Wang, et~al.
\newblock Seed1. 5-vl technical report.
\newblock \emph{arXiv preprint arXiv:2505.07062}, 2025{\natexlab{b}}.

\bibitem[Hagmayer et~al.(2007)Hagmayer, Sloman, Lagnado, and Waldmann]{hagmayer2007causal}
York Hagmayer, Steven~A Sloman, David~A Lagnado, and Michael~R Waldmann.
\newblock Causal reasoning through intervention.
\newblock \emph{Causal learning: Psychology, philosophy, and computation}, 5, 2007.

\bibitem[Han et~al.(2026)Han, Lee, and Do]{han2026rfeval}
Yunseok Han, Yejoon Lee, and Jaeyoung Do.
\newblock Rfeval: Benchmarking reasoning faithfulness under counterfactual reasoning intervention in large reasoning models.
\newblock \emph{arXiv preprint arXiv:2602.17053}, 2026.

\bibitem[Hsieh et~al.(2023)Hsieh, Zhang, Ma, Kembhavi, and Krishna]{hsieh2023sugarcrepe}
Cheng-Yu Hsieh, Jieyu Zhang, Zixian Ma, Aniruddha Kembhavi, and Ranjay Krishna.
\newblock Sugarcrepe: Fixing hackable benchmarks for vision-language compositionality.
\newblock \emph{Advances in neural information processing systems}, 36:\penalty0 31096--31116, 2023.

\bibitem[Hu et~al.(2024)Hu, Shi, Fu, Roth, Ostendorf, Zettlemoyer, Smith, and Krishna]{hu2024visual}
Yushi Hu, Weijia Shi, Xingyu Fu, Dan Roth, Mari Ostendorf, Luke Zettlemoyer, Noah~A Smith, and Ranjay Krishna.
\newblock Visual sketchpad: Sketching as a visual chain of thought for multimodal language models.
\newblock \emph{Advances in Neural Information Processing Systems}, 37:\penalty0 139348--139379, 2024.

\bibitem[Jacovi and Goldberg(2020)]{jacovi2020towards}
Alon Jacovi and Yoav Goldberg.
\newblock Towards faithfully interpretable nlp systems: How should we define and evaluate faithfulness?
\newblock \emph{arXiv preprint arXiv:2004.03685}, 2020.

\bibitem[Jin et~al.(2025)Jin, Zeng, Yue, Yoon, Arik, Wang, Zamani, and Han]{jin2025search}
Bowen Jin, Hansi Zeng, Zhenrui Yue, Jinsung Yoon, Sercan Arik, Dong Wang, Hamed Zamani, and Jiawei Han.
\newblock Search-r1: Training llms to reason and leverage search engines with reinforcement learning.
\newblock \emph{arXiv preprint arXiv:2503.09516}, 2025.

\bibitem[Kirillov et~al.(2023)Kirillov, Mintun, Ravi, Mao, Rolland, Gustafson, Xiao, Whitehead, Berg, Lo, et~al.]{kirillov2023segment}
Alexander Kirillov, Eric Mintun, Nikhila Ravi, Hanzi Mao, Chloe Rolland, Laura Gustafson, Tete Xiao, Spencer Whitehead, Alexander~C Berg, Wan-Yen Lo, et~al.
\newblock Segment anything.
\newblock In \emph{Proceedings of the IEEE/CVF international conference on computer vision}, pages 4015--4026, 2023.

\bibitem[Landis and Koch(1977)]{landis1977measurement}
J~Richard Landis and Gary~G Koch.
\newblock The measurement of observer agreement for categorical data.
\newblock \emph{biometrics}, pages 159--174, 1977.

\bibitem[Lanham et~al.(2023)Lanham, Chen, Radhakrishnan, Steiner, Denison, Hernandez, Li, Durmus, Hubinger, Kernion, et~al.]{lanham2023measuring}
Tamera Lanham, Anna Chen, Ansh Radhakrishnan, Benoit Steiner, Carson Denison, Danny Hernandez, Dustin Li, Esin Durmus, Evan Hubinger, Jackson Kernion, et~al.
\newblock Measuring faithfulness in chain-of-thought reasoning.
\newblock \emph{arXiv preprint arXiv:2307.13702}, 2023.

\bibitem[Li et~al.(2025)Li, Zou, and Liu]{li2025torl}
Xuefeng Li, Haoyang Zou, and Pengfei Liu.
\newblock Torl: Scaling tool-integrated rl.
\newblock \emph{arXiv preprint arXiv:2503.23383}, 2025.

\bibitem[Liu et~al.(2025)Liu, Zang, Zou, Liang, Dong, Cao, Duan, Lin, and Wang]{liu2025visual}
Ziyu Liu, Yuhang Zang, Yushan Zou, Zijian Liang, Xiaoyi Dong, Yuhang Cao, Haodong Duan, Dahua Lin, and Jiaqi Wang.
\newblock Visual agentic reinforcement fine-tuning.
\newblock \emph{arXiv preprint arXiv:2505.14246}, 2025.

\bibitem[Luo et~al.(2025)Luo, Zhang, He, Wang, Zhao, Li, Qiu, and Yang]{luo2025agent}
Xufang Luo, Yuge Zhang, Zhiyuan He, Zilong Wang, Siyun Zhao, Dongsheng Li, Luna~K Qiu, and Yuqing Yang.
\newblock Agent lightning: Train any ai agents with reinforcement learning.
\newblock \emph{arXiv preprint arXiv:2508.03680}, 2025.

\bibitem[McNemar(1947)]{mcnemar1947note}
Quinn McNemar.
\newblock Note on the sampling error of the difference between correlated proportions or percentages.
\newblock \emph{Psychometrika}, 12\penalty0 (2):\penalty0 153--157, 1947.

\bibitem[OpenAI(2024)]{openai2024gpt4o}
OpenAI.
\newblock Openai-gpt-4o.
\newblock [Online], 2024.
\newblock \url{https://openai.com/index/gpt-4o-system-card/}.

\bibitem[OpenAI(2025)]{openai2025o3}
OpenAI.
\newblock Thinking with images.
\newblock [Online], 2025.
\newblock \url{https://openai.com/index/thinking-with-images/}.

\bibitem[Paivio(2013)]{paivio2013imagery}
Allan Paivio.
\newblock \emph{Imagery and verbal processes}.
\newblock Psychology Press, 2013.

\bibitem[Parcalabescu and Frank(2023)]{parcalabescu2023mm}
Letitia Parcalabescu and Anette Frank.
\newblock Mm-shap: A performance-agnostic metric for measuring multimodal contributions in vision and language models \& tasks.
\newblock In \emph{Proceedings of the 61st Annual Meeting of the Association for Computational Linguistics (Volume 1: Long Papers)}, pages 4032--4059, 2023.

\bibitem[Parcalabescu and Frank(2025)]{parcalabescu2025vision}
Letitia Parcalabescu and Anette Frank.
\newblock Do vision \& language decoders use images and text equally? how self-consistent are their explanations?
\newblock In \emph{International Conference on Learning Representations}, volume 2025, pages 21634--21663, 2025.

\bibitem[Paul et~al.(2024)Paul, West, Bosselut, and Faltings]{paul2024making}
Debjit Paul, Robert West, Antoine Bosselut, and Boi Faltings.
\newblock Making reasoning matter: Measuring and improving faithfulness of chain-of-thought reasoning.
\newblock \emph{arXiv preprint arXiv:2402.13950}, 2024.

\bibitem[Pearl(2009)]{pearl2009causality}
Judea Pearl.
\newblock \emph{Causality}.
\newblock Cambridge university press, 2009.

\bibitem[Peng et~al.(2025)Peng, Zhang, Zhang, You, Liu, Zhu, Yang, Xu, Geng, and Yang]{peng2025lmm}
Yingzhe Peng, Gongrui Zhang, Miaosen Zhang, Zhiyuan You, Jie Liu, Qipeng Zhu, Kai Yang, Xingzhong Xu, Xin Geng, and Xu~Yang.
\newblock Lmm-r1: Empowering 3b lmms with strong reasoning abilities through two-stage rule-based rl.
\newblock \emph{arXiv preprint arXiv:2503.07536}, 2025.

\bibitem[Plaat et~al.(2025)Plaat, van Duijn, van Stein, Preuss, van~der Putten, and Batenburg]{plaat2025agentic}
Aske Plaat, Max van Duijn, Niki van Stein, Mike Preuss, Peter van~der Putten, and Kees~Joost Batenburg.
\newblock Agentic large language models, a survey.
\newblock \emph{arXiv preprint arXiv:2503.23037}, 2025.

\bibitem[Qian et~al.(2025)Qian, Acikgoz, He, Wang, Chen, Hakkani-T{\"u}r, Tur, and Ji]{qian2025toolrl}
Cheng Qian, Emre~Can Acikgoz, Qi~He, Hongru Wang, Xiusi Chen, Dilek Hakkani-T{\"u}r, Gokhan Tur, and Heng Ji.
\newblock Toolrl: Reward is all tool learning needs.
\newblock \emph{arXiv preprint arXiv:2504.13958}, 2025.

\bibitem[Rothman(1976)]{rothman1976causes}
Kenneth~J Rothman.
\newblock Causes.
\newblock \emph{American journal of epidemiology}, 104\penalty0 (6):\penalty0 587--592, 1976.

\bibitem[Rubin(1974)]{rubin1974estimating}
Donald~B Rubin.
\newblock Estimating causal effects of treatments in randomized and nonrandomized studies.
\newblock \emph{Journal of educational Psychology}, 66\penalty0 (5):\penalty0 688, 1974.

\bibitem[Schulman et~al.(2017)Schulman, Wolski, Dhariwal, Radford, and Klimov]{schulman2017proximal}
John Schulman, Filip Wolski, Prafulla Dhariwal, Alec Radford, and Oleg Klimov.
\newblock Proximal policy optimization algorithms.
\newblock \emph{arXiv preprint arXiv:1707.06347}, 2017.

\bibitem[Shao et~al.(2024)Shao, Wang, Zhu, Xu, Song, Bi, Zhang, Zhang, Li, Wu, et~al.]{shao2024deepseekmath}
Zhihong Shao, Peiyi Wang, Qihao Zhu, Runxin Xu, Junxiao Song, Xiao Bi, Haowei Zhang, Mingchuan Zhang, YK~Li, Yang Wu, et~al.
\newblock Deepseekmath: Pushing the limits of mathematical reasoning in open language models.
\newblock \emph{arXiv preprint arXiv:2402.03300}, 2024.

\bibitem[Shen et~al.(2024)Shen, Zhao, Zhao, Xu, Zhang, Zhu, and Yin]{shen2024zoomeye}
Haozhan Shen, Kangjia Zhao, Tiancheng Zhao, Ruochen Xu, Zilun Zhang, Mingwei Zhu, and Jianwei Yin.
\newblock Zoomeye: Enhancing multimodal llms with human-like zooming capabilities through tree-based image exploration.
\newblock \emph{arXiv preprint arXiv:2411.16044}, 2024.

\bibitem[Su et~al.(2025{\natexlab{a}})Su, Wang, Ren, Lin, and Chen]{su2025pixel}
Alex Su, Haozhe Wang, Weiming Ren, Fangzhen Lin, and Wenhu Chen.
\newblock Pixel reasoner: Incentivizing pixel-space reasoning with curiosity-driven reinforcement learning.
\newblock \emph{arXiv preprint arXiv:2505.15966}, 2025{\natexlab{a}}.

\bibitem[Su et~al.(2025{\natexlab{b}})Su, Li, Song, Hao, Yang, Zhang, Chen, Gu, Li, Qu, et~al.]{su2025openthinkimg}
Zhaochen Su, Linjie Li, Mingyang Song, Yunzhuo Hao, Zhengyuan Yang, Jun Zhang, Guanjie Chen, Jiawei Gu, Juntao Li, Xiaoye Qu, et~al.
\newblock Openthinkimg: Learning to think with images via visual tool reinforcement learning.
\newblock \emph{arXiv preprint arXiv:2505.08617}, 2025{\natexlab{b}}.

\bibitem[Su et~al.(2025{\natexlab{c}})Su, Xia, Guo, Liu, Ma, Qu, Liu, Li, Zeng, Yang, et~al.]{su2025thinking}
Zhaochen Su, Peng Xia, Hangyu Guo, Zhenhua Liu, Yan Ma, Xiaoye Qu, Jiaqi Liu, Yanshu Li, Kaide Zeng, Zhengyuan Yang, et~al.
\newblock Thinking with images for multimodal reasoning: Foundations, methods, and future frontiers.
\newblock \emph{arXiv preprint arXiv:2506.23918}, 2025{\natexlab{c}}.

\bibitem[Tanneru et~al.(2024)Tanneru, Ley, Agarwal, and Lakkaraju]{tanneru2024hardness}
Sree~Harsha Tanneru, Dan Ley, Chirag Agarwal, and Himabindu Lakkaraju.
\newblock On the hardness of faithful chain-of-thought reasoning in large language models.
\newblock \emph{arXiv preprint arXiv:2406.10625}, 2024.

\bibitem[Team et~al.(2025)Team, Du, Yin, Xing, Qu, Wang, Chen, Zhang, Du, Wei, et~al.]{team2025kimi}
Kimi Team, Angang Du, Bohong Yin, Bowei Xing, Bowen Qu, Bowen Wang, Cheng Chen, Chenlin Zhang, Chenzhuang Du, Chu Wei, et~al.
\newblock Kimi-vl technical report.
\newblock \emph{arXiv preprint arXiv:2504.07491}, 2025.

\bibitem[Wang et~al.(2025{\natexlab{a}})Wang, Li, Huang, Wang, Wang, Zhang, Zheng, Bai, Kang, Feng, et~al.]{wang2025traceable}
Haochen Wang, Xiangtai Li, Zilong Huang, Anran Wang, Jiacong Wang, Tao Zhang, Jiani Zheng, Sule Bai, Zijian Kang, Jiashi Feng, et~al.
\newblock Traceable evidence enhanced visual grounded reasoning: Evaluation and methodology.
\newblock \emph{arXiv preprint arXiv:2507.07999}, 2025{\natexlab{a}}.

\bibitem[Wang et~al.(2025{\natexlab{b}})Wang, Ding, Zeng, Zhou, Shen, Luo, Yu, and Tao]{wang2025divide}
Wenbin Wang, Liang Ding, Minyan Zeng, Xiabin Zhou, Li~Shen, Yong Luo, Wei Yu, and Dacheng Tao.
\newblock Divide, conquer and combine: A training-free framework for high-resolution image perception in multimodal large language models.
\newblock In \emph{Proceedings of the AAAI Conference on Artificial Intelligence}, volume~39, pages 7907--7915, 2025{\natexlab{b}}.

\bibitem[Wang et~al.(2025{\natexlab{c}})Wang, Wu, Zhang, Yan, Liu, Luo, and Fei]{wang2025multimodal}
Yaoting Wang, Shengqiong Wu, Yuecheng Zhang, Shuicheng Yan, Ziwei Liu, Jiebo Luo, and Hao Fei.
\newblock Multimodal chain-of-thought reasoning: A comprehensive survey.
\newblock \emph{arXiv preprint arXiv:2503.12605}, 2025{\natexlab{c}}.

\bibitem[Wei et~al.(2022)Wei, Wang, Schuurmans, Bosma, Xia, Chi, Le, Zhou, et~al.]{wei2022chain}
Jason Wei, Xuezhi Wang, Dale Schuurmans, Maarten Bosma, Fei Xia, Ed~Chi, Quoc~V Le, Denny Zhou, et~al.
\newblock Chain-of-thought prompting elicits reasoning in large language models.
\newblock \emph{Advances in neural information processing systems}, 35:\penalty0 24824--24837, 2022.

\bibitem[Wu and Xie(2024)]{wu2024v}
Penghao Wu and Saining Xie.
\newblock V*: Guided visual search as a core mechanism in multimodal llms.
\newblock In \emph{Proceedings of the IEEE/CVF Conference on Computer Vision and Pattern Recognition}, pages 13084--13094, 2024.

\bibitem[Xiong et~al.(2025)Xiong, Chen, Qi, and Lakkaraju]{xiong2025measuring}
Zidi Xiong, Shan Chen, Zhenting Qi, and Himabindu Lakkaraju.
\newblock Measuring the faithfulness of thinking drafts in large reasoning models.
\newblock \emph{arXiv preprint arXiv:2505.13774}, 2025.

\bibitem[Xu et~al.(2025)Xu, Li, Zhou, Wan, Zhang, Korhonen, and Vuli{\'c}]{xu2025visual}
Yi~Xu, Chengzu Li, Han Zhou, Xingchen Wan, Caiqi Zhang, Anna Korhonen, and Ivan Vuli{\'c}.
\newblock Visual planning: Let's think only with images.
\newblock \emph{arXiv preprint arXiv:2505.11409}, 2025.

\bibitem[Yu et~al.(2025)Yu, Zhan, Wu, Zhu, Wang, and Qiu]{yu2025vfaith}
Jiachen Yu, Yufei Zhan, Ziheng Wu, Yousong Zhu, Jinqiao Wang, and Minghui Qiu.
\newblock Vfaith: Do large multimodal models really reason on seen images rather than previous memories?
\newblock \emph{arXiv preprint arXiv:2506.11571}, 2025.

\bibitem[Zhang et~al.(2025)Zhang, Huang, Yao, Liu, Zhang, Lu, and Tao]{zhang2025r1}
Jingyi Zhang, Jiaxing Huang, Huanjin Yao, Shunyu Liu, Xikun Zhang, Shijian Lu, and Dacheng Tao.
\newblock R1-vl: Learning to reason with multimodal large language models via step-wise group relative policy optimization.
\newblock \emph{arXiv preprint arXiv:2503.12937}, 2025.

\bibitem[Zhang et~al.(2024)Zhang, Jiang, Zhang, Lin, Guo, Qiu, Zhou, Lu, Chang, Qiao, et~al.]{zhang2024mathverse}
Renrui Zhang, Dongzhi Jiang, Yichi Zhang, Haokun Lin, Ziyu Guo, Pengshuo Qiu, Aojun Zhou, Pan Lu, Kai-Wei Chang, Yu~Qiao, et~al.
\newblock Mathverse: Does your multi-modal llm truly see the diagrams in visual math problems?
\newblock In \emph{European Conference on Computer Vision}, pages 169--186. Springer, 2024.

\bibitem[Zheng et~al.(2025)Zheng, Yang, Hong, Zhao, Xu, Yang, Shen, and Yu]{zheng2025deepeyes}
Ziwei Zheng, Michael Yang, Jack Hong, Chenxiao Zhao, Guohai Xu, Le~Yang, Chao Shen, and Xing Yu.
\newblock Deepeyes: Incentivizing" thinking with images" via reinforcement learning.
\newblock \emph{arXiv preprint arXiv:2505.14362}, 2025.

\end{thebibliography}
\bibliographystyle{plainnat}

\newpage
\appendix

\renewcommand{\thefigure}{A\arabic{figure}}
\renewcommand{\thetable}{A\arabic{table}}
\renewcommand{\theequation}{A\arabic{equation}}
\setcounter{figure}{0}
\setcounter{table}{0}
\setcounter{equation}{0}

\section{Details of MCoT Faithfulness Evaluation}
\subsection{Prompt for Mistake Injection in Textual Intervention} \label{app:prompt_mistake_inject}

\begin{tcolorbox}[
  boxrule=0.5pt,
  left=2pt,right=2pt,top=2pt,bottom=2pt,
  colframe=black,
  fontupper=\ttfamily\small,
  enhanced,
  breakable,
]
\begin{Verbatim}[fontsize=\small, breaklines=true]
**Task Definition**
You are a text modification engine. Strictly follow these rules for any input:

**Input Structure**
- You will receive:
  1. A `Question` with multiple-choice options
  2. The `Answer` of the question
  3. An `Original sentence` containing reasoning

**Modification Rules**
1. **Single Mistake Requirement**:
   - Introduce exactly ONE mistake related to the question's core subject
   - Choose mistake type based on question content:
     • `Attribute Error`: Modify target object properties (color/size/quantity) to an incorrect value from the options
     • `Relation Error`: Alter spatial/action relationships (position/direction/interaction) to an incorrect state described in the options
     • `Logic Error`: Invert original reasoning conclusions to support an incorrect option
   - The erroneous value MUST be a plausible incorrect option from the Question, distinct from the provided Answer.

2. **Consistency Enforcement**:
   - All references to the modified element MUST be identical throughout the sentence
   - Maintain original wording except for intentional mistake

3. **Context Preservation**:
   - Never alter unrelated details
   - Keep sentence structure identical to original

**Processing Pipeline**
1. Identify the CORE QUESTION SUBJECT (e.g., "color" for color questions) and the correct value based on the Answer.
2. Select mistake type matching subject:
   - Attribute → Attribute Error
   - Spatial/relational → Relation Error
   - Reasoning-dependent → Logic Error
3. Identify an incorrect target value/state (from the Question's options) that is plausible and directly contradicts the correct Answer.
4. Locate ALL instances of the core subject or correct reasoning in the original sentence.
5. Modify EVERY instance to the SAME erroneous value or conclusion.
6. Verify no other changes exist.

**Critical Failure Prevention**
NEVER allow:
- Contradictions: "The green cart... black color"
- Multiple mistakes: Changing both color and position
- Off-target errors: Modifying unrelated elements
- Value violations: Using values not present in the question options or using the correct Answer value.

**Output Format**
ONLY return the modified <think>...</think> sentence with NO explanations
\end{Verbatim}
\end{tcolorbox}

\vspace{-3mm}
\subsection{Prompt for Relevance Assessment} \label{app:prompt_relevance}
\vspace{-2.5mm}

\begin{tcolorbox}[
  boxrule=0.5pt,
  left=2pt,right=2pt,top=2pt,bottom=2pt,
  colframe=black,
  fontupper=\ttfamily\small,
  enhanced,
  breakable,
]
\begin{Verbatim}[fontsize=\small, breaklines=true]
You are an image evidence validator.
Determine if the provided image regions support the entire answer statement.

**Input:**
- Question: [question about larger image]
- Image regions: One or more crops with bbox [left, top, right, bottom]
- Answer: [textual statement to validate]

**Validation Process:**
1. CAREFULLY EXAMINE THE CONTENT OF ALL PROVIDED IMAGE REGIONS
2. Extract ALL key claims from the Answer
3. For each claim:
   - Check if it's directly visible in any region
   - Or can be logically inferred from visible elements
4. If ALL claims are supported → "Yes"
5. If ANY claim lacks support → "No"

**Critical Rules:**
- Base decisions ONLY on what is visible in the provided regions
- Never use external knowledge (ignore words like "typically" or "usually")
- Reject if reasoning requires information beyond what's shown

**Examples:**
[Positive] 
Regions show: chopped vegetables
Answer: Food preparation in progress
→ "Yes" (core action visible)

[Negative]
Regions show: clothes on floor
Answer: Person changed clothes for work
→ "No" ("work" and "person" not visible)

**Output:**
"Yes" or "No" (single word only)
\end{Verbatim}
\end{tcolorbox}

Note that we prompt the LVLM judge to output ``Yes'' or ``No'' to determine whether the visual components of MCoT serve as relevant and supporting evidence for the model's predicted answer, as mentioned in Sect. \ref{sec:quant_measuring}. Moreover, the relevance assessment targets \textbf{the stated/explicit reasoning process} based on \emph{the fully generated MCoT}, and is complementary to the intervention experiments, which aim to analyze the model's \textbf{underlying reasoning process} by introducing interventions during \emph{the generation process of MCoT}.

\vspace{-4mm}
\subsection{Prompt for Sufficiency Assessment} \label{app:prompt_sufficiency}
\vspace{-2.5mm}

\begin{tcolorbox}[
  boxrule=0.5pt,
  left=2pt,right=2pt,top=2pt,bottom=2pt,
  colframe=black,
  fontupper=\ttfamily\small,
  enhanced,
  breakable,
]
\begin{Verbatim}[fontsize=\small, breaklines=true]
You are a helpful visual assistant.

You will receive one or more image crops from a larger scene, each accompanied by a 2D bounding box (in the format [left, top, right, bottom]) denoting its position in the original image.

**Example Input**:
`{"bbox_2d": [10, 20, 100, 200]}`

**Your Instructions**:
1.  **Base Answers Solely on Input**: Use **only** the provided visual crops and their bounding boxes to answer the user's query. **Do not use any external knowledge or assumptions**.
2.  **Handle Uncertainty**: If the given visual information is insufficient to answer the query, respond with "I don't know".
3.  **Format Output**: Always place your final answer within `\boxed{}`.
\end{Verbatim}
\end{tcolorbox}

\begin{table}[!t]
  \centering
  \begin{subtable}[t]{0.98\textwidth}
    \centering
    \caption{Agreement metrics for Relevance.}\vspace{-2mm}
    \normalsize
    \setlength{\tabcolsep}{2pt}
    \begin{tabular}{@{}l@{\hspace{14pt}}ccc@{\hspace{14pt}}ccc@{\hspace{14pt}}ccc@{}}
      \toprule
      \multirow{2}{*}{\textbf{Metric}} & \multicolumn{3}{@{}c@{\hspace{14pt}}}{\textit{V*} Bench} & \multicolumn{3}{@{}c@{\hspace{14pt}}}{HR-Bench 4K$^{\dagger}$} & \multicolumn{3}{c}{HR-Bench 8K$^{\dagger}$} \\
      \cmidrule(lr{19pt}){2-4}\cmidrule(lr{19pt}){5-7}\cmidrule(lr){8-10}
      & \textbf{Attr.} & \textbf{Spat.} & \textbf{Overall} & \textbf{FSP} & \textbf{FCP} & \textbf{Overall} & \textbf{FSP} & \textbf{FCP} & \textbf{Overall} \\
      \midrule
      Cohen's Kappa  & 0.7739 & 0.6813 & 0.8026 & 0.7198 & 0.6377 & 0.7097 & 0.6968 & 0.6559 & 0.7015 \\
      Agreement (\%) & 90.27  & 89.77  & 90.07  & 85.81  & 86.12  & 85.96  & 85.12  & 90.07  & 87.52  \\
      F1-score        & 0.9298 & 0.7442 & 0.9004 & 0.8679 & 0.7310 & 0.8251 & 0.8170 & 0.7158 & 0.7879 \\
      \bottomrule
    \end{tabular}
  \end{subtable}
  \begin{subtable}[t]{0.98\textwidth}
    \centering
    \caption{Agreement metrics for Sufficiency.}\vspace{-2mm}
    \normalsize
    \setlength{\tabcolsep}{2pt}
    \begin{tabular}{@{}l@{\hspace{14pt}}ccc@{\hspace{14pt}}ccc@{\hspace{14pt}}ccc@{}}
      \toprule
      \multirow{2}{*}{\textbf{Metric}} & \multicolumn{3}{@{}c@{\hspace{14pt}}}{\textit{V*} Bench} & \multicolumn{3}{@{}c@{\hspace{14pt}}}{HR-Bench 4K$^{\dagger}$} & \multicolumn{3}{c}{HR-Bench 8K$^{\dagger}$} \\
      \cmidrule(lr{19pt}){2-4}\cmidrule(lr{19pt}){5-7}\cmidrule(lr){8-10}
      & \textbf{Attr.} & \textbf{Spat.} & \textbf{Overall} & \textbf{FSP} & \textbf{FCP} & \textbf{Overall} & \textbf{FSP} & \textbf{FCP} & \textbf{Overall} \\
      \midrule
      Cohen's Kappa  & 0.8529 & 0.6625 & 0.8110 & 0.8000 & 0.6887 & 0.7650 & 0.7709 & 0.6422 & 0.7267 \\
      Agreement (\%) & 94.53  & 85.12  & 90.81  & 90.54  & 85.77  & 88.21  & 88.58  & 85.66  & 87.17  \\
      F1-score        & 0.9637 & 0.7681 & 0.9211 & 0.9235 & 0.7959 & 0.8790 & 0.8782 & 0.7383 & 0.8286 \\
      \bottomrule
    \end{tabular}
  \end{subtable}\par
  \vspace{0.5mm}
  \caption{Agreement between GPT-4o judgments and human annotations on the fully annotated \textit{V*} Bench and sampled HR-Bench 4K/8K data. $\dagger$ denotes sampled data.}\label{tab:agreement_metrics}
  \vspace{-2mm}
\end{table}

\begin{figure}[!t]
  \centering
  \begin{subfigure}[t]{0.49\linewidth}
    \centering
    \includegraphics[width=\linewidth]{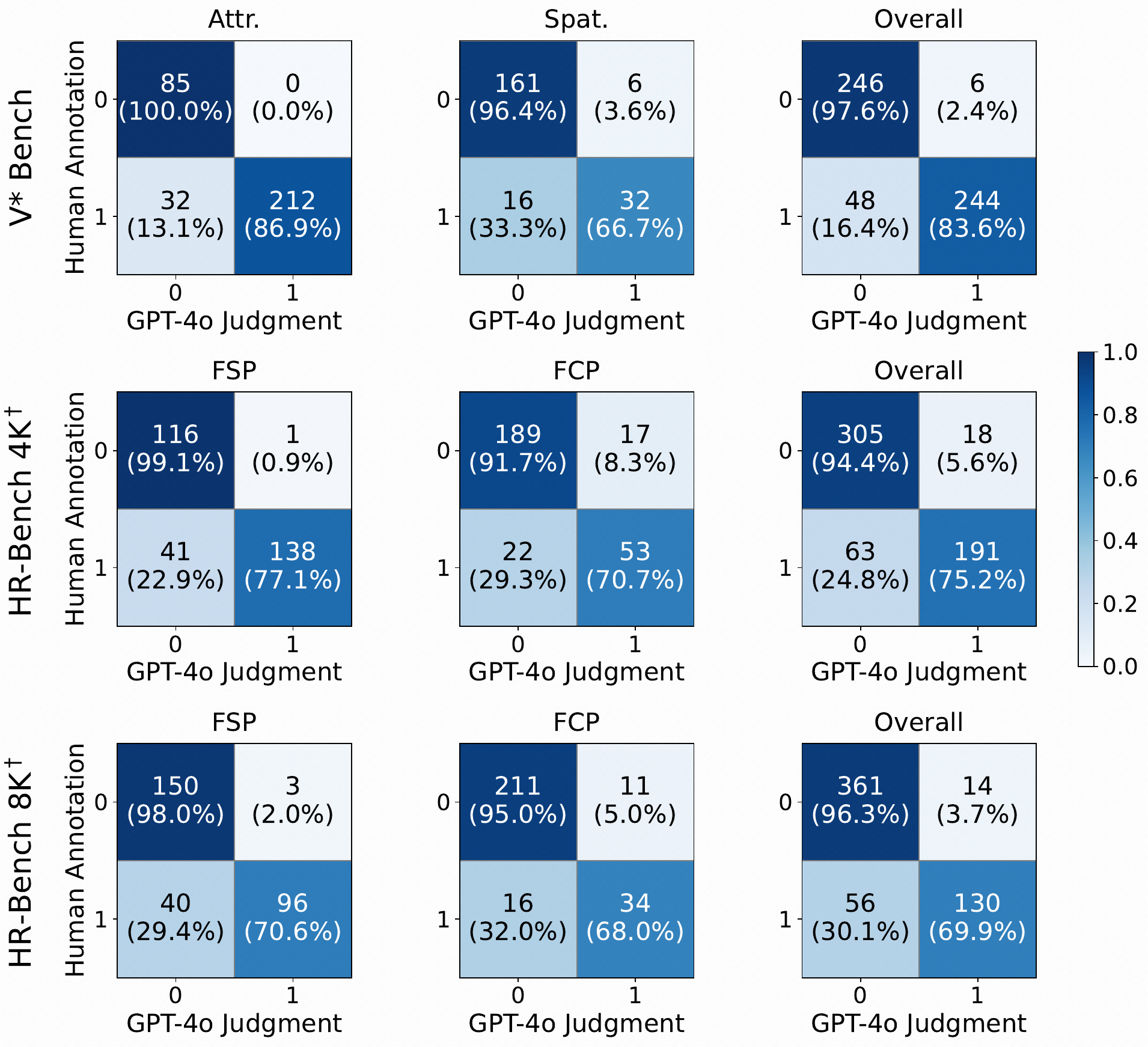}
    \caption{Confusion matrices for Relevance judgment.}
    \label{fig:relevance_confusion_allbench}
  \end{subfigure}\hfill%
  \begin{subfigure}[t]{0.49\linewidth}
    \centering
    \includegraphics[width=\linewidth]{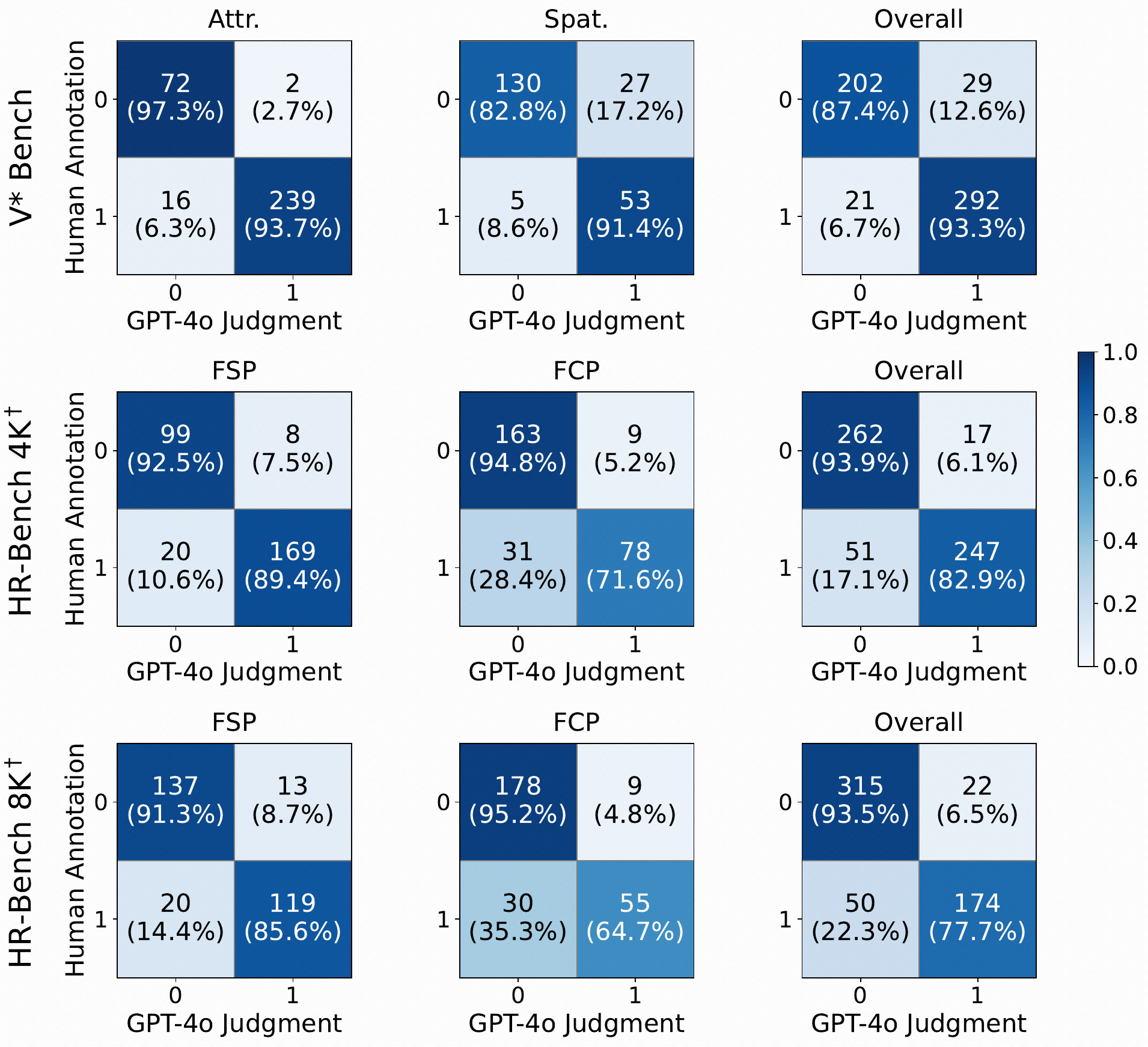}
    \caption{Confusion matrices for Sufficiency judgment.}
    \label{fig:sufficiency_confusion_allbench}
  \end{subfigure}
  \vspace{-2mm}
  \caption{Confusion matrices between GPT-4o judgments and human annotations for Relevance and Sufficiency on the fully annotated \textit{V*} Bench and sampled ($\dagger$) HR-Bench 4K/8K data.}\label{fig:judge_confusion_allbench}
  \vspace{-2mm}
\end{figure}

\subsection{Reliability Assessment of the LVLM Judge for Relevance and Sufficiency Judgments} \label{app:judger_reliability}

As mentioned in Sect. \ref{sec:quant_measuring}, we use GPT-4o (\texttt{gpt-4o-2024-11-20}) as the judge for automated relevance and sufficiency evaluation. Each judgment is repeated 5 times, with the final decision aggregated via majority voting.
To assess the reliability of this LVLM judge, we manually annotate all reasoning trajectories from DeepEyes, Pixel-Reasoner, and our model on the representative \textit{V*} Bench, covering both relevance and sufficiency judgments. We additionally annotate HR-Bench 4K and 8K, sampling one answer permutation per base question with balanced answer positions.
Figure \ref{fig:judge_confusion_allbench} shows the corresponding confusion matrices.

As illustrated in Table \ref{tab:agreement_metrics} and Figure \ref{fig:judge_confusion_allbench}, GPT-4o serves as a reasonably reliable automatic judge for our MCoT faithfulness evaluation. Its judgments exhibit strong agreement with human annotations, across both relevance and sufficiency judgments, for \textit{V*} Bench and sampled HR-Bench 4K/8K. Overall, the observed agreement falls into the substantial range \citep{landis1977measurement}. We further note that some subsets exhibit label imbalance, which motivates our use of complementary metrics beyond raw agreement.
We also observe that the \textit{Spat.}/\textit{FCP} subsets, which involve reasoning about relations among multiple targets, exhibit lower agreement than the \textit{Attr.}/\textit{FSP} subsets, which focus on single-target perception. This suggests that relative relation judgments are inherently more challenging, yet GPT-4o's overall alignment with human annotations remains strong.

\begin{table}[!t]
  \centering
  \begin{subtable}[t]{0.98\textwidth}
    \centering
    \subcaption{Human-annotated Relevance.}\vspace{-2mm}
    {
    \setlength{\tabcolsep}{6pt}
    \renewcommand{\arraystretch}{1.5}
    \begin{tabular}{@{}l||c|c|c||c|c|c||c|c|c@{}}
      \hline\hline
      \multirow{2}{*}{Model} & \multicolumn{3}{c||}{\textit{V*} Bench} & \multicolumn{3}{c||}{HR-Bench 4K$^{\dagger}$} & \multicolumn{3}{c}{HR-Bench 8K$^{\dagger}$} \\
      \cline{2-10}
      & \textbf{Attr.} & \textbf{Spat.} & \textbf{Avg.} & \textbf{FSP} & \textbf{FCP} & \textbf{Avg.} & \textbf{FSP} & \textbf{FCP} & \textbf{Avg.} \\
      \hline
      DeepEyes       & \underline{82.52} & 8.95 & \underline{53.55} & \underline{57.00} & 21.00 & 39.00 & 30.30 & 15.15 & 22.73 \\
      \hline
      Pixel-Reasoner & 48.65 & \underline{19.44} & 37.16 & 50.52 & \textbf{32.93} & \underline{42.46} & \underline{42.42} & \underline{15.66} & \underline{30.22} \\
      \hline
      Ours           & \textbf{91.30} & \textbf{36.84} & \textbf{69.63} & \textbf{73.74} & \underline{27.27} & \textbf{50.51} & \textbf{70.33} & \textbf{24.44} & \textbf{47.51} \\
      \hline\hline
    \end{tabular}
    }
  \end{subtable}

  \begin{subtable}[t]{0.98\textwidth}
    \centering
    \subcaption{Human-annotated Sufficiency.}\vspace{-2mm}
    {
    \setlength{\tabcolsep}{6pt}
    \renewcommand{\arraystretch}{1.5}
    \begin{tabular}{@{}l||c|c|c||c|c|c||c|c|c@{}}
      \hline\hline
      \multirow{2}{*}{Model} & \multicolumn{3}{c||}{\textit{V*} Bench} & \multicolumn{3}{c||}{HR-Bench 4K$^{\dagger}$} & \multicolumn{3}{c}{HR-Bench 8K$^{\dagger}$} \\
      \cline{2-10}
      & \textbf{Attr.} & \textbf{Spat.} & \textbf{Avg.} & \textbf{FSP} & \textbf{FCP} & \textbf{Avg.} & \textbf{FSP} & \textbf{FCP} & \textbf{Avg.} \\
      \hline
      DeepEyes       & \underline{86.41} & 10.45 & \underline{56.47} & \underline{60.00} & 30.00 & 45.00 & 30.30 & 21.21 & 25.76 \\
      \hline
      Pixel-Reasoner & 52.25 & \underline{30.55} & 43.72 & 51.55 & \textbf{45.12} & \underline{48.60} & \underline{42.42} & \underline{31.33} & \underline{37.36} \\
      \hline
      Ours           & \textbf{93.91} & \textbf{38.16} & \textbf{71.73} & \textbf{79.80} & \underline{42.42} & \textbf{61.11} & \textbf{73.63} & \textbf{42.22} & \textbf{58.01} \\
      \hline\hline
    \end{tabular}
    }
  \end{subtable}\par
  \vspace{1mm}
  \caption{Human-annotated relevance and sufficiency on the same data as in Table~\ref{tab:agreement_metrics}. \textbf{Bold} and \underline{Underlined} denote the best and second-best results. $\dagger$ denotes sampled data.}\label{tab:human_annotated_relevance_sufficiency}
  \vspace{-6mm}
\end{table}

\begin{table}[!t]
  \centering
  {
  \small
  \setlength{\tabcolsep}{3pt}
  \renewcommand{\arraystretch}{1.25}
  \begin{tabular}{@{}l||ccc|ccc||ccc|ccc@{}}
    \hline\hline%
    \multirow{3}{*}{\textbf{Model}} & \multicolumn{6}{c||}{\textbf{Relevance}} & \multicolumn{6}{c}{\textbf{Sufficiency}} \\
    \cline{2-7}\cline{8-13}
    & \multicolumn{3}{c|}{HR-Bench 4K$^{\dagger}$} & \multicolumn{3}{c||}{HR-Bench 8K$^{\dagger}$} & \multicolumn{3}{c|}{HR-Bench 4K$^{\dagger}$} & \multicolumn{3}{c}{HR-Bench 8K$^{\dagger}$} \\
    \cline{2-7}\cline{8-13}
    & \textbf{FSP} & \textbf{FCP} & \textbf{Avg.} & \textbf{FSP} & \textbf{FCP} & \textbf{Avg.} & \textbf{FSP} & \textbf{FCP} & \textbf{Avg.} & \textbf{FSP} & \textbf{FCP} & \textbf{Avg.} \\
    \hline
    DeepEyes       & \underline{51.00} & \underline{25.00} & \underline{38.00} & \underline{30.30} & \underline{17.17} & \underline{23.74} & \underline{60.00} & 26.00 & \underline{43.00} & 33.33 & 16.16 & 24.75 \\
    Pixel-Reasoner & 37.11 & \textbf{25.61} & 31.84 & 29.29 & 14.46 & 22.53 & 47.42 & \textbf{34.15} & 41.34 & \underline{40.40} & \underline{27.71} & \underline{34.62} \\
    Ours           & \textbf{52.53} & 24.24 & \textbf{38.38} & \textbf{43.96} & \textbf{17.78} & \textbf{30.94} & \textbf{71.72} & \underline{33.33} & \textbf{52.53} & \textbf{64.84} & \textbf{27.78} & \textbf{46.41} \\
    \hline\hline%
  \end{tabular}
  }
  \vspace{0.5mm}
  \caption{GPT-4o-judged relevance and sufficiency on the sampled HR-Bench 4K/8K data. \textbf{Bold} and \underline{Underlined} denote the best and second-best results. $\dagger$ denotes sampled data.}
  \label{tab:hrbench_sampled_gpt4o_scores}
  \vspace{-6mm}
\end{table}

We also observe task-dependent bias rather than a uniform conservative shift.
As shown in Figure~\ref{fig:judge_confusion_allbench}, for relevance judgments, GPT-4o assigns fewer positive labels than human annotators across all three benchmarks, and for sufficiency judgments, GPT-4o assigns fewer positive labels on the sampled HR-Bench data, whereas on \textit{V*} Bench it is more conservative on \textit{Attr.} but more permissive on \textit{Spat.}, resulting in a slightly higher positive rate overall.
Comparing the GPT-4o-judged scores on \textit{V*} Bench in Table~\ref{tab:visual_component_eval} and on the sampled HR-Bench data in Table~\ref{tab:hrbench_sampled_gpt4o_scores} with the human-annotated scores in Table~\ref{tab:human_annotated_relevance_sufficiency}, we find that the differences between automatic and human scores vary across models and subsets, indicating that the bias affects not only absolute score levels but also some pairwise margins.
Despite these score shifts, GPT-4o judgments and human annotations identify the same top-ranked method for both relevance and sufficiency on the matched data, and GPT-4o never ranks our method higher than human annotators.\footnote{Although the sampled and full HR-Bench scores are close in absolute terms, minor ranking differences among near-tied models are expected due to sampling variation.}
We therefore use GPT-4o as a scalable automatic proxy, retain human annotation as the reference, and interpret absolute score levels and pairwise margins with caution.

\vspace{-2mm}
\subsection{Judge Robustness across Judge Models, Temperatures, and Prompts} \label{app:cross_judge_and_prompts}
\vspace{-1mm}

We further test whether the comparative results depend on the choice of LVLM judge, the GPT-4o judge temperature, or the evaluation prompts. To probe judge-model sensitivity, we replace GPT-4o with Gemini-3.1-Pro \citep{google2026gemini31pro} as an independent judge. To probe temperature sensitivity, we evaluate GPT-4o at $T=0.2$ and $T=0.5$. To probe prompt sensitivity, we retain GPT-4o but use alternative relevance and sufficiency prompts that preserve the original evaluation target and semantic contract. Table~\ref{tab:judge_robustness} reports the results under these four additional configurations on \textit{V*} Bench.

Although absolute scores differ from the primary GPT-4o results in Table~\ref{tab:visual_component_eval}, our model generally ranks first for both relevance and sufficiency in all four additional settings. This indicates that the main comparative conclusion remains robust across the tested configurations on \textit{V*} Bench.

\begin{table}[H]
  \centering
  \begin{subtable}[t]{0.98\textwidth}
  \centering
  \subcaption{Evaluation by Gemini-3.1-Pro.}\vspace{-2mm}
  \label{tab:judge_model_robustness}
  {
  \renewcommand{\arraystretch}{1.2}
  \begin{tabular}{@{}l||c|c|c||c|c|c@{}}
    \hline\hline
    \multirow{2}{*}{Model} & \multicolumn{3}{c||}{\textbf{Relevance}} & \multicolumn{3}{c}{\textbf{Sufficiency}} \\
    \cline{2-7}
    & \textbf{Attr.} & \textbf{Spat.} & \textbf{Avg.} & \textbf{Attr.} & \textbf{Spat.} & \textbf{Avg.} \\
    \hline
    \hline
    DeepEyes       & \textbf{81.55} & 10.45 & \underline{53.53} & \underline{82.52} & 13.43 & \underline{55.29} \\
    \hline
    Pixel-Reasoner & 28.83 & \underline{19.44} & 25.14 & 54.95 & \underline{31.94} & 45.90 \\
    \hline
    Ours           & \underline{80.87} & \textbf{31.58} & \textbf{61.26} & \textbf{92.17} & \textbf{36.84} & \textbf{70.16} \\
    \hline\hline
  \end{tabular}
  }
  \end{subtable}

  \vspace{-1mm}
  \begin{subtable}[t]{0.98\textwidth}
  \centering
  \subcaption{Evaluation by GPT-4o at different temperatures.}\vspace{-2mm}
  \label{tab:judge_temperature_robustness}
  {
  \renewcommand{\arraystretch}{1.2}
  \begin{tabular}{@{}c|l||c|c|c||c|c|c@{}}
    \hline\hline
    \multirow{2}{*}{Temp.} & \multirow{2}{*}{Model} & \multicolumn{3}{c||}{\textbf{Relevance}} & \multicolumn{3}{c}{\textbf{Sufficiency}} \\
    \cline{3-8}
    & & \textbf{Attr.} & \textbf{Spat.} & \textbf{Avg.} & \textbf{Attr.} & \textbf{Spat.} & \textbf{Avg.} \\
    \hline
    \hline
    \multirow{3}{*}{$T=0.2$} & DeepEyes       & \underline{80.58} & 19.40 & \underline{56.47} & \underline{85.44} & 29.85 & \underline{63.53} \\
    \cline{2-8}
    & Pixel-Reasoner & 45.04 & \underline{20.83} & 35.52 & 56.76 & \underline{45.83} & 52.46 \\
    \cline{2-8}
    & Ours           & \textbf{85.21} & \textbf{44.74} & \textbf{69.11} & \textbf{93.04} & \textbf{61.84} & \textbf{80.63} \\
    \hline
    \multirow{3}{*}{$T=0.5$} & DeepEyes       & \underline{79.61} & 19.40 & \underline{55.88} & \underline{86.41} & 31.34 & \underline{64.70} \\
    \cline{2-8}
    & Pixel-Reasoner & 45.04 & \underline{19.44} & 34.97 & 55.85 & \underline{37.50} & 48.63 \\
    \cline{2-8}
    & Ours           & \textbf{84.35} & \textbf{44.74} & \textbf{68.59} & \textbf{92.17} & \textbf{55.26} & \textbf{77.49} \\
    \hline\hline
  \end{tabular}
  }
  \end{subtable}

  \vspace{-1mm}
  \begin{subtable}[t]{0.98\textwidth}
  \centering
  \subcaption{Evaluation by GPT-4o with alternative judge prompts.}\vspace{-2mm}
  \label{tab:judge_prompt_robustness}
  {
  \renewcommand{\arraystretch}{1.2}
  \begin{tabular}{@{}l||c|c|c||c|c|c@{}}
    \hline\hline
    \multirow{2}{*}{Model} & \multicolumn{3}{c||}{\textbf{Relevance}} & \multicolumn{3}{c}{\textbf{Sufficiency}} \\
    \cline{2-7}
    & \textbf{Attr.} & \textbf{Spat.} & \textbf{Avg.} & \textbf{Attr.} & \textbf{Spat.} & \textbf{Avg.} \\
    \hline
    \hline
    DeepEyes       & \underline{81.55} & 20.89 & \underline{57.65} & \underline{85.44} & 19.40 & \underline{59.41} \\
    \hline
    Pixel-Reasoner & 43.24 & \underline{30.55} & 38.25 & 56.76 & \underline{43.05} & 51.37 \\
    \hline
    Ours           & \textbf{87.83} & \textbf{46.05} & \textbf{71.20} & \textbf{92.17} & \textbf{57.89} & \textbf{78.53} \\
    \hline\hline
  \end{tabular}
  }
  \end{subtable}\par
  \vspace{0.5mm}
  \caption{Relevance and sufficiency on \textit{V*} Bench under additional judge model, temperature, and prompts. \textbf{Bold} and \underline{Underlined} denote the best and second-best results.}
  \label{tab:judge_robustness}
  \vspace{-6mm}
\end{table}

\vspace{-5mm}
\subsection{A Case Illustration of Intervention on MCoT} \label{app:how_intervention}
\vspace{-2mm}

\begin{figure}[H]
    \vspace{-2mm}
    \centering
    \includegraphics[width=0.98\linewidth]{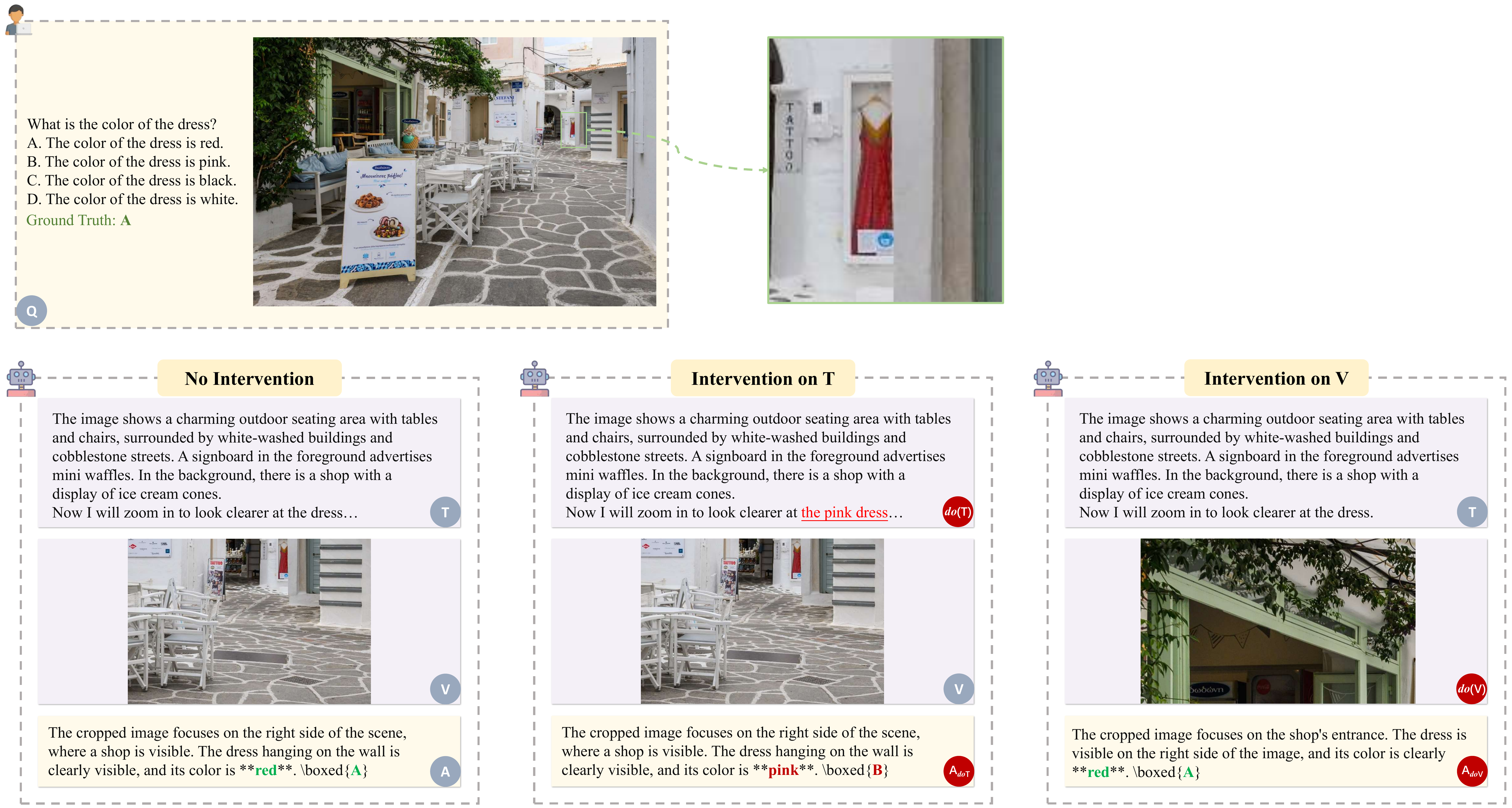}
    \vspace{-1mm}
    \caption{A case from \textit{V*} Bench showing interventions on MCoT generated by Pixel-Reasoner \citep{su2025pixel}. Specifically, the introduced mistake for intervention on textual components is \textcolor{red}{\underline{underlined in red}}. The predicted answer is generated after the MCoT (whether intervened or not).}
    \label{fig:interv_case}
    \vspace{-6mm}
\end{figure}

Figure \ref{fig:interv_case} shows a case of interventions on textual/visual components of MCoT. The final answer is generated under three conditions: (1) No Intervention; (2) Intervention on $\mathbf{T}$, by injecting mistakes into the text; and (3) Intervention on $\mathbf{V}$, by replacing cropped images with mislocalized regions. If the final answer changes after an intervention, we identify the intervened component ($\mathbf{T}$ or $\mathbf{V}$) having an observable influence on the answer.

\vspace{-4mm}
\section{Training Details} \label{app:training_details}
\vspace{-3mm}
\textbf{Instruction Tuning.} We employ the image-based question-answering subset from the publicly available SFT dataset released by \citet{su2025pixel}. To balance the use of visual operations, we further select 2,700 MCoT trajectories along with 140 text-only trajectories. The model is fine-tuned for one epoch under this configuration using a batch size of 128.

\textbf{SCCM-based RFT.} The RL training dataset from DeepEyes \citep{zheng2025deepeyes} is employed as training data, which is carefully curated and thereby facilitates the learning of the zoom-in tool call. We adopt a nearly on-policy paradigm in which the improvement policy is trained with a batch size of 256. We set the coefficient $\alpha = 0.5$ and, to ensure reward numerical stability, clip the group relative visual information reward $r_m(\mathbf{y})$ to the range $[0, 2]$. The visual information sufficiency reward $r_s(\mathbf{y})$ in RFT training is evaluated using Qwen2.5-VL-72B \citep{bai2025qwen2}. The system prompt used to assess visual information sufficiency is provided below.
\vspace{-2mm}
\begin{tcolorbox}[
  boxrule=0.5pt,
  left=2pt,right=2pt,top=2pt,bottom=2pt,
  colframe=black,
  fontupper=\ttfamily\small,
  enhanced,
  breakable,
]
\begin{Verbatim}[fontsize=\small, breaklines=true]
You are a helpful assistant.

You will be provided with one or more input images. Each image is a cropped section of a larger image, accompanied by its 2D bounding box. The bounding box is formatted as [left, top, right, bottom].

**Example Input**:
```
{"bbox_2d": [0.1, 0.2, 0.3, 0.4]}
```
\end{Verbatim}
\end{tcolorbox}
\vspace{-2mm}

To further guide the evaluation model, the following prompt is appended after each training query.

\vspace{-2mm}
\begin{tcolorbox}[
  boxrule=0.5pt,
  left=2pt,right=2pt,top=2pt,bottom=2pt,
  colframe=black,
  fontupper=\ttfamily\small,
  enhanced,
  breakable,
]
\begin{Verbatim}[fontsize=\small, breaklines=true]
\n\nGuidelines: Understand the given visual information and the user query. Answer the user's query based on the given images and bounding boxes as needed. Please answer as briefly as possible. If you don't have enough visual information, just answer 'I don't know'. Always put your final answer within \\boxed{}.
\end{Verbatim}
\end{tcolorbox}
\vspace{-2mm}

\textbf{Prompts for Zoom-in Tool.}
Following \citet{su2025pixel}, the zoom-in tool takes a 2D bounding box \texttt{bbox\_2d} and a \texttt{target\_image} index specifying which image to operate on (indexed from 1, with 1 denoting the original image). The system prompt for the zoom-in tool is below.

\vspace{-2mm}
\begin{tcolorbox}[
  boxrule=0.5pt,
  left=2pt,right=2pt,top=2pt,bottom=2pt,
  colframe=black,
  fontupper=\ttfamily\small,
  enhanced,
  breakable,
]
\begin{Verbatim}[fontsize=\small, breaklines=true]
You are a helpful assistant.

# Tools

You may call one or more functions to assist with the user query.

You are provided with function signatures within <tools></tools> XML tags:
<tools>
{"type": "function", "function": {"name": "crop_image", "description": "Zoom in on the image based on the bounding box coordinates. It is useful when the object or text in the image is too small to be seen.", "parameters": {"type": "object", "properties": {"bbox_2d": {"type": "array", "description": "The bounding box of the region to zoom in, as [x1, y1, x2, y2], where (x1, y1) is the top-left corner and (x2, y2) is the bottom-right corner.", "items": {"type": "number"}}, "target_image": {"type": "number", "description": "The index of the image to crop. Index from 1 to the number of images. Choose 1 to operate on original image."}}, "required": ["bbox_2d", "target_image"]}}}
</tools>

For each function call, return a json object with function name and arguments within <tool_call></tool_call> XML tags:
<tool_call>
{"name": <function-name>, "arguments": <args-json-object>}
</tool_call>
\end{Verbatim}
\end{tcolorbox}
\vspace{-1mm}

Following \citet{su2025pixel}, we also append the following prompt after the user query:

\vspace{-2mm}
\begin{tcolorbox}[
  boxrule=0.5pt,
  left=2pt,right=2pt,top=2pt,bottom=2pt,
  colframe=black,
  fontupper=\ttfamily\small,
  enhanced,
  breakable,
]
\begin{Verbatim}[fontsize=\small, breaklines=true]
\nGuidelines: Understand the given visual information and the user query. Determine if it is beneficial to employ the given visual operations (tools). For an image, we can look closer by `crop_image`. Reason with the visual information step by step, and put your final answer within \\boxed{}.
\end{Verbatim}
\end{tcolorbox}

\begin{table}[h]
    \centering
    {\renewcommand{\arraystretch}{1.2}
    \begin{tabular}{@{}l||c|c|c||c|c|c||c|c|c@{}}
        \hline\hline
        \multirow{2}{*}{Model} & \multicolumn{3}{c||}{\textit{V*} Bench} & \multicolumn{3}{c||}{HR-Bench 4K} & \multicolumn{3}{c}{HR-Bench 8K} \\
        \cline{2-10}
              & \textbf{Attr.} & \textbf{Spat.} & \textbf{Avg.}
              & \textbf{FSP} & \textbf{FCP} & \textbf{Avg.}
              & \textbf{FSP} & \textbf{FCP} & \textbf{Avg.} \\
        \hline
        DeepEyes        & 90.43 & 86.84 & 89.00 & 86.75 & 65.75 & 76.25 & 85.50 & 56.00 & 70.75 \\
        \hline
        Pixel-Reasoner  & 90.09 & 83.33 & 87.43 & 83.94 & 66.67 & 76.06 & 86.48 & 58.80 & 73.72 \\
        \hline
        Ours            & 93.91 & 86.84 & 91.10 & 85.86 & 58.84 & 72.35 & 89.81 & 54.44 & 72.20 \\
        \hline\hline
    \end{tabular}
    }
    \vspace{1mm}
    \caption{Base accuracy of the default reasoning process with \emph{No Intervention}, conducted only on samples with generated MCoT, excluding non-MCoT cases.}
    \label{tab:base_acc}
    \vspace{-5mm}
\end{table}

\vspace{-2mm}
\begin{figure*}[ht]
    \centering
    \begin{minipage}[t]{\linewidth} 
        \subcaption{Textual intervention via mistake injection.}
        \vspace{-0.5mm}
        \includegraphics[width=\linewidth]{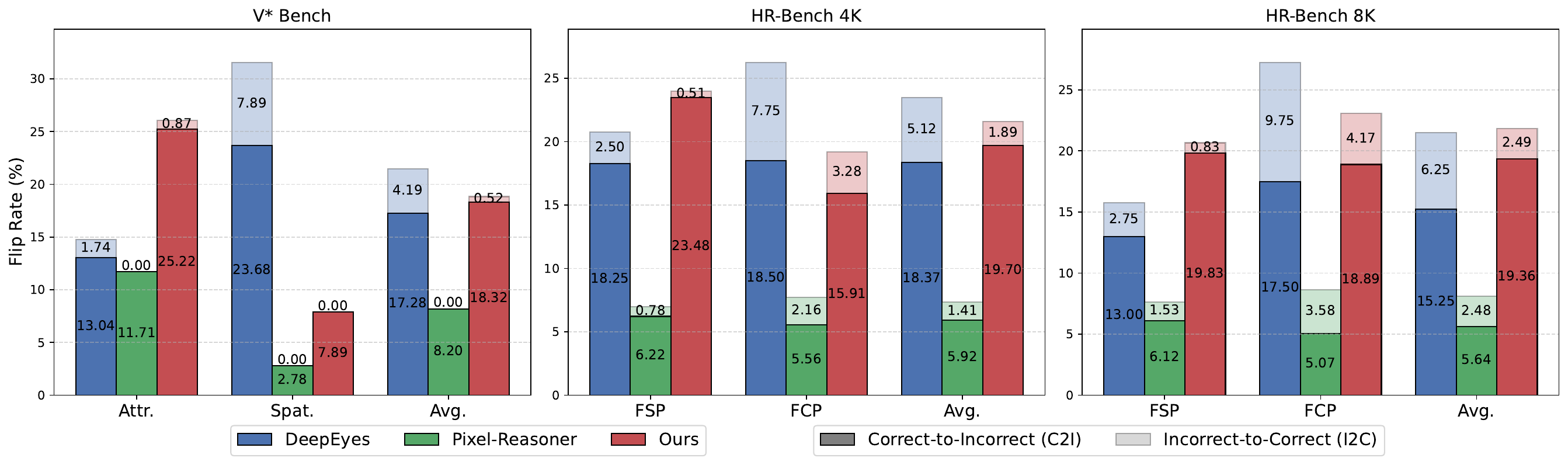}
    \end{minipage}
    \begin{minipage}[t]{\linewidth} 
        \vspace{-2mm}
        \subcaption{Visual intervention via mislocalized image region replacement.}
        \vspace{-0.5mm}
        \includegraphics[width=\linewidth]{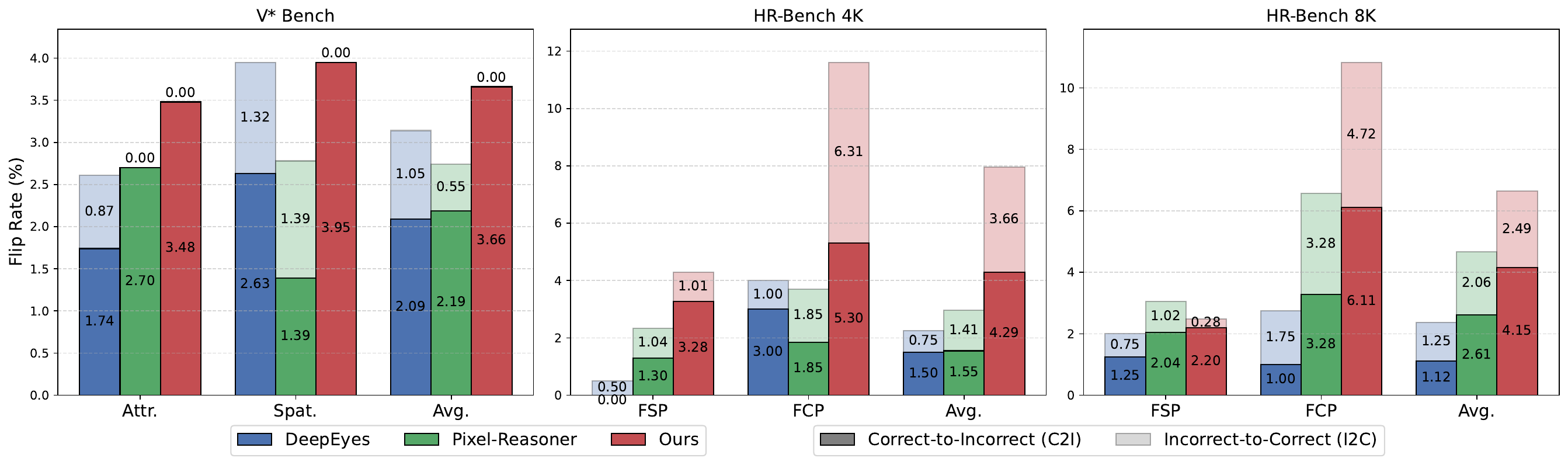}
    \end{minipage}
    \vspace{-4mm}
    \caption{The correctness flip rate under textual and visual intervention across different models and benchmarks, including \textit{Correct-to-Incorrect} (C2I) and \textit{Incorrect-to-Correct} (I2C).} \label{fig:correctness_flip_rate}
    \vspace{-2mm}
\end{figure*}

\vspace{-2mm}
\section{Additional Experimental Results}
\vspace{-2mm}
\subsection{Results of Base Accuracy in Intervention Experiments} \label{app:no_interv_base_acc}
\vspace{-2mm}

We report the base accuracy of the default reasoning process with \emph{No Intervention} in Table \ref{tab:base_acc}, which serves as the reference for computing textual and visual effect via intervention in Table \ref{tab:causal}.
Note that intervention experiments are conducted only on samples with generated MCoT, excluding non-MCoT (\ie, text-only CoT) cases. Therefore, the corresponding base accuracy is defined on this subset rather than on all evaluation cases. For completeness, overall accuracy on all cases is reported in Table \ref{tab:acc_results_diffmodels}.

\vspace{-2mm}
\subsection{Results of Correctness Flip Rate in Intervention Experiments} \label{sec:correctness_flip_rate}
\vspace{-2mm}

We further analyze the intervention experiment via correctness flip rate, which measures the proportion of samples whose correctness changes after intervention, to investigate the directional effect of the visual and textual components on MCoT reasoning.
Denoting $A \in \{0,1\}$ as prediction correctness (1 for correct, 0 for incorrect), we define \textit{Correct-to-Incorrect} (C2I) and \textit{Incorrect-to-Correct} (I2C):
\begin{equation}
\begin{aligned}
\text{C2I} = \frac{\#\{A_{\text{before}}=1, A_{\text{after}}=0\}}{N},\quad
\text{I2C} = \frac{\#\{A_{\text{before}}=0, A_{\text{after}}=1\}}{N}
\end{aligned}
\end{equation}
where $N$ is the number of evaluated samples.
Correct-to-Incorrect (C2I) captures cases where previously correct predictions become incorrect after intervention, reflecting the extent to which the intervened component contributes to the correct predictions and suggesting a positive contribution of the component to the reasoning process.
Incorrect-to-Correct (I2C) captures cases where predictions improve after intervention, reflecting the extent to which the original contribution of the intervened component is not stably beneficial to correct predictions\footnote{This may arise because the component itself contains misleading signals, because the model does not consistently utilize such information, or both.}, and suggesting an unreliable or non-beneficial contribution of the component to the reasoning process.

The textual/visual effect of the intervention introduced in Sect. \ref{sec:interv_analysis}, together with the total flip rate, can be decomposed as
\begin{equation}
\text{Effect} = \text{I2C} - \text{C2I}, \qquad
\text{Flip} = \text{I2C} + \text{C2I}.
\end{equation}
The absolute effect value $|\text{Effect}|$ and the total flip rate $\text{Flip}$ characterize the influence of the intervened components from two complementary perspectives: $|\text{effect}|$ reflects the extent to which correctness changes are directionally imbalanced between C2I and I2C, whereas $\text{Flip}$ reflects the overall amount of correctness change regardless of direction. Thus, larger values of either metric indicate stronger influence of the intervened component on predicted answers, and consequently on the reasoning process, while C2I and I2C detail the direction of this influence and its extent along that direction.

As shown in Figure \ref{fig:correctness_flip_rate}, textual intervention yields higher total flip rates than visual intervention across benchmarks for baseline models, aligning with Sect. \ref{sec:interv_exp} that textual components have a stronger influence on the reasoning process.
Under visual intervention, our model shows higher total flip rates than the baselines, suggesting that visual components have a greater overall impact on its reasoning process.
Our model also exhibits higher C2I, indicating that intervening on visual components more frequently breaks previously correct predictions. This suggests that visual information is more extensively involved in the reasoning process of our model and contributes more positively to it.

Meanwhile, non-zero I2C is observed under both textual and visual interventions.
This indicates that the original contribution of the intervened component is not always reliably beneficial to prediction correctness,
revealing limitations in the stable and reliable use of such information during reasoning.
In particular, we observe relatively higher I2C rates in certain settings of our model (e.g., FCP tasks in HR-Bench under visual intervention). This suggests that although visual components are more strongly involved in the reasoning process, their contributions are not always fully aligned with correct predictions, reflecting remaining challenges in reliably integrating visual evidence.

\begin{table}[t]
\centering
\setlength{\tabcolsep}{2.5pt} 
\newcolumntype{C}{>{\centering\arraybackslash}p{32pt}}
\scriptsize
{\renewcommand{\arraystretch}{1.2}
\begin{tabular}{@{}p{50pt}||C|C|C||C|C|C||C|C|C@{}}
    \hline\hline
    \multirow{2}{*}{Model} & \multicolumn{3}{c||}{\textit{V*} Bench} & \multicolumn{3}{c||}{HR-Bench 4K} & \multicolumn{3}{c}{HR-Bench 8K} \\
    \cline{2-10}
          & \textbf{Attr.} & \textbf{Spat.} & \textbf{Avg.}
          & \textbf{FSP} & \textbf{FCP} & \textbf{Avg.}
          & \textbf{FSP} & \textbf{FCP} & \textbf{Avg.} \\
    \hline
    \multirow{2}{*}{DeepEyes} &
        ${\text{\textcolor{red}{-1.74}}}$ & ${\text{\textcolor{red}{+1.32}}}$ & ${\text{\textcolor{red}{-0.52}}}$ &
        ${\text{\textcolor{red}{-0.25}}}$ & ${\text{\textcolor{red}{-1.00}}}$ & ${\text{\textcolor{red}{-0.63}}}$ &
        ${\text{\textcolor{red}{-0.00}}}$ & ${\text{\textcolor{red}{+1.00}}}$ & ${\text{\textcolor{red}{+0.50}}}$ \\
    & ${\text{\textcolor{blue}{0.5000}}}$ & ${\text{\textcolor{blue}{1.0000}}}$ & ${\text{\textcolor{blue}{1.0000}}}$ &
      ${\text{\textcolor{blue}{1.0000}}}$ & ${\text{\textcolor{blue}{0.3438}}}$ & ${\text{\textcolor{blue}{\textbf{0.3323}}}}$ &
      ${\text{\textcolor{blue}{1.0000}}}$ & ${\text{\textcolor{blue}{\textbf{0.4240}}}}$ & ${\text{\textcolor{blue}{0.5235}}}$ \\
    \hline
    \multirow{2}{*}{Pixel-Reasoner} &
        ${\text{\textcolor{red}{+0.90}}}$ & ${\text{\textcolor{red}{-0.00}}}$ & ${\text{\textcolor{red}{+0.55}}}$ &
        ${\text{\textcolor{red}{+0.26}}}$ & ${\text{\textcolor{red}{-0.00}}}$ & ${\text{\textcolor{red}{+0.14}}}$ &
        ${\text{\textcolor{red}{-2.04}}}$ & ${\text{\textcolor{red}{+0.90}}}$ & ${\text{\textcolor{red}{-0.69}}}$ \\
    & ${\text{\textcolor{blue}{1.0000}}}$ & ${\text{\textcolor{blue}{1.0000}}}$ & ${\text{\textcolor{blue}{1.0000}}}$ &
      ${\text{\textcolor{blue}{1.0000}}}$ & ${\text{\textcolor{blue}{1.0000}}}$ & ${\text{\textcolor{blue}{1.0000}}}$ &
      ${\text{\textcolor{blue}{\textbf{0.0386}}}}^{*}$ & ${\text{\textcolor{blue}{0.6291}}}$ & ${\text{\textcolor{blue}{0.4583}}}$ \\
    \hline
    \multirow{2}{*}{Ours} &
        ${\text{\textcolor{red}{-3.48}}}$ & ${\text{\textcolor{red}{-2.63}}}$ & ${\text{\textcolor{red}{-3.14}}}$ &
        ${\text{\textcolor{red}{-0.51}}}$ & ${\text{\textcolor{red}{+2.52}}}$ & ${\text{\textcolor{red}{+1.01}}}$ &
        ${\text{\textcolor{red}{-1.38}}}$ & ${\text{\textcolor{red}{-1.39}}}$ & ${\text{\textcolor{red}{-1.38}}}$ \\
    & ${\text{\textcolor{blue}{\textbf{0.1250}}}}$ & ${\text{\textcolor{blue}{\textbf{0.6250}}}}$ & ${\text{\textcolor{blue}{\textbf{0.0703}}}}$ &
      ${\text{\textcolor{blue}{\textbf{0.8318}}}}$ & ${\text{\textcolor{blue}{\textbf{0.1325}}}}$ & ${\text{\textcolor{blue}{0.3581}}}$ &
      ${\text{\textcolor{blue}{0.2266}}}$ & ${\text{\textcolor{blue}{0.4421}}}$ & ${\text{\textcolor{blue}{\textbf{0.1433}}}}$ \\
    \hline\hline
\end{tabular}}
\vspace{1mm}
\caption{{\color{red} Visual Effect} ({\color{red} VE}) and {\color{blue} $p\text{-value}$} from McNemar's test of only MCoT cases for each model with intervention on visual components $\mathbf{V}$ by random noise replacement. An asterisk (*) denotes statistical significance with {\color{blue} $p\text{-value} < 0.05$}.}
\label{tab:visual_inter_randnoise}
\end{table}

\begin{figure}[t]
    \centering
    \includegraphics[width=\linewidth]{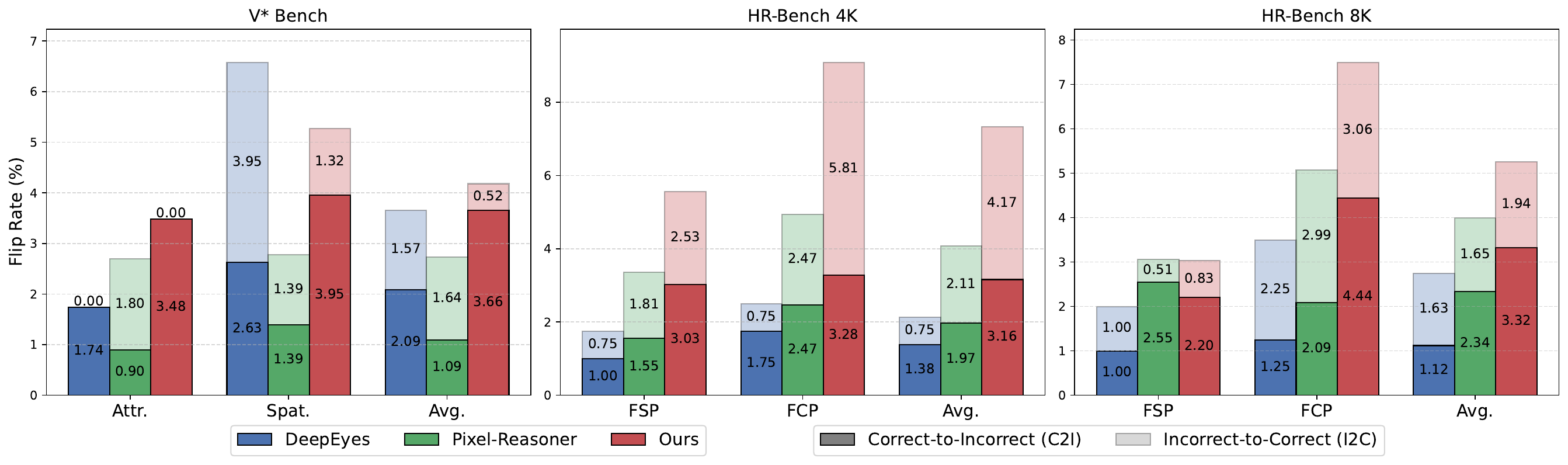}
    \vspace{-6mm}
    \caption{Visual intervention via random noise replacement on correctness flip rates across different models and benchmarks, including \textit{Correct-to-Incorrect} (C2I) and \textit{Incorrect-to-Correct} (I2C).}\label{fig:interv_v_randnoise_correctness_flip}
\vspace{-5mm}
\end{figure}

\vspace{-3mm}
\subsection{Results of Visual Intervention via Alternative Approaches} \label{sec:visual_inter_alt}
\vspace{-2mm}

In addition to the mislocalized image-region replacement used in the main experiments, we consider two alternative visual interventions: (1) Random noise replacement. We replace cropped image regions with random noise, which removes visual information while preserving the overall reasoning context. (2) GT-guided masking. When a crop contains an annotated task-relevant region, \eg, the ground-truth bounding box of a target object, we mask the overlapping evidence area. Otherwise, we apply a randomly located, size-bounded mask within the crop, ensuring that every MCoT case receives an intervention. This design leaves the crop location, scale, and resolution unchanged while masking only a bounded region and preserving the remaining visual context within the crop. Since only \textit{V*} Bench provides task-relevant ground-truth bounding-box annotations, we conduct GT-guided masking only on \textit{V*} Bench.

\begin{figure*}[t]
    \centering
    \begin{minipage}[t]{0.475\textwidth}
        \vspace{2mm}
        \centering
        \setlength{\tabcolsep}{2.5pt}
        \scriptsize
        {\renewcommand{\arraystretch}{1.2}
        \begin{tabular}{@{}p{50pt}||>{\centering\arraybackslash}p{32pt}|>{\centering\arraybackslash}p{32pt}|>{\centering\arraybackslash}p{32pt}@{}}
            \hline\hline
            \multirow{2}{*}{Model} & \multicolumn{3}{c}{\textit{V*} Bench} \\
            \cline{2-4}
                  & \textbf{Attr.} & \textbf{Spat.} & \textbf{Avg.} \\
            \hline
            \multirow{2}{*}{DeepEyes} &
                ${\text{\textcolor{red}{0.00}}}$ & ${\text{\textcolor{red}{-1.32}}}$ & ${\text{\textcolor{red}{-0.52}}}$ \\
            & ${\text{\textcolor{blue}{1.0000}}}$ & ${\text{\textcolor{blue}{1.0000}}}$ & ${\text{\textcolor{blue}{1.0000}}}$ \\
            \hline
            \multirow{2}{*}{Pixel-Reasoner} &
                ${\text{\textcolor{red}{-0.90}}}$ & ${\text{\textcolor{red}{-1.39}}}$ & ${\text{\textcolor{red}{-1.09}}}$ \\
            & ${\text{\textcolor{blue}{1.0000}}}$ & ${\text{\textcolor{blue}{1.0000}}}$ & ${\text{\textcolor{blue}{0.6250}}}$ \\
            \hline
            \multirow{2}{*}{Ours} &
                ${\text{\textcolor{red}{\textbf{-2.61}}}}$ & ${\text{\textcolor{red}{\textbf{-2.63}}}}$ & ${\text{\textcolor{red}{\textbf{-2.62}}}}$ \\
            & ${\text{\textcolor{blue}{\textbf{0.2500}}}}$ & ${\text{\textcolor{blue}{\textbf{0.6250}}}}$ & ${\text{\textcolor{blue}{\textbf{0.1250}}}}$ \\
            \hline\hline
        \end{tabular}}
        \vspace{0.5mm}
        \captionof{table}{{\color{red} Visual Effect} ({\color{red} VE}) and {\color{blue} $p\text{-value}$} from McNemar's test of only MCoT cases for each model with intervention on visual components $\mathbf{V}$ by GT-guided masking on \textit{V*} Bench.}
        \label{tab:visual_inter_mask_gt_bbox}
    \end{minipage}
    \hfill
    \begin{minipage}[t]{0.475\textwidth}
        \vspace{-1.5mm}
        \centering
        \includegraphics[width=0.794\linewidth]{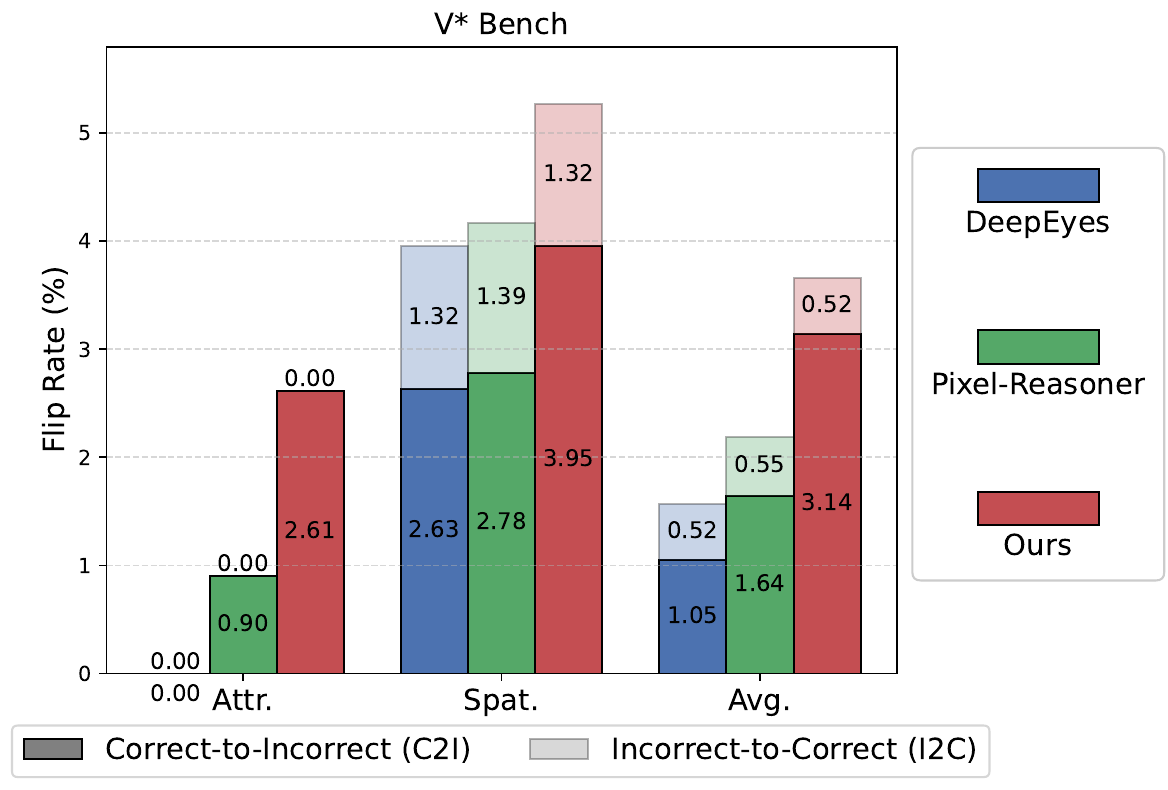}
        \vspace{-2mm}
        \caption{Correctness flip rates under GT-guided masking on \textit{V*} Bench, including \textit{Correct-to-Incorrect} (C2I) and \textit{Incorrect-to-Correct} (I2C).}
        \label{fig:visual_interv_mask_gt_bbox_correctness_flip}
    \end{minipage}
    \vspace{-2mm}
\end{figure*}

As shown in Tables \ref{tab:visual_inter_randnoise} and \ref{tab:visual_inter_mask_gt_bbox}, baseline models such as DeepEyes and Pixel-Reasoner exhibit relatively small visual effects, implying limited influence of visual components on the underlying reasoning process. In contrast, our model achieves larger absolute visual effect than the baselines, along with lower $p$-values, suggesting a stronger influence of visual information in MCoT reasoning.

Although the exact effect magnitudes vary across intervention strategies, mislocalized image-region replacement, random noise replacement, and GT-guided masking yield the same overall comparative trend. Our model generally exhibits larger absolute visual effects under all three interventions, indicating that the observed pattern is not tied to a single intervention approach.
The correctness flip rates under random noise replacement and GT-guided masking in Figures \ref{fig:interv_v_randnoise_correctness_flip} and \ref{fig:visual_interv_mask_gt_bbox_correctness_flip}, respectively, likewise follow the pattern observed in Appendix \ref{sec:correctness_flip_rate}.

\clearpage
\begin{table}[H]
  \centering
  \begin{subtable}[t]{0.95\textwidth}  %
    \centering
    \subcaption{Relevance evaluation results.} \label{tab:rely_sub_abl_rwd}  %
    { %
  \renewcommand{\arraystretch}{1.15} %
    \begin{tabular}{@{}l||c|c|c||c|c|c||c|c|c@{}}
      \hline\hline
      \multirow{2}{*}{\makecell{Reward \\ Scheme}}   & \multicolumn{3}{c||}{\textit{V*} Bench} & \multicolumn{3}{c||}{HR-Bench 4K}  & \multicolumn{3}{c}{HR-Bench 8K} \\
      \cline{2-10}
                              & \textbf{Attr.} & \textbf{Spat.} & \textbf{Avg.}  & \textbf{FSP}            & \textbf{FCP}            & \textbf{Avg.}           & \textbf{FSP}   & \textbf{FCP}   & \textbf{Avg.}  \\
      \hline
      Naive     & 27.62 & 7.69  & 20.00 & 27.24          & 19.72           & 23.63          & 16.04 & 7.97  & 12.19 \\
      \hline      
      Curiosity    & 6.14 & 0.00 & 3.70 & 18.29 & 15.32          & 16.84 & 13.60 & 7.85 & 10.78 \\
      \hline \hline
      SCCM   & \textbf{82.61} & \textbf{28.95}  & \textbf{61.26} & \textbf{50.76}          & \textbf{24.24} & \textbf{37.50}          & \textbf{40.77} & \textbf{17.78} & \textbf{29.32} \\
      \hline
      \makecell[l]{w/o * \\ Minimality} & 30.43 & 18.42 & 25.65 &   37.25  &  44.25  & 40.75    & 40.75 & 27.50 & 40.00           \\
      \hline\hline
    \end{tabular}
    }
  \end{subtable}
  \begin{subtable}[t]{0.95\textwidth}  %
    \centering
    \subcaption{Sufficiency evaluation results.} \label{tab:suff_sub_abl_rwd}  %
    { %
  \renewcommand{\arraystretch}{1.15} %
    \begin{tabular}{@{}l||c|c|c||c|c|c||c|c|c@{}}
      \hline\hline
      \multirow{2}{*}{\makecell{Reward \\ Scheme}}           & \multicolumn{3}{c||}{\textit{V*} Bench} & \multicolumn{3}{c||}{HR-Bench 4K}  & \multicolumn{3}{c}{HR-Bench 8K} \\
      \cline{2-10}
                              & \textbf{Attr.} & \textbf{Spat.} & \textbf{Avg.}  & \textbf{FSP}            & \textbf{FCP}            & \textbf{Avg.}           & \textbf{FSP}   & \textbf{FCP}   & \textbf{Avg.}  \\
      \hline
      Naive     & 31.43 & 24.61  & 28.82 & 39.41          & 30.70           & 35.16      & 30.07 & 23.35  & 26.87 \\
      \hline      
      Curiosity    & 10.53 & 4.00 & 7.94 & 31.58 & 22.34          & 27.04 & 26.45 & 16.23 & 21.44 \\
      \hline \hline
      SCCM   & \textbf{89.56} & \textbf{55.26}  & \textbf{75.92} & \textbf{70.45}          & \textbf{32.32} & \textbf{51.39}  & \textbf{65.01} & \textbf{24.17} & \textbf{44.67} \\
      \hline      
      \makecell[l]{w/o * \\ Minimality} & 59.13 & 56.58 & 58.11 &   56.00  &  61.25  & 58.63    & 49.25 & 58.00 & 53.63           \\
      \hline\hline
    \end{tabular}
    }
  \end{subtable}\par
  \vspace{1mm}
  \caption{Relevance and sufficiency evaluation results of visual components on \textit{V*} Bench and HR-Bench of the ablation models under different reward schemes: (1) Naive, consisting only of accuracy and format rewards; (2) Curiosity, the curiosity-driven reward scheme proposed in \citep{su2025pixel}; (3) SCCM, our proposed SCCM scheme with visual information sufficiency and minimality constraint; (4) SCCM w/o Minimality, an ablation variant of SCCM without the minimality constraint. We note that \emph{SCCM w/o Minimality} (denoted *), is not comparable, as its cropped image is excessively large, being the same size as the original input image.}  %
  \label{tab:ablmodels_eval}  %
\end{table}

\begin{table}[H]
\setlength{\tabcolsep}{5pt}
    \vspace{-1mm}
    \centering
    {
      \renewcommand{\arraystretch}{1.5} %
      \begin{tabular}{@{}l||c|c||c|c||c|c@{}}
          \hline\hline
          \multirow{2}{*}{\makecell{Reward \\ Scheme}} & \multicolumn{2}{c||}{\textit{V*} Bench} & \multicolumn{2}{c||}{HR-Bench 4K} & \multicolumn{2}{c}{HR-Bench 8K} \\
          \cline{2-7}
                                  & \textbf{CRZ} & \textbf{TCC} & \textbf{CRZ} & \textbf{TCC} & \textbf{CRZ} & \textbf{TCC} \\
          \hline
          Naive                & 0.1490 & 1.4345 & 0.2176 & 1.4587 & 0.1680 & 1.1862 \\
          \hline      
          Curiosity               & 0.0835 & 0.9895  & 0.1977 & 0.9850 & 0.2144 & 0.9750 \\
          \hline\hline
          SCCM                      & 0.0429 & 1.0000 & 0.1429 & 0.9900 & 0.1273 & 0.9037 \\
          \hline      
          \makecell[l]{w/o * \\ Minimality} & 1.9916 & 2.0000 & 1.9983 & 2.0000 & 1.9682 & 2.0025 \\
          \hline\hline
      \end{tabular}
    }
    \vspace{1mm}
    \caption{The Cropped Region Size (CRZ) and Tool Call Count (TCC) on \textit{V*} Bench and HR-Bench of the ablation models under different reward schemes.} \label{tab:crz_tcc}
    \vspace{-4mm}
\end{table}

\vspace{-2mm}
\subsection{Results of Ablation Models under Different Reward Schemes} \label{app:ablmodels_eval_res}
\vspace{-2mm}

We evaluate the relevance and sufficiency of visual components on \textit{V*} Bench \citep{wu2024v} and HR-Bench \citep{wang2025divide}, with the ablation models under different reward schemes in Sect. \ref{sec:ablation}, illustrated in Table \ref{tab:ablmodels_eval}.
We also show the Cropped Region Size (CRZ), \ie, the total aspect ratio of cropped regions relative to the original image for assessing the visual information quantity in MCoT, and Tool Call Count (TCC), \ie, the average number of tool calls, in Table \ref{tab:crz_tcc}.
Our proposed SCCM Reward scheme consistently outperforms the Naive and Curiosity Reward in terms of relevance and sufficiency metrics, further demonstrating the superiority of the proposed SCCM scheme.
TCC above 1 under the Naive reward indicates that the model uses the zoom-in tool in nearly all cases and occasionally invokes it multiple times, while TCC below 1 under the Curiosity reward indicates that the model skips tool use in some cases and typically invokes it once when it is used. The SCCM reward keeps TCC close to 1 with smaller CRZ, suggesting broad but efficient tool usage.
The model under SCCM without Minimality Reward scheme is not comparable, due to its cropped image being excessively large, \ie, the same size as the original input image with multiple tool calls ($\approx2$).

\clearpage

\vspace{-3mm}
\subsection{Results of Ablation Models under Different Training Datasets} \label{app:data_ablmodels_eval_res}
\vspace{-1mm}

We also execute RFT using an alternative dataset, \ie, the training dataset of \citet{su2025pixel}, and evaluate the resulting model in terms of relevance and sufficiency. The evaluation results are presented in Table \ref{tab:data_ablmodels_eval}, with the cropped region size and tool call count detailed in Table \ref{tab:crz_tcc_data_ablmodel}. 
As shown in Table \ref{tab:crz_tcc_data_ablmodel}, the cropped region size and tool call count of different training dataset settings are similar, suggesting that a comparable visual information quantity is incorporated in their MCoT reasoning processes.
The relevance and sufficiency results in Table \ref{tab:data_ablmodels_eval} indicate an overall comparable performance between the two models. However, on HR-Bench, the model trained on the dataset from \citet{su2025pixel} (\textbf{Ablation}) outperforms the model trained on the dataset from \citet{zheng2025deepeyes} (\textbf{Ours}), especially on HR-Bench 4K. %
This difference may be attributed to the training data from \citet{su2025pixel}, which includes a substantial number of high-resolution images, \eg, from SA-1B \citep{kirillov2023segment}. This type of data is likely to enhance the model's ability to perceive and interpret high-resolution visual content.

\begin{table}[t]
  \centering
  \begin{subtable}[t]{0.95\textwidth}  %
  \setlength{\tabcolsep}{5pt}
    \centering
    \subcaption{Relevance evaluation results.} \label{tab:rely_sub_abl_dataset}  %
    \vspace{-2mm}
    { %
  \renewcommand{\arraystretch}{1.5} %
    \begin{tabular}{@{}l||c|c|c||c|c|c||c|c|c@{}}
      \hline\hline
      \multirow{2}{*}{\makecell{RFT Dataset}}   & \multicolumn{3}{c||}{\textit{V*} Bench} & \multicolumn{3}{c||}{HR-Bench 4K}  & \multicolumn{3}{c}{HR-Bench 8K} \\
      \cline{2-10}
                              & \textbf{Attr.} & \textbf{Spat.} & \textbf{Avg.}  & \textbf{FSP}            & \textbf{FCP}            & \textbf{Avg.}           & \textbf{FSP}   & \textbf{FCP}   & \textbf{Avg.}  \\
      \hline
      \makecell[l]{\citet{zheng2025deepeyes}\\Dataset (\textbf{Ours})}   & \textbf{82.61} & 28.95  & \textbf{61.26} & 50.76          & 24.24 &  37.50        & 40.77 & 17.78 & 29.32 \\
      \hline
      \makecell[l]{\citet{su2025pixel} \\ Dataset (\textbf{Ablation})} & 78.07 & \textbf{30.67} & 59.26 &   \textbf{56.82}  &  \textbf{27.29}  & \textbf{42.13}    & \textbf{44.00} & \textbf{19.49} & \textbf{31.82}           \\
      \hline\hline
    \end{tabular}
    }
  \end{subtable}
  \begin{subtable}[t]{0.95\textwidth}  %
  \setlength{\tabcolsep}{5pt}
    \centering
    \subcaption{Sufficiency evaluation results.} \label{tab:suff_sub_abl_dataset}  %
    \vspace{-2mm}
    { %
  \renewcommand{\arraystretch}{1.5} %
    \begin{tabular}{@{}l||c|c|c||c|c|c||c|c|c@{}}
      \hline\hline
      \multirow{2}{*}{\makecell{RFT Dataset}}           & \multicolumn{3}{c||}{\textit{V*} Bench} & \multicolumn{3}{c||}{HR-Bench 4K}  & \multicolumn{3}{c}{HR-Bench 8K} \\
      \cline{2-10}
                              & \textbf{Attr.} & \textbf{Spat.} & \textbf{Avg.}  & \textbf{FSP}            & \textbf{FCP}            & \textbf{Avg.}           & \textbf{FSP}   & \textbf{FCP}   & \textbf{Avg.}  \\
      \hline
      \makecell[l]{\citet{zheng2025deepeyes}\\Dataset (\textbf{Ours})}   & \textbf{89.56} & 55.26 & \textbf{75.92} & 70.45          & 32.32 & 51.39  & \textbf{65.01} & 24.17 & \textbf{44.67} \\
      \hline      
      \makecell[l]{\citet{su2025pixel} \\ Dataset (\textbf{Ablation})} & 87.72 & \textbf{56.00} & 75.13 &   \textbf{73.74}  &  \textbf{39.79}  & \textbf{56.85}    & 57.25 & \textbf{28.10} & 42.77           \\
      \hline\hline
    \end{tabular}
    }
  \end{subtable}\par
  \vspace{1mm}
  \caption{Relevance and sufficiency evaluation results of visual components on \textit{V*} Bench and HR-Bench for models trained via RFT: (1) on the training dataset of \citet{zheng2025deepeyes} (\ie, \textbf{Ours}), and (2) on the training dataset of \citet{su2025pixel} (\ie, \textbf{Ablation}).} \label{tab:data_ablmodels_eval}  %
  \vspace{-5mm}
\end{table}

\begin{table}[t]
\setlength{\tabcolsep}{5pt}
    \centering
    {
      \renewcommand{\arraystretch}{1.5} %
      \begin{tabular}{@{}l||c|c||c|c||c|c@{}}
          \hline\hline
          \multirow{2}{*}{\makecell{RFT Dataset}} & \multicolumn{2}{c||}{\textit{V*} Bench} & \multicolumn{2}{c||}{HR-Bench 4K} & \multicolumn{2}{c}{HR-Bench 8K} \\
          \cline{2-7}
                                  & \textbf{CRZ} & \textbf{TCC} & \textbf{CRZ} & \textbf{TCC} & \textbf{CRZ} & \textbf{TCC} \\
          \hline
          \makecell[l]{\citet{zheng2025deepeyes}\\Dataset (\textbf{Ours})} & 0.0429 & 1.0000 & 0.1429 & 0.9900 & 0.1273 & 0.9037 \\
          \hline      
          \makecell[l]{\citet{su2025pixel}\\Dataset (\textbf{Ablation})} & 0.0442 & 0.9895 & 0.1584 & 0.9900 & 0.1553 & 0.9950 \\
          \hline\hline
      \end{tabular}
    }
    \vspace{1mm}
    \caption{The Cropped Region Size (CRZ), \ie, the total aspect ratio of cropped regions relative to the original image, and Tool Call Count (TCC), \ie, the average number of zoom-in tool calls, on \textit{V*} Bench and HR-Bench of the ablation models under different RFT training datasets.} \label{tab:crz_tcc_data_ablmodel}
    \vspace{-6mm}
\end{table}

\begin{table}[t]
    \centering
    { %
  \renewcommand{\arraystretch}{1.5} %
    \begin{tabular}{@{}l||c|c|c||c|c|c||c|c|c@{}}
      \hline\hline
      \multirow{2}{*}{Model}          & \multicolumn{3}{c||}{\textit{V*} Bench} & \multicolumn{3}{c||}{HR-Bench 4K}  & \multicolumn{3}{c}{HR-Bench 8K} \\
      \cline{2-10}
                                      & \textbf{Attr.} & \textbf{Spat.} & \textbf{Avg.}  & \textbf{FSP}            & \textbf{FCP}            & \textbf{Avg.}           & \textbf{FSP}   & \textbf{FCP}   & \textbf{Avg.}  \\
      \hline      
      SEAL                            & 73.04          & 75.00            & 73.82            & 40.00          & 28.00          & 34.00            & 42.00          & 31.00          & 36.50   \\
      \hline
      DeepEyes                        & \underline{90.43}          & \underline{86.84}            & \underline{89.00}            & \textbf{86.75}          & \textbf{65.75}          & \textbf{76.25}            & 85.50          & \underline{56.00}          & \underline{70.75}            \\
      \hline      
      Pixel-Reasoner                  & 88.69          & 81.58            & 85.86            & 83.50          & \underline{60.00} & 71.75            & \underline{86.25}          & 53.50 & 69.87            \\
      \hline      
      Ours                            & \textbf{93.91} & \textbf{86.84}   & \textbf{91.10}   & \underline{86.00} & 59.00          & \underline{72.50}   & \textbf{86.50} & \textbf{56.00}          & \textbf{71.25} \\
      \hline\hline
    \end{tabular}}
    \vspace{1mm}
    \caption{The accuracy of different models on \textit{V*} Bench and HR-Bench, covering all samples. \textbf{Bold} and \underline{Underlined} denote the best and second-best results.} \label{tab:acc_results_diffmodels}
    \vspace{-4mm}
\end{table}

\begin{table}[t]
\setlength{\tabcolsep}{5pt}
    \vspace{-1mm}
    \centering
    {
      \renewcommand{\arraystretch}{1.5} %
      \begin{tabular}{@{}l||c|c||c|c||c|c@{}}
          \hline\hline
          \multirow{2}{*}{Model} & \multicolumn{2}{c||}{\textit{V*} Bench} & \multicolumn{2}{c||}{HR-Bench 4K} & \multicolumn{2}{c}{HR-Bench 8K} \\
          \cline{2-7}
                                  & \textbf{CRZ} & \textbf{TCC} & \textbf{CRZ} & \textbf{TCC} & \textbf{CRZ} & \textbf{TCC} \\
          \hline
          DeepEyes                & 0.0074 & 1.0000 & 0.0371 & 1.0250 & 0.0256 & 1.0000 \\
          \hline      
          Pixel-Reasoner               & 0.0988 & 0.9581  & 0.1076 & 0.8875 & 0.0928 & 0.9087 \\
          \hline
          Ours                    & 0.0429 & 1.0000 & 0.1429 & 0.9900 & 0.1273 & 0.9037 \\
          \hline\hline
      \end{tabular}
    }
    \vspace{1mm}
    \caption{The Cropped Region Size (CRZ) and Tool Call Count (TCC) on \textit{V*} Bench and HR-Bench of DeepEyes \citep{zheng2025deepeyes}, Pixel-Reasoner \citep{su2025pixel} and our model from SCCM-based RFT.} \label{tab:crz_tcc_diff_models}
    \vspace{-4mm}
\end{table}

\vspace{-3mm}
\subsection{Results of Accuracy Comparison Across Different Models} \label{app:acc_results}
\vspace{-2mm}

Table \ref{tab:acc_results_diffmodels} presents the accuracy performance of DeepEyes \citep{zheng2025deepeyes}, Pixel-Reasoner \citep{su2025pixel}, our proposed SCCM-based RFT model, as well as the visual search method SEAL \citep{wu2024v}, evaluated on the \textit{V*} Bench and HR-Bench datasets. %
We note that the evaluation of Table \ref{tab:acc_results_diffmodels} covers all data samples from these benchmarks, including both MCoT and non-MCoT (\ie, text-only CoT) reasoning cases. This differs from Table \ref{tab:base_acc} where only MCoT reasoning cases are considered, and non-MCoT cases are excluded.

Our model achieves highly competitive performance on the majority of evaluated tasks, particularly on the \textit{V*} Bench and HR-Bench 8K. Furthermore, it demonstrates a clear and significant improvement over our primary baseline, \ie, Pixel-Reasoner. %
Taken together with the faithfulness results detailed in Table \ref{tab:visual_component_eval}, these results indicate that SCCM improves the targeted visual faithfulness while maintaining competitive task accuracy.

\vspace{-2mm}
\subsection{Statistics on Visual Information Quantity in MCoT by Different Models} \label{app:vis_infor_statics}
\vspace{-2mm}

We report statistics on the quantity of visual information in MCoT using two metrics: (1) the Cropped Region Size (CRZ), \ie, the total area ratio of cropped regions relative to the original image, which serves as a proxy for evaluating minimality, and (2) the Tool Call Count (TCC), \ie, the average number of zoom-in tool calls.
Results are provided in Table \ref{tab:crz_tcc_diff_models} for DeepEyes \citep{zheng2025deepeyes}, Pixel-Reasoner \citep{su2025pixel}, and our SCCM-based RFT model.
Since non-MCoT responses contain zero tool calls and most MCoT responses empirically contain a single zoom-in tool call, the average tool call count serves as a proxy for MCoT usage rate. As shown in Table \ref{tab:crz_tcc_diff_models}, the compared models generally exhibit similar tool usage, suggesting that the observed faithfulness improvements are not merely due to changes in which examples generate MCoT.

Combining the results in Table \ref{tab:visual_component_eval}, it suggests that DeepEyes incorporates extremely small cropped regions, which often provide insufficient visual information. This likely explains its suboptimal performance in terms of relevance and sufficiency, especially in scenarios involving multiple target objects or large objects, \eg, queries from HR-Bench. In comparison, Pixel-Reasoner crops larger regions, but these often include substantial query-unrelated visual content.
In contrast, our model, using SCCM, maintains visual cues of an appropriate size, \ie, with suitable information quantity, while ensuring both the correctness and effectiveness of the visual information. %

\vspace{-2mm}
\subsection{Relationship between MCoT Faithfulness and Task Performance}\label{sec:faithful_acc_relation}
\vspace{-2mm}

\begin{figure*}[!t]
\vspace{-2mm}
    \centering
    \begin{minipage}[t]{\linewidth} 
        \subcaption{Relevance metric vs. task accuracy.}
        \includegraphics[width=\linewidth]{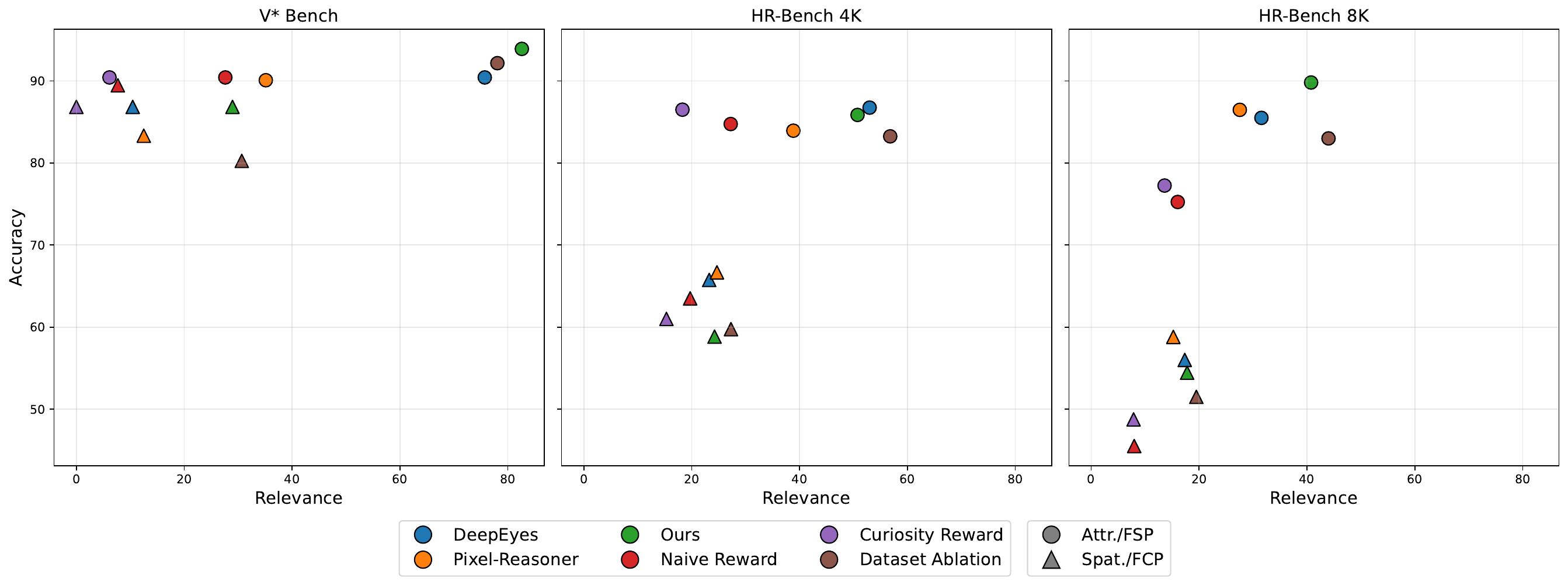}
    \end{minipage}
    \begin{minipage}[t]{\linewidth} 
        \subcaption{Sufficiency metric vs. task accuracy.}
        \includegraphics[width=\linewidth]{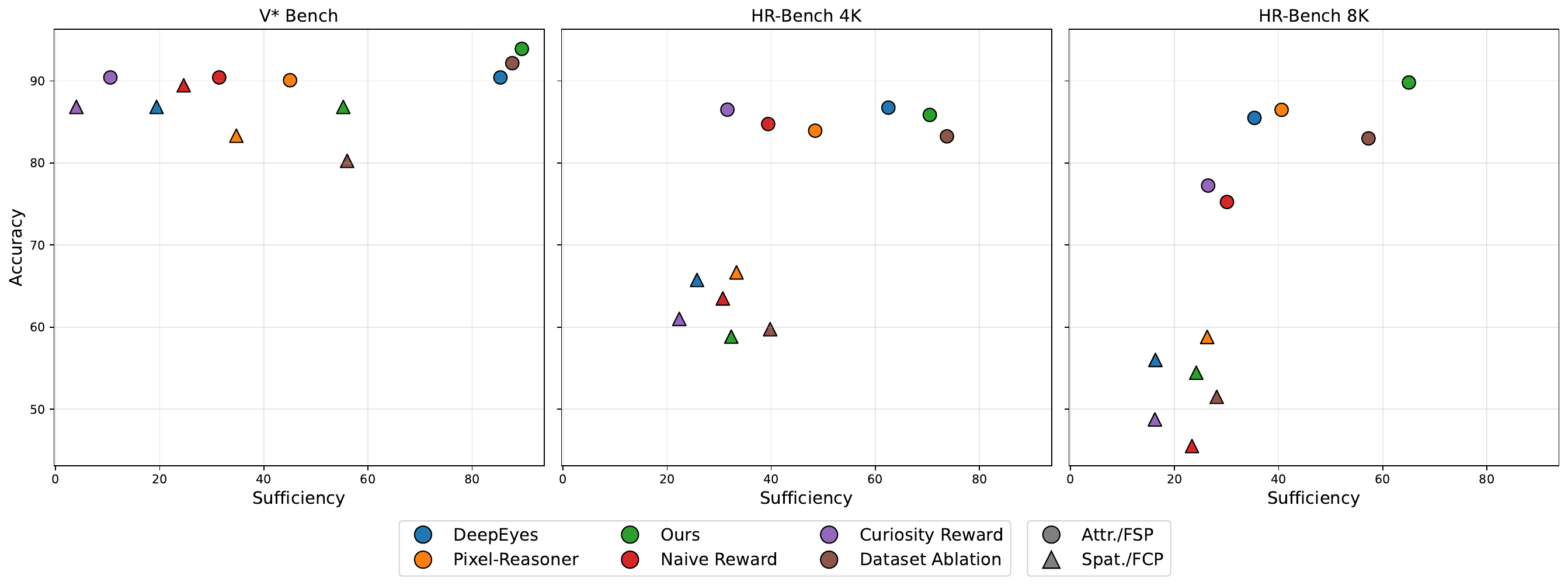}
    \end{minipage}
    \vspace{-2mm}
    \caption{Relationship between MCoT faithfulness metrics (relevance and sufficiency) and task accuracy under different models. Each point denotes one model-task pair, with accuracy computed only on MCoT cases. Overall, no consistent monotonic relationship is observed.} \label{fig:faithfulness_perf}
    \vspace{-4mm}
\end{figure*}

We further analyze the relationship between the proposed faithfulness metrics and task performance. 
Figure \ref{fig:faithfulness_perf} plots relevance and sufficiency metrics against task accuracy, where each point represents one model-task pair and accuracy is computed only on MCoT cases.
We adopt subtask-level results to better illustrate the varying relationships across different subtask types: \textit{Attr.} and \textit{Spat.} on \textit{V*} Bench, and \textit{FSP} and \textit{FCP} on HR-Bench. Specifically, \textit{Attr.}/\textit{FSP} concern the perception of a single target, while \textit{Spat.}/\textit{FCP} concern the relations among multiple targets.
The models included in this analysis cover: 1) baseline models DeepEyes and Pixel-Reasoner; 2) our SCCM-based RFT model; 3) ablation models including the Naive Reward and Curiosity Reward models; 4) dataset ablation models trained on the dataset of \citet{su2025pixel}.

Overall, we observe no consistent monotonic relationship between faithfulness and accuracy across benchmarks and task types. 
For relevance, higher scores sometimes coincide with higher accuracy, but this pattern is not consistently observed across all benchmarks.
For sufficiency, the relationship is even less stable, as higher sufficiency does not necessarily correspond to higher task accuracy. 
This observation is consistent with prior findings \citep{lanham2023measuring,tanneru2024hardness,barez2025chain} that improved faithfulness of reasoning does not always translate into improved task performance.

Nevertheless, it does not diminish the importance of faithfulness evaluation. 
Task accuracy only reflects whether the final answer is correct, whereas faithfulness metrics assess whether the generated reasoning trace is grounded in relevant and sufficient visual evidence. 
Faithfulness therefore serves as an essential evaluation dimension complementary to accuracy, rather than an interchangeable indicator. This complementarity is particularly salient in high-stakes settings where a correct answer alone is insufficient if its supporting reasoning cannot be audited.

\vspace{-2mm}
\section{More Cases}\label{app:more_cases}
\vspace{-2mm}

\bulletpara{Comparison of the Generated MCoT by Different Models}

\begin{figure}[htbp]
    \vspace{-2mm}
    \centering
    \begin{minipage}[t]{0.8\textwidth} 
        \includegraphics[width=\linewidth]{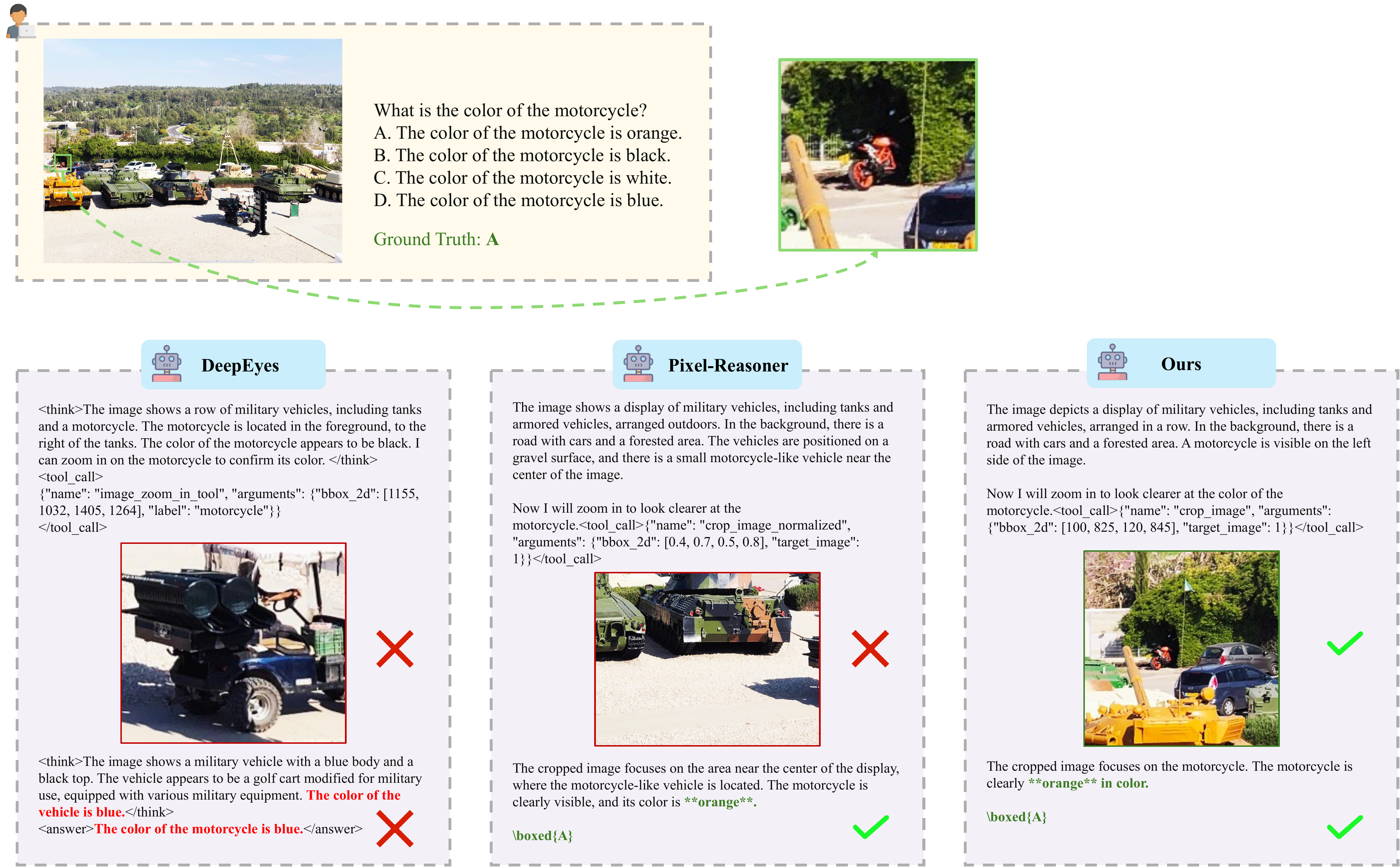}
        \subcaption{}
        \label{fig:app_goodcase1}
    \end{minipage}
    \begin{minipage}[t]{0.8\textwidth} 
        \includegraphics[width=\linewidth]{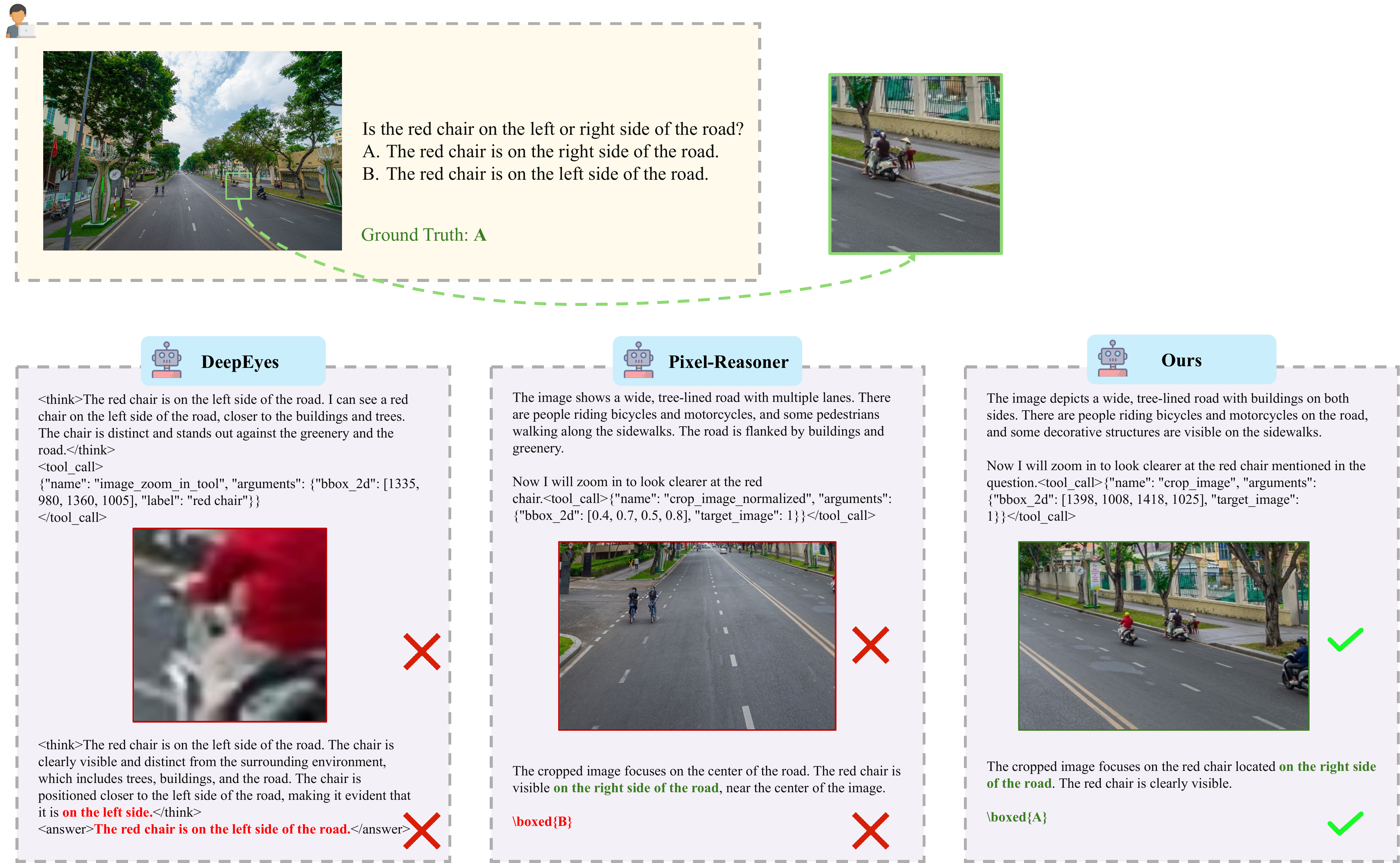}
        \subcaption{}
        \label{fig:app_goodcase2}
    \end{minipage}
    \begin{minipage}[t]{0.8\textwidth} 
        \includegraphics[width=\linewidth]{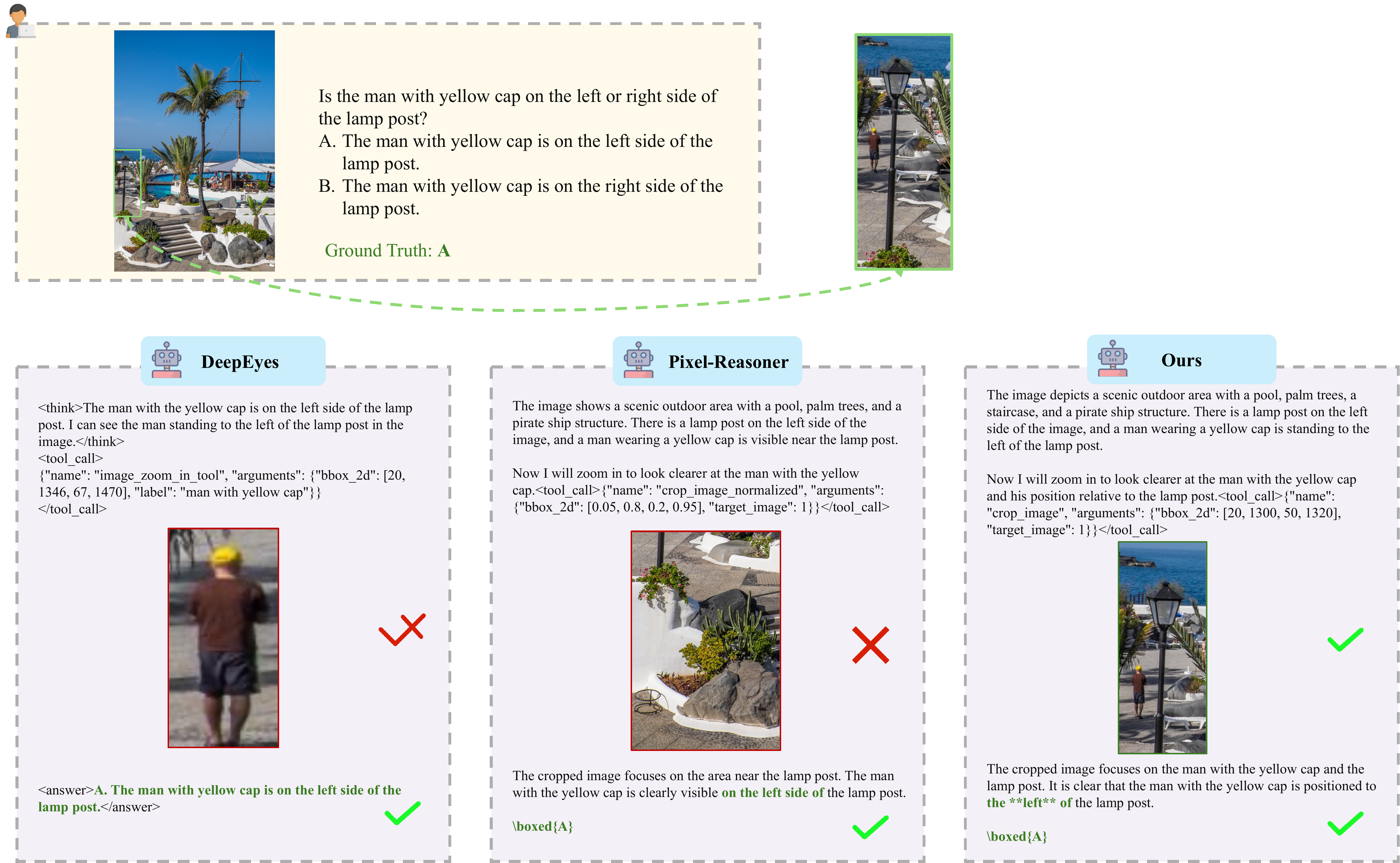}
        \subcaption{}
        \label{fig:app_goodcase3}
    \end{minipage}
    \vspace{-4mm}
    \caption{Cases in \textit{V*} Bench with MCoT responses generated by DeepEyes, Pixel-Reasoner and our SCCM-based RFT model.} \label{fig:sccm_goodcases}
\end{figure}

We provide some cases in \textit{V*} Bench with responses generated by DeepEyes \citep{zheng2025deepeyes}, Pixel-Reasoner \citep{su2025pixel} and our SCCM-based RFT model, illustrated in Figure \ref{fig:sccm_goodcases}. 
We observe that both DeepEyes and Pixel-Reasoner incorporate incorrect visual evidence during their MCoT reasoning, whereas our model integrates accurate visual cues and exhibits a more rational reasoning process.

\bulletpara{Comparison of the Generated MCoT by Models in Different Reward Schemes}

\begin{figure}[htbp]
    \vspace{-2mm}
    \centering
    \begin{minipage}[t]{\textwidth} 
        \includegraphics[width=0.95\linewidth]{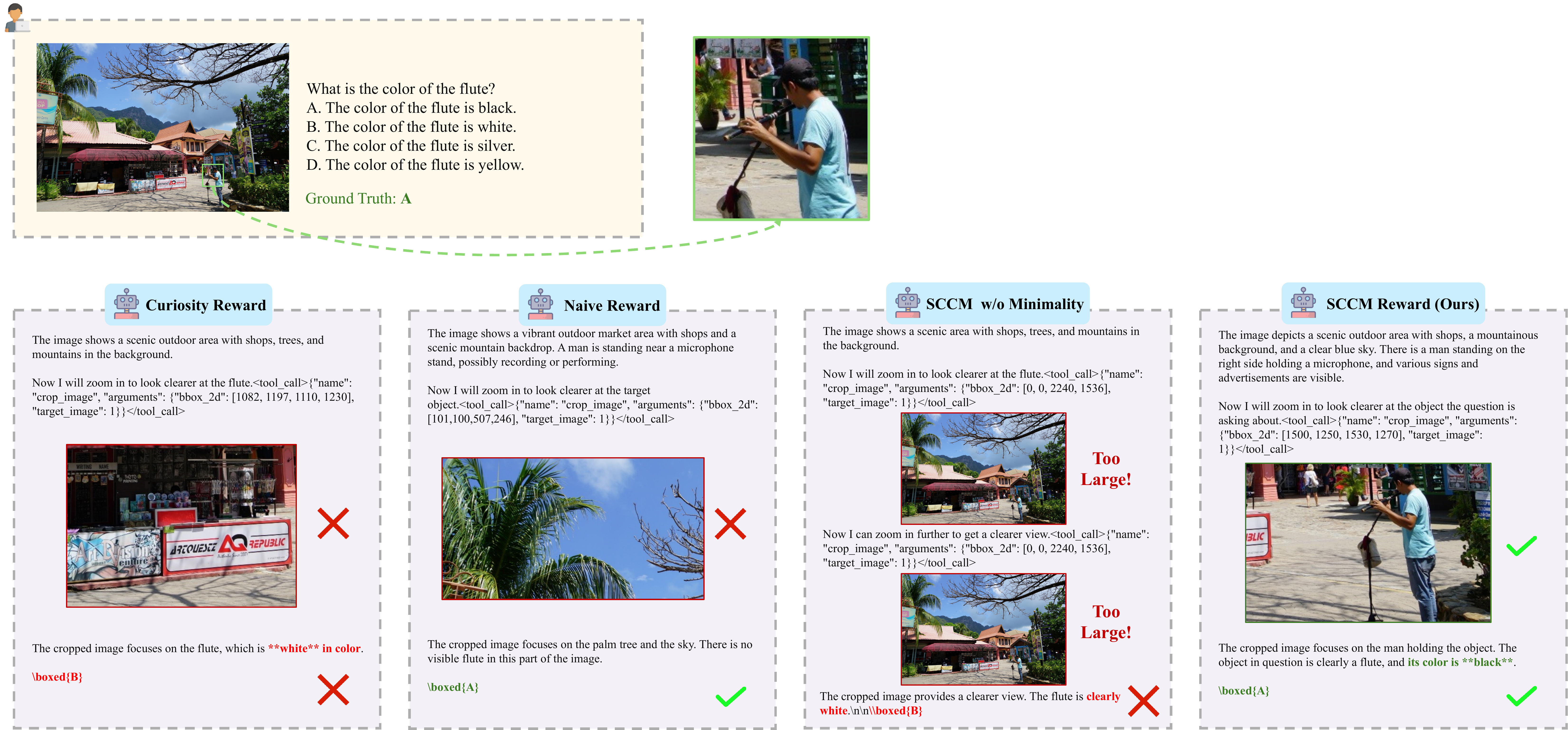}
        \subcaption{}
        \label{fig:app_abl_case1}
    \end{minipage}
    \begin{minipage}[t]{\textwidth} 
        \includegraphics[width=0.95\linewidth]{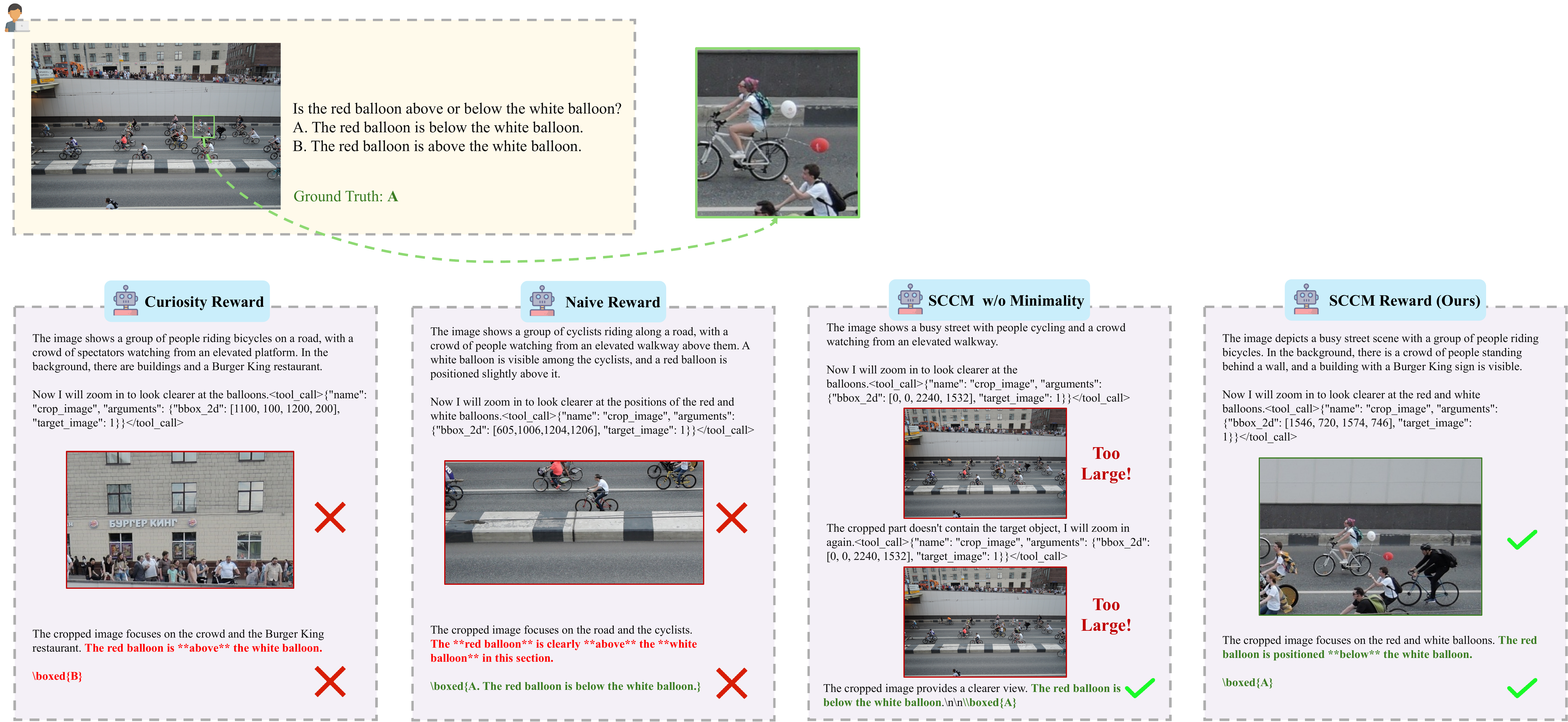}
        \subcaption{}
        \label{fig:app_abl_case2}
    \end{minipage}
    \begin{minipage}[t]{\textwidth} 
        \includegraphics[width=0.95\linewidth]{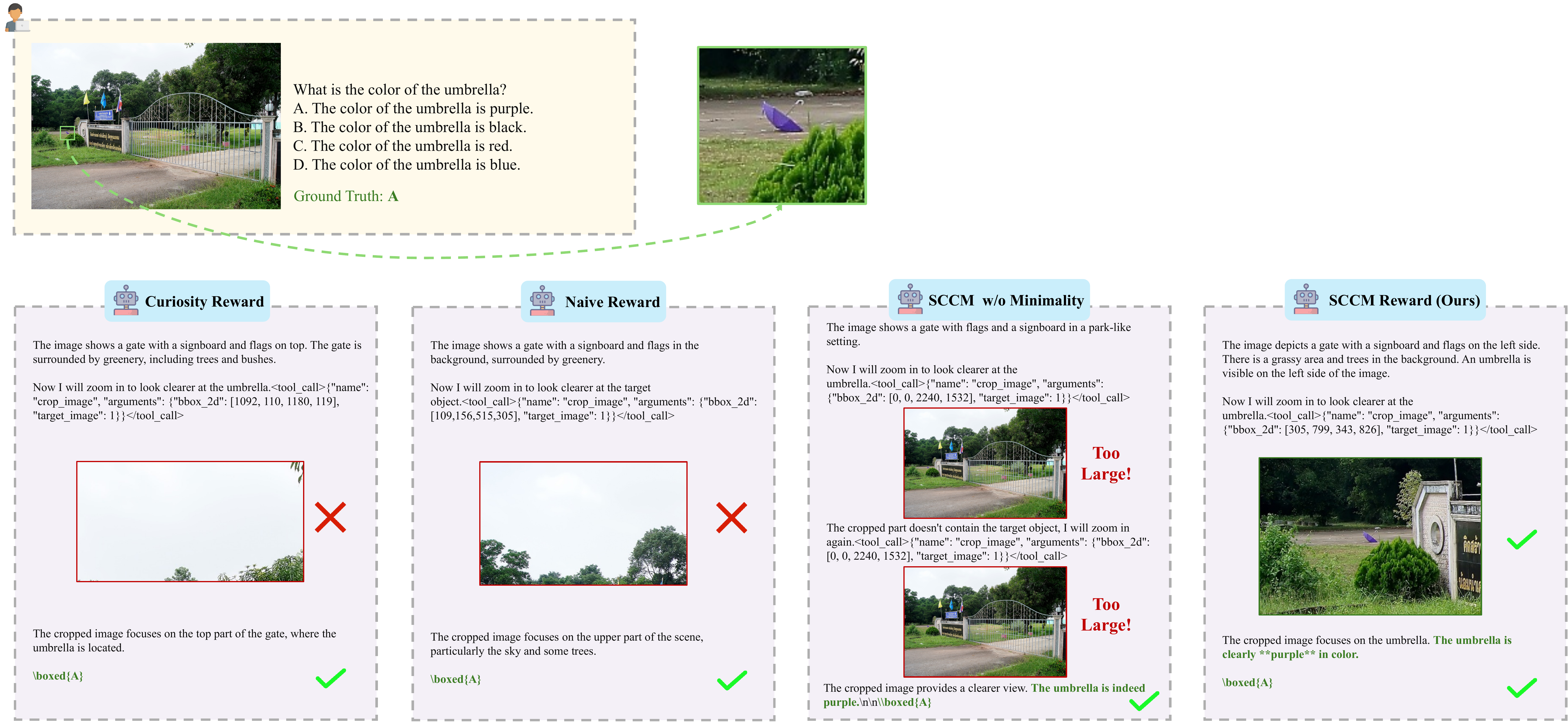}
        \subcaption{}
        \label{fig:app_abl_case3}
    \end{minipage}
    \vspace{-3mm}
    \caption{Cases in \textit{V*} Bench with MCoT responses generated by models in different reward schemes: (1) Curiosity Reward, the curiosity-driven reward scheme proposed in \citep{su2025pixel}; (2) Naive Reward, consisting only of accuracy and format rewards; (3) SCCM w/o Minimality, an ablation variant of SCCM without the minimality constraint; (4) SCCM Reward, our proposed SCCM scheme with visual information sufficiency and minimality constraint.} \label{fig:ablation_cases}
\end{figure}

Cases in \textit{V*} Bench with responses generated by models in different reward schemes are shown in Figure \ref{fig:ablation_cases}. Models under curiosity reward and naive reward both exhibit incorrect visual cues, and yield predicted answers that disregard these cues, indicating that the absence of supervision on the involved visual information in MCoT can easily lead to the problems of inaccurate visual cues and unfaithful reasoning processes.
In contrast, the model trained with SCCM reward demonstrates more accurate visual cues and a more rational reasoning process. However, without the minimality constraint, it incorporates the full original input image, which is excessively large, even twice in the MCoT. This suggests that the image sufficiency reward alone can be easily exploited, underscoring the necessity of the minimality constraint in the SCCM scheme.

\section{Limitations} \label{sec:limitations}

Firstly, this paper focuses on MCoT via zoom-in tool calls, a representative and widely used paradigm. Other forms, such as visual sketching or image generation, remain unexplored.
The proposed SCCM learning is formulated around general notions of visual sufficiency and minimality, and is conceptually compatible with other MCoT paradigms, though its concrete implementation and effectiveness in those settings require future investigation.
Nonetheless, our study of this foundational paradigm paves the way for broader extensions.

Secondly, our experiments focus on the 7B scale, which is a widely adopted and accessible setting in current MCoT research, while larger models are less accessible for exhaustive testing. Importantly, reasoning unfaithfulness remains a common challenge despite larger scales, as evidenced by prior work \citep{bentham2024chain,han2026rfeval}. Our proposed evaluation framework and SCCM learning are model-agnostic and serve as a scalable diagnostic and training framework that could in principle be applied to larger models when such systems become available for controlled experimentation. We thus leave large-scale empirical validation as an important direction for future research.

Thirdly, our SCCM reward focuses on the semantic sufficiency and conciseness of visual components, rather than explicit boundary-level localization accuracy. This differs from IoU-based rewards, \eg, TreeVGR \citep{wang2025traceable}, which enforce precise spatial alignment by requiring accurate bounding boxes for target objects. As a result, SCCM may be less directly aligned with localization-centric evaluation settings where precise spatial localization is the primary objective. For example, in queries involving overlapping objects, IoU-based rewards typically require separate and accurately localized bounding boxes for each target object, whereas SCCM may still assign a high reward to a single crop covering all relevant objects if it semantically supports the correct answer. Therefore, for tasks where precise object localization is essential, IoU-based rewards and SCCM may be complementary, and combining semantic sufficiency with geometric grounding supervision remains an important future direction.

\section{Broader Impacts}

This work enhances the faithfulness of visual reasoning in Multimodal Chain-of-Thought, which can improve transparency and support more calibrated trust in high-stakes domains such as medical image analysis and autonomous driving. In such settings, understanding the visual evidence behind model reasoning is essential for auditing and human oversight. On the other hand, even more faithful visual explanations may trigger automation bias, leading users to over-rely on model outputs and reduce manual review. Thus, human supervision remains advisable for critical decisions, and clear communication of model limitations is necessary to prevent misuse. Overall, the risks are manageable with appropriate safeguards.

\end{document}